\documentclass[twoside,11pt]{article}

\usepackage{jmlr2e}

\usepackage{amsfonts,amsmath,latexsym,amssymb}
\usepackage{graphicx}
\usepackage[font=footnotesize]{subcaption}
\usepackage{acronym}
\usepackage{algorithm}
\usepackage{psfrag}
\usepackage{xcolor}
\usepackage{dsfont}
\usepackage{adjustbox}
\usepackage[noend]{algpseudocode}
\usepackage{nicefrac}
\usepackage{setspace}
\usepackage[colorinlistoftodos]{todonotes}\reversemarginpar
\usepackage{multicol}
\usepackage{multirow}
\usepackage{xr-hyper}
\usepackage{wrapfig}
 \usepackage{hyperref}
 \usepackage{breakurl}
 \usepackage{zref-base}
\usepackage{atveryend}
  \makeatletter
\AfterLastShipout{%
  \if@filesw   
    \begingroup
      \zref@setcurrent{default}{\the\value{equation}}%
      \zref@wrapper@immediate{%
        \zref@labelbyprops{eq-last}{default}%
      }%
            \zref@setcurrent{default}{\the\value{figure}}%
      \zref@wrapper@immediate{%
        \zref@labelbyprops{fig-last}{default}%
      }%
    \endgroup
  \fi
}
\makeatother
 
\newtheorem{teorema}{Theorem}

\newcommand\V[1]  { \mathbf{#1} }
\newcommand\B[1]  { \boldsymbol{#1} }

\newcommand\set[1] {\mathcal{#1}}
\newcommand\up[1] {\mathrm{#1}}

\definecolor{mypink}{rgb}{1, 0.288, 1}
\definecolor{myorange}{rgb}{1, 0.647, 0.208}
\definecolor{myblue}{rgb}{0.169, 0.282, 0.847}
\definecolor{myblue1}{rgb}{0.502, 0.702, 1}
\definecolor{myblue2}{rgb}{0, 0, 1}
\definecolor{myblue3}{rgb}{0,0, 1}
\definecolor{mygreen}{rgb}{0.274, 0.647, 0.208}
\definecolor{mygreen2}{rgb}{0, 0.738, 1}
\definecolor{myred1}{rgb}{1,0.5, 0.5}
\definecolor{myred2}{rgb}{0,0.676, 0}
\definecolor{myred3}{rgb}{0.83,0, 0}
\definecolor{mypurple2}{rgb}{0.586,0.863, 0.414}

\acrodef{SGD}[Nesterov's-SG]{subgradient method}
\acrodef{ASM}[ASM]{accelerated subgradient method}
\acrodef{AMRC}[AMRC]{adaptive minimax risk classifier}
\acrodef{MRC}[MRC]{minimax risk classifier}
\acrodef{ERM}[ERM]{empirical risk minimization}
\acrodef{RKHS}[RKHS]{reproducing kernel Hilbert space}
\acrodef{RRM}[RRM]{robust risk minimization}
\acrodef{DWM}[DWM]{dynamic weighted majority}
\acrodef{FOGD}[FOGD]{Fourier online gradient descent}
\acrodef{SP}[SP]{shifting perceptron}
\acrodef{RBP}[RBP]{randomized budget perceptron}
\acrodef{NOGD}[NOGD]{Nystr\"om online gradient descent}
\acrodef{OGD}[OGD]{online Gradient Descent}
\acrodef{RMSE}[RMSE]{root mean square error}
\acrodef{NORMA}[NORMA]{naive online regularized risk minimization algorithm}
\acrodef{RFF}[RFF]{random Fourier features}
\acrodef{MSE}[MSE]{mean squared error}
\acrodef{ess}[ESS]{effective sample size}
\acrodef{ood}[OOD]{out-of-distribution}
\acrodef{id}[ID]{in-distribution}
\acrodef{AUE}[AUE]{accuracy updated ensemble}
\acrodef{VMRC}[VMRC]{methods for varying distribution based on minimax risk classifier}
\acrodef{ELLA}[ELLA]{efficient continual learning algorithm}
\acrodef{GEM}[GEM]{gradient episodic memory}
\acrodef{LMRC}[LMRC]{continual learning methods based on minimax risk classifier}
\acrodef{MER}[MER]{meta-experience replay}
\acrodef{EWC}[EWC]{elastic weight consolidation}
\acrodef{MDA}[MDA]{multi-source domain adaptation}
\acrodef{MTL}[MTL]{multi-task learning}
\acrodef{SCD}[SCD]{supervised classification under concept drift}
\acrodef{CL}[CL]{continual learning}

\usepackage{lastpage}
\jmlrheading{26}{2025}{1-\pageref{LastPage}}{3/24; Revised
1/25}{1/25}{24-0343}{Ver\'onica \'Alvarez, Santiago Mazuelas, and Jose A. Lozano}
\ShortHeadings{Supervised Learning with Evolving Tasks and Performance Guarantees}{\'Alvarez, Mazuelas, and Lozano}

\title{Supervised Learning with Evolving Tasks and \\Performance Guarantees}

\begin{document}
 \sloppy

\author{\name Ver\'onica \'Alvarez \email valvarez@bcamath.org \\
       \addr Basque Center for Applied Mathematics (BCAM)\\
       Bilbao 48009, Spain
       \AND
       \name Santiago Mazuelas \email smazuelas@bcamath.org \\
       \addr Basque Center for Applied Mathematics (BCAM)\\
             IKERBASQUE-Basque Foundation for Science\\
       Bilbao 48009, Spain
              \AND
       \name Jose A. Lozano \email jlozano@bcamath.org \\
       \addr        Intelligent Systems Group, University of the Basque Country UPV/EHU\\
       Basque Center for Applied Mathematics (BCAM)\\
      San Sebasti\'an, Spain,  Bilbao 48009, Spain}

\editor{Pierre Alquier}

\maketitle

\begin{abstract}%
Multiple supervised learning scenarios are composed by a sequence of classification tasks. For instance, multi-task learning and continual learning aim to learn a sequence of tasks that is either fixed or grows over time. 
Existing techniques for learning tasks that are in a sequence are tailored to specific scenarios, lacking adaptability to others. In addition, most of existing techniques consider situations in which the order of the tasks in the sequence is not relevant. However, it is common that tasks in a sequence are evolving in the sense that consecutive tasks often have a higher similarity. This paper presents a learning methodology that is applicable to multiple supervised learning scenarios and adapts to evolving tasks. Differently from existing techniques, we provide computable tight performance guarantees and analytically characterize the increase in the effective sample size. Experiments on benchmark datasets show the performance improvement of the proposed methodology in multiple scenarios and the reliability of the presented performance guarantees.
\end{abstract}


 \section{Introduction} \label{sec:intro}

\begin{figure}
\centering
                  \psfrag{t}[][][0.8][90]{Time}
                  \psfrag{5}[][][1]{$5$ years old}
                  \psfrag{25}[][][1]{$25$ years old}
                  \psfrag{50}[][][1]{$50$ years old}
                  \psfrag{75}[][][1]{$75$ years old}
                  \psfrag{b}[l][l][1]{ }
                  \psfrag{Sc}[l][l][1]{\textcolor{mypurple2}{SCD}}
                  \psfrag{CL}[l][l][1]{\textcolor{myred2}{CL}}
                   \psfrag{Online}[l][l][1]{{Online learning}}
                  \psfrag{MDA}[l][l][1]{\textcolor{mygreen2}{MDA}}
                  \psfrag{MTL}[l][l][1]{\textcolor{myblue2}{MTL}}
                  \psfrag{Batchabcdef}[l][l][1]{{Batch learning}}
                  \psfrag{Source}[l][l][1]{Source}
                  \psfrag{S}[l][l][1]{Source}
                  \psfrag{T}[l][l][1]{Target}
                  \psfrag{targetabc}[l][l][1]{and target}
         \includegraphics[width=\textwidth]{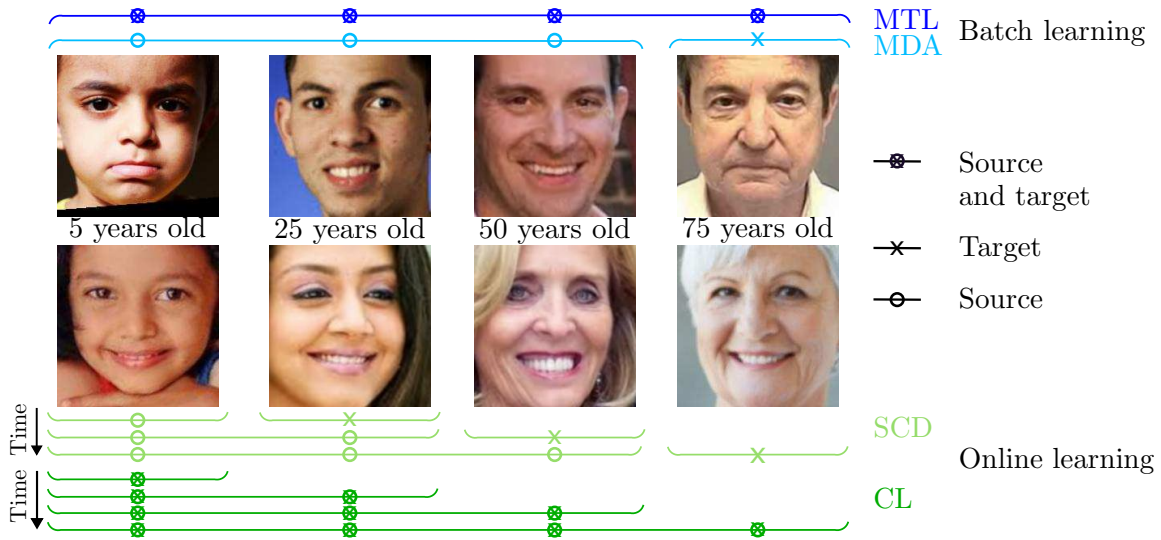}
\caption{Multiple supervised learning scenarios are composed by a sequence of tasks. Tasks are commonly evolving 
in the sense that consecutive tasks often have a higher similarity (e.g., gender classification in pictures of people with similar ages). The proposed methodology is applicable to multiple supervised learning scenarios and adapts to evolving tasks.}
\label{fig:intro}
\end{figure}

 There are multiple supervised learning scenarios composed by a sequence of tasks (classification problems). These scenarios mainly differ in the specific source tasks that provide information for learning, and the target tasks aimed to be learned (see Figure~\ref{fig:intro}). For instance, in \ac{MDA}, the goal is to learn a target task by leveraging information from all the tasks in the sequence (e.g., \citet{Mansour2009, Li2022, li2024subspace, yang2024dane}); while in \ac{MTL}, the goal is to learn simultaneously the whole sequence by leveraging information from all the tasks (e.g., \citet{Zhang2018, Lin2020, bengio2009curriculum, pentina2015curriculum, hu2024revisiting, knight2024multi}). Curriculum learning is a specific framework within the broader category of \ac{MTL} in which  tasks are ordered by difficulty, starting with simpler tasks and gradually progressing to more complex ones \citep{pentina2015curriculum, weinshall2018curriculum, bengio2009curriculum, bell2022effect}.

 In addition to the batch learning scenarios, there are also online learning scenarios where the sequence of tasks grows over time. For instance, in \ac{SCD}, the goal is to learn at each time step the last task in the sequence by leveraging information from the preceding tasks of the current sequence (e.g., \citet{elwell2011incremental, brzezinski2013reacting, guo2024dynamical, fedeli2023iwda}); while in \ac{CL}, the goal is to learn at each time step the current sequence by leveraging information from all the tasks (e.g., \citet{ruvolo2013ella, henning2021posterior, thapabayesian, wang2024comprehensive}). 

Learning tasks that are in a sequence holds promise to significantly improve performance by leveraging information from different tasks (\citealt{ruvolo2013ella, lopez2017gradient, chen2018lifelong, wang2024comprehensive}). Such transfer of information can enable accurate classification even in cases where each task has a reduced sample size (e.g., less than 100 samples), thus significantly increasing the \ac{ess} of each task. However, this transfer is hindered by the use of information from tasks characterized by different underlying distributions (e.g., negative transfer and catastrophic forgetting) (\citealt{henning2021posterior, kirkpatrick2017overcoming, hurtado2021optimizing, chen2024stability, evron2022catastrophic}).

\textbf{Related work:} Existing techniques for learning tasks that are in a sequence are tailored to one specific supervised learning scenario and are not suitable for others. For instance, methods designed for batch learning scenarios utilize samples from all the tasks in the sequence (\citealt{Evgeniou2004, Zhang2015, knight2024multi, hu2024revisiting}) and are not suitable for online learning scenarios. In addition, methods focused on one target task cannot effectively learn the whole sequence of tasks. Existing techniques designed for each supervised learning scenario follow different strategies. Common \ac{MDA} methods weight the source samples to minimize the distribution discrepancy between the source and target tasks (\citealt{Mansour2009, bai2022temporal, yang2024dane}). 
\ac{MTL} methods usually learn shared representations using samples from all the tasks \citep{Tripuraneni2020, zhang2021survey}.  Most of existing methods for \ac{SCD} learn the last task by slightly updating the classification rule from the preceding task \citep{elwell2011incremental, lu2019adaptive}. Methods for \ac{CL} often learn parameters using a pool of stored samples from all the preceding tasks together with the samples from the last task \citep{lopez2017gradient, henning2021posterior}.

Most of existing techniques consider situations in which the order of the tasks in the sequence is not relevant \citep{baxter2000model, maurer2016benefit, denevi2019online}. However, tasks in a sequence are commonly evolving in the sense that consecutive tasks often have a higher similarity \citep{bartlett1992learning, hanneke2019statistical, mazzetto2024adaptive, pentina2015lifelong}. For instance, in the problem of classifying face images of different ages \citep{zhifei2017cvpr}, the similarity between consecutive tasks (face images of consecutive ages) is significantly higher (see Figure~\ref{fig:intro}). The existing techniques that adapt to evolving tasks account for a scalar rate of change by using a learning rate \citep{orabona:2008, shen:2019}, 
 weight factor \citep{pavlidis:2011, bai2022temporal}, 
or window size (\citealt{bifet:2007, pentina2015lifelong, mazzetto2024adaptive}). Specifically, a slow/fast rate of change is tackled by using a low/high learning rate, weight factor, or smaller/larger window size. More sophisticated techniques designed for \ac{SCD} account for a time-varying scalar rate of change by using time-varying learning rates \citep{shen:2019}, weight factors \citep{pavlidis:2011}, or window sizes (\citealt{bifet:2007, xie2024evolving}). However, in common practical scenarios, the tasks' changes cannot be adequately addressed accounting only for a scalar rate of change. Such inadequacy is due to the fact that tasks' changes are commonly multidimensional, i.e., different statistical characteristics describing the tasks often change in a different manner.

Most of conventional techniques do not provide performance guarantees since tasks are characterized by different underlying distributions. Existing techniques that provide computable performance guarantees are not designed for evolving tasks \citep{amit2018meta, farid2021generalization}. On the other hand, existing techniques designed for evolving tasks provide performance guarantees in terms of distributions discrepancies that cannot be computed in practice \citep{long1999complexity, mohri2012new, pentina2015lifelong}. 
 
\textbf{Our contribution:} This paper presents a learning methodology that is applicable to multiple supervised learning scenarios and provides computable tight performance guarantees in terms of error probabilities. In addition, the proposed methodology adapts to evolving tasks accounting for multidimensional changes between consecutive tasks. Specifically, the main contributions in the paper are as follows.
 \begin{itemize}
\item We establish a learning methodology for evolving tasks based on \acp{MRC}. Such methodology is applicable to scenarios such as \ac{MDA}, \ac{SCD}, \ac{MTL}, and \ac{CL}.
\item We develop learning techniques that provide multidimensional adaptation to evolving tasks by estimating multiple statistical characteristics of the evolving underlying distributions.
\item We show that the proposed methodology can provide computable tight performance guarantees for evolving tasks and increase the \ac{ess} of each task using information from other tasks.
\item We numerically quantify the performance improvement of the proposed methodology in multiple scenarios with reduced sample sizes. In addition, we assess the reliability
of the performance guarantees presented.
 \end{itemize}
 
The rest of the paper is organized as follows. Section~\ref{sec:pre} briefly describes the problem formulation and \acp{MRC} that use linear combinations of general feature mappings. Section~\ref{sec:gf} presents the learning methodology for evolving tasks and describes the performance guarantees. Section~\ref{sec:evolution} proposes sequential techniques for efficient learning classification rules for all target tasks and Section~\ref{sec:ess} analytically characterizes the \ac{ess} increase. We describe the methods' implementation in scenarios with evolving tasks in Section~\ref{sec:scenarios}. Then, Section~\ref{sec:nr} provides multiple numerical results and Section~\ref{sec:conc} draws the conclusions.

This paper extends the conference papers~\citet{alvarez22, alvarez2023minimax} by establishing a general methodology with computable performance guarantees applicable to multiple supervised learning scenarios such as MDA, MTL, SCD, and CL. \citet{alvarez22} proposed techniques for SCD, while ~\citet{alvarez2023minimax} proposed techniques for CL. The additional results presented in this paper are a general learning methodology that is applicable to multiple machine learning scenarios. We present an extension to MDA, MTL, and CL of the results in \citet{alvarez22} for SCD that provide multidimensional adaptation to time changes with computable
performance guarantees. In addition, we present an extension to MDA, MTL, and SCD of the results in \citet{alvarez2023minimax} for CL that characterize the increase of the ESS. Furthermore, we assess the evolving
assumption and the multidimensional changes in real datasets and present several algorithmic extensions including techniques that account for high-order time dependences among tasks.

\section{Preliminaries} \label{sec:pre}

This section describes the main notations used in the paper and multiple supervised learning scenarios composed by a sequence of tasks. Then, we briefly describe \ac{MRC} methods that minimize the worst-case error probability over an uncertainty set. For the readers’ convenience, we provide in Table~\ref{notions} a list with the main notions used in the paper and their corresponding notations.

\begin{table}[ht]
\caption{Main notations used in the paper}
\label{notions}
\centering
\renewcommand{\arraystretch}{1.3}
\adjustbox{max width=\textwidth}{%
\begin{tabular}{ll}
\hline
Notation &Meaning\\\hline\hline
$\Delta(\set{X}\times\set{Y})$& set of probability distributions over instances $\set{X}$ and labels $\set{Y}$\\
$ \text{T}(\set{X},\set{Y})$& set of classification rules from instances $\set{X}$ to labels $\set{Y}$\\
$\up{h}(y|x)$&probability assigned by classification rule $\up{h}$ to label $y$ for instance $x$\\
$\ell(\up{h},\up{p})$& expected 0-1 loss of classification rule $\up{h}$ w.r.t. distribution $\up{p}$\\
$R(\up{h})$& risk (error probability) of classification rule $\up{h}$\\
$\Phi:\set{X}\times\set{Y}\to\mathbb{R}^m$&feature mapping with dimension $m$\\
$M$& bound of the feature mapping ($\|\Phi(x, y)\|_\infty \leq M$)\\
$\set{U}$&uncertainty set of distributions given by expectations' constraints as in \eqref{eq:us}\\
$\up{p}_j$ & underlying distribution that characterizes the $j$-th task\\
$D_j$& sample set of the $j$-th task\\
$n_j$& sample size of $D_j$\\
$\Phi_j$& random variable given by the feature mapping of samples from the $j$-th task\\
$\B{\sigma}_j$& estimate of standard deviation of the $j$-th task\\
$\B{\tau}_j^\infty$ &expectation of feature mapping corresponding to the $j$-th task\\
$\B{w}_j = \B{\tau}_j^\infty - \B{\tau}_{j-1}^\infty$& change between tasks\\
$\B{d}_j$& estimate of the expected quadratic change between the $j$-th task and the $j-1$-th task\\
$\set{U}_j^\infty$ & uncertainty set of the $j$-th task given by the true expectation\\
$R_j^\infty$&  minimum worst-case error probability of the $j$-th task\\
$\B{\mu}_j^\infty$& classifier parameter of the $j$-th task corresponding with $\set{U}_j^\infty$\\
$\B{\mu}_j$& classifier parameter of the $j$-th task\\
$\up{h}_j$ & classification rule using single-task learning\\
$\up{h}_j^j$ & classification rule using forward learning\\
$\up{h}_j^k$ & classification rule using forward and backward learning in a sequence of $k$ tasks\\
$\set{U}_j$, $\set{U}_j^j$, $\set{U}_j^k$&uncertainty sets using single-task, forward, and forward and backward learning\\
$R(\set{U}_j)$, $R(\set{U}_j^j)$, $R(\set{U}_j^k)$&minimax risks using single-task, forward, and forward and backward learning\\
$\B{\tau}_j$, $\B{\tau}_j^j$, $\B{\tau}_j^k$&mean vector using single-task, forward, and forward and backward learning\\
$\B{\lambda}_j$, $\B{\lambda}_j^j$, $\B{\lambda}_j^k$&confidence vectors using single-task, forward, and forward and backward learning\\
$\B{s}_j$, $\B{s}_j^j$, $\B{s}_j^k$&MSE vectors using single-task, forward, and forward and backward learning\\
$\B{\eta}_j^j, \B{\eta}_j^k$ & smoothing gains using forward, and forward and backward learning\\
$n_j^j$, $n_j^k$&ESSs using forward, and forward and backward learning\\
\hline
\end{tabular}}
\end{table}

\paragraph{Notation} Calligraphic letters represent sets; bold lowercase letters represent vectors; bold uppercase letters represent matrices; $\V{I}$ and $\mathbb{I}\{\cdot\}$ denote the identity matrix and the indicator function, respectively; $(\cdot)_+$ denotes the positive part of its argument; $\text{sign}(\cdot)$  denotes the vector given by the signs of the argument components; $\V{e}_i$ denotes the $i$-th vector in a standard basis, $\| \cdot \|_1$ and $\| \cdot\|_{\infty}$ denote the $1$-norm and the infinity norm of its argument, respectively; $\preceq$ and $\succeq$ denote vector inequalities; and $\mathbb{E}_{\up{p}}\{\,\cdot\,\}$ and $ \mathbb{V}\text{ar}_{\up{p}}\{\cdot\}$ denote the component-wise expectation and the component-wise variance of its argument with respect to distribution $\up{p}$. For a vector $\V{v}$, $v^{(i)}$ and $\V{v}^\top$ denote the $i$-th component and the transpose of $\V{v}$, respectively. In addition,  non-linear operators acting on vectors denote component-wise operations. For instance, $|\V{v}|$ and $\V{v}^2$ denote the vector formed by the absolute value and the square of each component, respectively, and for a vector $\V{u}$, $\V{v}/\V{u}$ denotes the vector formed by the component-wise division.

\subsection{Problem Formulation}\label{sec:pf}

Let $\set{X}$ be a set of instances and $\set{Y}$ be a set of labels with $\set{Y}$ represented by $\{1, 2, \ldots, |\set{Y}|\}$, we denote by $\Delta(\set{X}\times\set{Y})$ the set of probability distributions over \mbox{$\set{X}\times\set{Y}$}. In addition, we denote by $\text{T}(\set{X},\set{Y})$ the set of classification rules and by $\up{h}(y|x)$ the probability with which classification rule $\up{h}\in\text{T}(\set{X}, \set{Y})$ assigns label $y \in \set{Y}$ to instance $x \in \set{X}$ ($\up{h}(y|x) \in \{0, 1\}$ for deterministic classification rules). Supervised classification techniques use a sample set $D=\{(x_i,y_i)\}_{i=1}^n$ formed by $n$ i.i.d. samples from distribution $\up{p}^*\in\Delta(\set{X}\times\set{Y})$ to find a classification rule $\up{h}\in \text{T}(\set{X}, \set{Y})$ with small expected loss $\ell(\up{h}, \up{p}^*)$. In the following, we utilize the $0$-$1$-loss so that the expected loss of the classification rule $\up{h}$ with respect to the underlying distribution is the error probability of such rule $\up{h}$ that is denoted by $R(\up{h})$. 

In the addressed setting, there are multiple sample sets $D_1, D_2, \ldots$ where each sample set $D_j$ is composed by $n_j$ i.i.d. instance-label pairs. Sample sets $D_1, D_2, \ldots$ correspond to different classification tasks characterized by underlying distributions $\up{p}_1, \up{p}_2, \ldots$. Learning methods use the available sample sets to obtain classification rules for the target tasks. Such setting describes multiple supervised learning scenarios that mainly differ in the specific target tasks they aim to learn and the source tasks from which they acquire information for learning (see Fig.~\ref{fig:intro}). For instance, \ac{MDA} methods use sample sets $D_1, D_2, \ldots, D_{k-1}$ from $k-1$ source tasks to obtain a classification rule $\up{h}_k$ for the $k$-th target task; while \ac{MTL} methods use $k$ sample sets $D_1, D_2, \ldots, D_k$ to obtain classification rules $\up{h}_{1}, \up{h}_{2}, \ldots, \up{h}_{k}$ so all the tasks in the sequence are both source and target. In addition to these batch learning scenarios, the above setting also describes online learning scenarios where sample sets $D_1, D_2, \ldots$ corresponding with different tasks arrive over time. For instance, \ac{SCD} methods use, at each step $k$, sample sets $D_1, D_2, \ldots, D_{k-1}$ from $k-1$ source tasks to obtain a classification rule $\up{h}_k$ for the $k$-th target task; while \ac{CL} methods use, at each step $k$, sample sets $D_1, D_2, \ldots, D_k$ to obtain classification rules $\up{h}_{1}, \up{h}_{2}, \ldots, \up{h}_{k}$ so all the tasks in the current sequence are both source and target. In this paper, we establish a general methodology for learning tasks in a sequence that can be used in all the above mentioned scenarios. 

Most existing techniques are designed for situations where tasks’ similarities do not depend on the order of the task in the sequence \citep{baxter2000model, maurer2016benefit, denevi2019online}. These settings are usually mathematically modeled assuming that the tasks' distributions
\begin{equation}
\label{eq:iid}
\tag{i.i.d.-A} \up{p}_1, \up{p}_{2}, \ldots \text{ are independent and identically distributed.}
\end{equation}
In particular, each distribution $\up{p}_{j}$ for $ j = 1, 2, \ldots$ is sampled from a fixed task meta-distribution \citep{baxter2000model, maurer2016benefit, denevi2019online}. In this paper, we develop techniques designed for evolving tasks where consecutive tasks are significantly more similar. These settings can be mathematically modeled assuming that the tasks' distributions \begin{equation}
\label{eq:td}
\tag{Evo-A}\up{p}_1, \up{p}_{2}, \ldots \text{ form a random walk with independent and zero-mean increments.}\end{equation} 
In this case, each distribution $\up{p}_{j}$  for $ j = 1, 2, \ldots$ is sampled from a task meta-distribution that is continuously evolving over tasks. Under the~\eqref{eq:td} assumption, we have that $\up{p}_{j+1} = \up{p}_j + \varepsilon_{j+1}$ where \mbox{$\varepsilon_{j+1}$} for $j = 1, 2, \ldots$ are independent and zero-mean. The~\eqref{eq:td} assumption is similar to the ``changing task environments'' assumption used in Pentina and Lampert (2015) and the distribution change between consecutive tasks in~\citet{bartlett1992learning, hanneke2019statistical}. However, these papers assume that the distribution slightly changes between consecutive tasks, while we assume that the distribution evolves as a random walk with independent increments.  Section~\ref{sec:gf} below further shows how the assumption used in the paper can describe common real datasets better than the conventional~\eqref{eq:iid} assumption.

\subsection{Minimax risk classifiers}\label{sec:mrc}

\acp{MRC} are based on robust risk minimiztion (RRM) also known as distributionally robust learning  \citep{farnia2016minimax, FatAnq:16}. RRM-based methods differ in the uncertainty set considered. For instance, \citet{shafieezadeh2019regularization} utilize uncertainty sets determined by  Wasserstein distances and \cite{duchi2019variance} utilize uncertainty sets determined by f-divergences. However, these uncertainty sets are not easy to obtain in scenarios with varying distributions. \acp{MRC} \citep{mazuelas:2020, MazRomGru:22} determine uncertainty sets using expectation estimates that can be effectively obtained in the addressed settings. In particular, these uncertainty sets are especially suitable for learning from a sequence of tasks because they can effectively leverage samples from multiple tasks characterized by different distributions.  

The uncertainty set used in the proposed methods is given by constraints on the expectation of a feature mapping $\Phi: \set{X} \times \set{Y} \rightarrow \mathbb{R}^m$ as
\begin{equation}
\label{eq:us}
\mathcal{U} = \{\up{p} \in \Delta (\mathcal{X} \times \mathcal{Y}) : \left|\mathbb{E}_{\up{p}}\{\Phi(x, y)\}  - {\B{\tau}} \right| \preceq \B{\lambda}\}
 \end{equation}
where $|\, \cdot \, |$ denotes the vector formed by the absolute value of each component in the argument, ${\B{\tau}}$ denotes the vector of expectation estimates corresponding with the feature mapping~$\Phi$, and $\B{\lambda} \succeq \V{0}$ is a confidence vector that accounts for inaccuracies in the estimate. Feature mappings are vector-valued functions over $\set{X}\times\set{Y}$. For instance, such  mappings can be defined by multiple features over instances together with a one-hot encoding of labels, as follows \citep{mohri:2018}
\begin{equation}
\label{eq:feature_mapping}
\Phi(x, y) = \V{e}_y \otimes \Psi(x) = [\mathbb{I}\{y = 1\}\Psi(x)^\top, \mathbb{I}\{y = 2\}\Psi(x)^\top, ..., \mathbb{I}\{y = |\set{Y}|\}\Psi(x)^\top]^\top
\end{equation}
where $\V{e}_y$ is the $y$-th vector in the standard basis of $\mathbb{R}^{|\set{Y}|}$, $\otimes$ denotes the Kronecker product, and the map $\Psi:\set{X} \rightarrow \mathbb{R}^q$ represents instances as real vectors.  The feature mapping $\Phi$ represents each instance-label pair $(x, y)$ by an $m$-dimensional real vector with $m = |\set{Y}|q$, so that $\Phi(x, y)$ is composed by $|\set{Y}|$ $q$-dimensional blocks with values $\Psi(x)$ in the block corresponding to $y$ and zero otherwise. General types of scalar features can be used for MRC learning including those given by thresholds/decision stumps \citep{lebanon2001boosting}, the last layer in a neural network (NN) \citep{bengio2013representation}, and random features corresponding to a reproducing kernel Hilbert space \citep{rahimi2007random}.  
 
Given the uncertainty set $\set{U}$, \ac{MRC} rules minimize the worst-case error probability and are solutions of the optimization problem
 \begin{equation}
 \label{eq:minmaxrisk}
\underset{\up{h} \in \text{T}(\mathcal{X}, \mathcal{Y})}{\min} \, \underset{\up{p} \in \mathcal{U}}{\max} \; \ell(\up{h}, \up{p})
\end{equation}
where $\ell(\up{h}, \up{p})$ denotes the expected loss of classification rule $\up{h}$ for distribution $\up{p}$. In the following, we utilize the 0-1-loss so that $\ell(\up{h}, \up{p})= \mathbb{E}_{\up{p}}\{1 - \up{h}(y|x)\}$  and the expected loss with respect to the underlying distribution becomes the error probability of the classification rule. In addition, the optimal value of~\eqref{eq:minmaxrisk} is denoted by $R(\set{U})$ and referred to as the minimax risk for uncertainty set $\set{U}$.

The \ac{MRC} rule $\up{h}$ assigns label $\hat{y} \in \mathcal{Y}$ to instance $x \in \mathcal{X}$ with probability given by linear-affine combinations of the components of the feature mapping as (see eq. (11) in~\citet{MazRomGru:22})
\begin{equation}
\label{eq:prob}
\up{h}(\hat{y}|x) = \left\{\begin{matrix}\left(\Phi(x, \hat{y})^{\top} \B{\mu}^*  -  \varphi(\B{\mu}^*) \right)_+ /c_{x} & \text{if} \, c_{x} \neq 0\\
1/|\mathcal{Y}| & \text{if} \, c_{x} = 0 \end{matrix}\right.
\end{equation}
with
\begin{align*}
\varphi(\B{\mu}^*) &= {\underset{x \in \mathcal{X}, \set{C} \subseteq \mathcal{Y}}{\max}} \Big{(}\sum_{y \in \set{C}}\Phi(x, y)^{\top}\B{\mu}^* - 1\Big{)}/|\set{C}|, \;\; c_{x}=\sum_{y \in \mathcal{Y}} \left(\Phi(x, y)^{\top} \B{\mu}^* - \varphi(\B{\mu}^*) \right)_+.
\end{align*}
The vector parameter $\B{\mu}^*$ is the solution of the convex optimization problem (see eq. (6) in~\citet{MazRomGru:22})
\begin{equation}
\label{eq:mrc}
\underset{\B{\mu}}{\min} \; 1 - {\B{\tau}}^{\top} \B{\mu} + \varphi(\B{\mu}) + \B{\lambda}^{\top} \left|\B{\mu}\right|
\end{equation}
given by the Fenchel-Lagrange dual of~\eqref{eq:minmaxrisk} \citep{mazuelas:2020, mazuelas:2022};  and parameters $\B{\mu}^*$ correspond to the Lagrange multipliers of constraints in~\eqref{eq:us}. Note that the label that maximizes the probability in~\eqref{eq:prob} is given by $\hat{y} \in \arg \max_{y \in \set{Y}} {\Phi}(x,y)^{\top} \B{\mu}^*$. The deterministic classification rule $\up{h}^\up{d}$ that assigns such label $\hat{y}$ to instance $x$ will be referred in the following as deterministic MRC.  \acp{MRC} are specifically designed for classification problems. The extension of MRCs to regression problems is not straightforward because MRCs do not consider a parametric space of possible rules. 

The baseline approach of single-task learning obtains a classification rule $\up{h}_j$ for each \mbox{$j$-th} task leveraging information only from the sample set $\mbox{$D_j = \{(x_{j, i}, y_{j, i})\}_{i= 1}^{n_j}$}$ given by $n_j$ i.i.d. samples from distribution $\up{p}_j$.  For single-task learning, \acp{MRC} can obtain mean and confidence vectors as \citep{mazuelas:2020}
\begin{equation}
\label{eq:tau1}
\B{\tau}_j = \frac{1}{n_j} \sum_{i = 1}^{n_j} \Phi\left(x_{j, i}, y_{j, i}\right), \; \hspace{0.2cm} \B{s}_j = \frac{\B{\sigma}_j^2}{n_j},\; \hspace{0.2cm} \B{\lambda}_j =   \lambda_0 \sqrt{\B{s}_j} \; \; \in \; \; \mathbb{R}^m
\end{equation}
where  $\lambda_0 >0$ controls the regularization strength in the methods proposed. The mean vector $\B{\tau}_j$ corresponds to the sample mean and provides an unbiased estimator of the feature mapping expectation $\B{\tau}_j^\infty = \mathbb{E}_{\up{p}_j}\{\Phi(x, y)\}$. The vector $\B{\sigma}_j^2$ is given by an estimate of the feature mapping variance.  
Specifically, the $i$-th component of vector $\B{\sigma}_j^2$ is an estimate of the variance of the $i$-th component of the feature mapping, denoted by $\mathbb{V}\text{ar}_{\up{p}_j}\{\Phi^{(i)}(x,y)\}$. In particular, if $\B{\sigma}_j^2$ is given by the sample variance the vector $\B{s}_j$ in~\eqref{eq:tau1} corresponds to the \acp{MSE} of the mean vector $\B{\tau}_j$.

In the following sections, we describe techniques that obtain the mean and \ac{MSE} vectors in scenarios that are composed by a sequence of tasks. Once such vectors are obtained, the proposed methodology obtains the classifier parameter $\B{\mu}_{j}$ for each $j$-th task solving the convex optimization problem in~\eqref{eq:mrc} that can be efficiently addressed using conventional methods~\citep{nesterov2015quasi, tao:2019}.

\section{Learning evolving tasks}\label{sec:gf}
  
This section assesses the evolving assumption in Section~\ref{sec:pre} and presents the methodology for learning evolving tasks. 

\subsection{Evolving tasks}\label{sec:time_dependent_tasks}

In the following, we describe the main aspects of evolving tasks and the inadequacy of conventional modeling assumptions. In particular, we analyze the similarities among underlying distributions $\up{p}_j$, for $j = 1, 2, \ldots$, by assessing the similarities among the corresponding vectors formed by the expectations of a feature mapping. For each $j$-th task, such a mean vector $\B{\tau}_j^\infty = \mathbb{E}_{\up{p}_j}\{\Phi(x, y)\}$ represents the statistical characteristics of the underlying distribution $\up{p}_j$ as measured by the feature mapping  $\Phi: \set{X} \times \set{Y} \rightarrow \mathbb{R}^m$. 

\begin{figure}
\centering
\begin{subfigure}[t]{0.48\textwidth}
     \psfrag{Airlines datasetabcdefghijklmnopq}[l][l][0.7]{``Airlines'' dataset}
                  \psfrag{Faces}[l][l][0.7]{``UTKFaces'' dataset}
                  \psfrag{Lags}[][][0.7]{Lags}
                  \psfrag{Partial autocorrelation}[][][0.7]{Partial autocorrelation}
                  \psfrag{1}[][][0.5]{$1$}
                  \psfrag{4}[][][0.5]{$4$}
                  \psfrag{8}[][][0.5]{$8$}
                  \psfrag{12}[][][0.5]{$12$}
                  \psfrag{16}[][][0.5]{$16$}
                  \psfrag{20}[][][0.5]{$20$}
                  \psfrag{-0.2}[][][0.5]{$-0.2$}
                  \psfrag{0}[][][0.5]{$0$}
                  \psfrag{0.2}[][][0.5]{$0.2$}
                  \psfrag{0.6}[][][0.5]{$0.6$}
         \includegraphics[width=\textwidth]{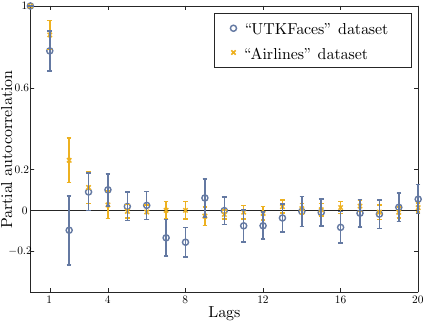}
\caption{Averaged partial autocorrelation of mean vectors components +/- their standard deviations shows the better adequacy of the~\eqref{eq:td} assumption for evolving tasks.}
\label{fig:partial_autocorrelation}
\end{subfigure}
\hfill
\begin{subfigure}[t]{0.48\textwidth}
     \psfrag{Task}[][][0.7]{Task}
     \psfrag{data1abcdefghijklmnijklmnop}[l][l][0.7]{Component 1 $y = 1$}
     \psfrag{data2}[l][l][0.7]{Component 5 $y = 1$}
     \psfrag{data3}[l][l][0.7]{Component 1 $y = 2$}
                  \psfrag{Mean vector}[][][0.7]{Mean vector components}
                  \psfrag{Lags}[l][l][0.7]{Lags}
                  \psfrag{Partial Autocorrelation}[l][l][0.7]{Partial autocorrelation}
                  \psfrag{0.4}[][][0.5]{$0.4$}
                  \psfrag{0.8}[][][0.5]{$0.8$}
                  \psfrag{8}[][][0.5]{$8$}
                  \psfrag{1.2}[][][0.5]{$1.2$}
                  \psfrag{16}[][][0.5]{$16$}
                  \psfrag{20}[][][0.5]{$20$}
                  \psfrag{-0.2}[][][0.5]{$-0.2$}
                  \psfrag{0}[][][0.5]{$0$}
                  \psfrag{10}[][][0.5]{$10$}
                  \psfrag{200}[][][0.5]{$200$}
                  \psfrag{400}[][][0.5]{$400$}
                  \psfrag{600}[][][0.5]{$600$}
                  \psfrag{-3}[][][0.5]{$-3$}
         \includegraphics[width=\textwidth]{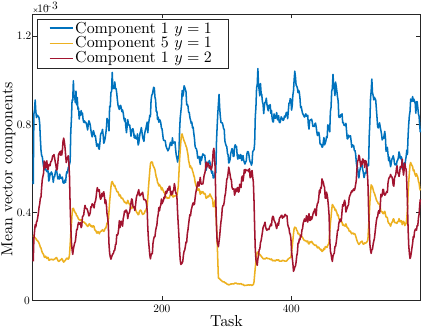}
\caption{Mean vector components using ``Airlines'' dataset shows the multidimensional changes in tasks' distributions.}
\label{fig:tau_components}
\end{subfigure}
\caption{In common practical scenarios, tasks are evolving and tasks' changes are multidimensional.}
\label{fig:evo}
\end{figure} 

For a sequence of evolving tasks, the similarity between two tasks depends on their distance in the sequence. In particular, the similarity between $\up{p}_{j+t}$ and $\up{p}_{j}$ is often higher for small $t$, and consecutive tasks  ($t = 1$) are significantly more similar. This similarity between consecutive tasks can be better described by the~\eqref{eq:td} assumption in Section~\ref{sec:pf} instead of the usual~\eqref{eq:iid} assumption. The~\eqref{eq:td} assumption considers independent changes between consecutive distributions, while the~\eqref{eq:iid} assumption considers independent distributions. 
 Note that if the tasks' distributions satisfy~\eqref{eq:td}, the difference between the distributions of the $j$-th and the $(j+t)$-th tasks has zero-mean and variance \mbox{$\mathbb{V}\text{ar}\{\up{p}_{j+t}-\up{p}_j\}=\sum_{i=1}^t \mathbb{V}\text{ar}\{\up{p}_{j+i}-\up{p}_{j+i-1}\}$} that increases with~$t$. On the other hand, if the tasks' distributions satisfy~\eqref{eq:iid}, the difference between the distributions of the $j$-th and the $(j+t)$-th tasks has zero-mean and variance \mbox{$\mathbb{V}\text{ar}\{\up{p}_{j+t}-\up{p}_j\}=\mathbb{V}\text{ar}\{\up{p}_{j+1}-\up{p}_j\} = 2\mathbb{V}\text{ar}\{\up{p}_1\}$} that does not depend on $t$. 
 
The better adequacy of the~\eqref{eq:td} assumption for evolving tasks can be assessed in real datasets by using partial autocorrelations, as shown in Figure~\ref{fig:partial_autocorrelation}. This figure shows the averaged partial autocorrelation of the mean vector components +/- their standard deviations for different lags using “Airlines” and “UTKFaces” datasets (see dataset characteristics in Table~\ref{tab:datasets} of Appendix~\ref{app:res}). The  ``Airlines'' dataset is composed by tasks corresponding to airplane delays of a specific time period and the ``UTKFaces'' dataset is composed by tasks corresponding to the classification of face images of a specific age (see also Fig.~\ref{fig:intro}). The partial autocorrelation is a usual measure of the correlation between observations of a process (see Section 4 in \citet{cowpertwait2009introductory}). For instance, the partial autocorrelation indicates that a process is a random walk if there are significant correlations at the first lag, followed by much smaller correlations; while the partial autocorrelation indicates that the observations are i.i.d. if the correlations are not significant at any lag. Figure~\ref{fig:partial_autocorrelation} shows that the partial autocorrelation of the sequence of mean vectors are clearly larger than zero at lag $1$. Hence, the figure shows that the~\eqref{eq:td} assumption better describes the tasks' distributions than the~\eqref{eq:iid} assumption in real datasets. 

Figure~\ref{fig:partial_autocorrelation} also shows that higher-order dependences among tasks can be adequate for certain datasets. For instance, the figure shows that the partial autocorrelations of the mean vector components are larger than zero at lag 2 for ``Airlines'' dataset. Accounting for higher-order dependences can better capture the similarities among tasks in certain datasets. Most of the results in the paper are described for the simple case that accounts for first order similarities between consecutive tasks via the above discussed~\eqref{eq:td} assumption. Section~\ref{subsec:time_increment} shows extensions of the proposed methodology accounting for higher-order dependences.

Figure~\ref{fig:tau_components} shows that the changes in tasks' distributions are often multidimensional, that is, different statistical characteristics of the underlying distribution often change in a different manner. This figure shows 3 mean vector components (component $1$ for $y = 1$, component $5$ for $y = 1$, and component $1$ for $y = 2$) for different tasks using ``Airlines'' dataset. Such figure illustrates a clearly different change in each mean vector component that reflects multidimensional tasks' changes. For instance, the expected values of the first instance component for class $y=2$ (component 1 $y=2$) often increases in cases where other components decrease. In addition, the expected values of the fifth component for class $y = 1$ (component 5 $y = 1$) exhibit slower changes than those for component 1 with \( y = 1 \). 
{Existing methods for evolving tasks account for a scalar rate of change that cannot capture the multidimensional tasks' changes. In the following, we propose techniques that account for the change in each component in the mean vector by using a vector $\B{d}_j$ that assesses the expected quadratic change between consecutive tasks.}

\subsection{Learning methodology}

    \begin{figure}
         \centering
     \psfrag{7}[][][0.9]{$\set{U}_{j-1}^k$}
     \psfrag{1}[][][0.9]{$\set{U}_k^k$}
     \psfrag{2}[][][0.9]{$\set{U}_{j+1}^k$}
          \psfrag{3}[][][0.9]{$\set{U}_{1}^k$}
          \psfrag{t}[][][0.9]{$\B{\lambda}_{j-1}$}
     \psfrag{l}[][][0.9]{$\B{\lambda}_j$}
     \psfrag{k}[][][0.9]{$R(\set{U}_{2}^k)$}
     \psfrag{f}[][][0.9]{$R(\set{U}_j^k)$}
     \psfrag{5}[][][0.9]{$R(\set{U}_{j+1}^k)$}
     \psfrag{z}[][][0.9]{$\set{U}_2^k$}
     \psfrag{0}[][][0.9]{$\set{U}_j^k$}
      \psfrag{8}[][][0.9]{$\up{h}_{j-1}^k$}
      \psfrag{d}[][][0.9]{$\up{h}_{j}^k$}
      \psfrag{h}[][][0.9]{$\up{h}_k^k$}
        \psfrag{.}[][][0.9]{$\ldots$}
      \psfrag{2}[][][0.9]{$\set{U}_{j+1}^k$}
            \psfrag{s}[][][0.9]{$D_k$}
                  \psfrag{i}[][][0.9]{$\up{h}_{1}^k$}
      \psfrag{x}[l][][1]{Sample sets}
     \psfrag{r}[][][0.9]{$D_1$}
     \psfrag{9}[][][0.9]{$R(\set{U}_{j-1}^k)$}
     \psfrag{6}[][][0.9]{$R(\set{U}_k^k)$}
     \psfrag{4}[][][0.9]{$\up{h}_{j+1}^k$}
     \psfrag{o}[l][][1]{Classification rules}
       \psfrag{n}[l][][1]{Uncertainty sets}
            \psfrag{p}[l][][1]{Performance}
       \psfrag{g}[l][][1]{guarantees}
              \psfrag{b}[l][][0.9]{}
         \psfrag{a}[][][0.9]{$D_{j}$}
          \psfrag{e}[][][0.9]{$D_{j+1}$}
          \psfrag{c}[][][0.9]{$D_{j-1}$}
          \psfrag{q}[][][0.9]{$\B{d}_j$}
           \psfrag{y}[][][0.9]{$\B{d}_{j+1}$}
           \psfrag{u}[l][][1]{}
           \psfrag{t}[l][][1]{}
           \psfrag{l}[][][0.9]{$R(\set{U}_1^k)$}
     \includegraphics[width=0.98\textwidth]{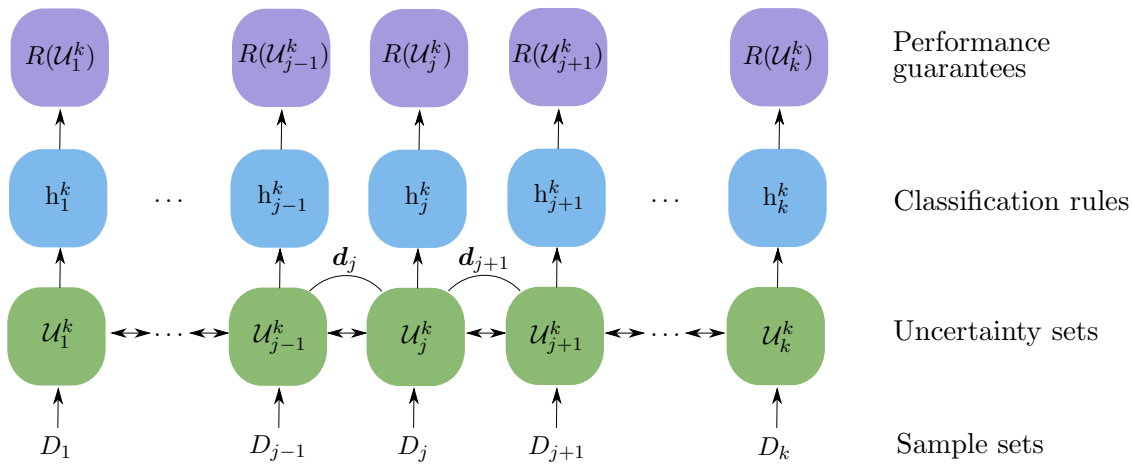}
    \caption{The proposed methodology obtains an uncertainty set $\set{U}_j^k$ for each $j$-th task using the sample set $D_j$, the adjacent uncertainty sets $\set{U}_{j-1}^k, \set{U}_{j+1}^k$, and the change between consecutive tasks $\B{d}_j, \B{d}_{j+1}$. Then, the uncertainty set is used to obtain the classification rule $\up{h}_j^k$ together with the minimax risk $R(\set{U}_j^k)$ that directly gives performance guarantees.}
       \label{fig:diagrama_general}
    \end{figure}  

Figure~\ref{fig:diagrama_general} depicts the flow diagram of the proposed learning methodology that is applicable to multiple supervised learning scenarios and provides computable tight performance guarantees.

For a sequence of $k$ tasks, the proposed learning methodology obtains an MRC rule for each $j$-th task leveraging information from the $k$ tasks and accounting for the change between consecutive tasks. Specifically, for each $j$-th task, the proposed methods obtain an uncertainty set $\set{U}_j^k$ as in~\eqref{eq:us} by using uncertainty sets, $\set{U}_{j-1}^k$, $\set{U}_{j+1}^k$, for adjacent tasks and the expected quadratic change between consecutive tasks $\B{d}_j$, $\B{d}_{j+1}$ (see Fig.~\ref{fig:diagrama_general}). Then, given the uncertainty set $\set{U}_j^k$, we obtain the classification rule $\up{h}_j^k$ and the minimax risk $R(\set{U}_j^k)$ solving the convex optimization problem~\eqref{eq:mrc}.  For instance, in online learning scenarios such as SCD, at each step $k$ an uncertainty set $\set{U}_k^k$ for each $k$-th task can be obtained by using $\set{U}_{k-1}^{k-1}$ and $\B{d}_{k}$; then such uncertainty set is used to obtain the classification rule $\up{h}_k^k$ and the minimax risk $R(\set{U}_k^k)$.  Section~\ref{sec:evolution} describes techniques that recursively obtain uncertainty sets for each task accounting for multidimensional tasks' changes and Section~\ref{sec:scenarios} describes techniques to solve the convex optimization problem in~\eqref{eq:mrc} based on subgradient methods \citep{nesterov2015quasi, tao:2019}. 

The proposed methodology provides computable tight performance guarantees for the error probability of each $j$-th task by assessing the minimax risks. Let $R(\up{h}_j^k)$ denote the error probability of the classification rule $\up{h}_j^k$ that is determined by the parameter $\B{\mu}_j^k$. Then, using the bounds of Theorem 7 in \citet{mazuelas:2022} in the addressed setting, we have that 
\begin{align}
 \label{eq:prob_error_continual}
R(\up{h}_j^{k}) & \leq R(\mathcal{U}_j^{k}) + \left(|\B{\tau}_j^\infty - \B{\tau}_j^k| - \B{\lambda}_j^k\right)^{\top} |\B{\mu}_j^{k}| 
\end{align}
where $R(\set{U}_j^k)$ and $\set{U}_j^k$ denote the minimax risk and the uncertainty set given as in~\eqref{eq:us} by mean and confidence vectors $\B{\tau}_j^k$ and $\B{\lambda}_j^k$. The above inequality provides computable tight bounds for error probabilities given by the minimax risk. Specifically, the minimax risk $R(\set{U}_j^k)$ directly provides a bound if the error in the mean vector estimate satisfies \mbox{$|\B{\tau}_j^\infty - \B{\tau}_{j}^{k}| \preceq \B{\lambda}_j^{k}$}. In other cases, the minimax risk $R(\set{U}_j^k)$ still provides approximate bounds as long as the difference \mbox{$|\B{\tau}_j^\infty - \B{\tau}_j^k| - \B{\lambda}_j^k$} is not substantial. Bound in~\eqref{eq:prob_error_continual} is tight in the sense that it would coincide with the \ac{MRC} error if the underlying distribution is the worst-case distribution in $\set{U}_j^k$ (the distribution in $\set{U}_j^k$ with the largest Bayes risk). Section~\ref{sec:nr} shows that the presented bounds can also provide tight performance guarantees in practice.

The learning methodology is applicable to multiple supervised learning scenarios including online (\ac{SCD} and \ac{CL}) and batch learning scenarios (\ac{MDA} and \ac{MTL}). The proposed techniques effectively use information from all the tasks in the sequence to obtain a classification rule for the last task in \ac{SCD} and \ac{MDA} or a sequence of classification rules in \ac{MTL} and \ac{CL}. Section~\ref{sec:scenarios} describes the implementation of the proposed methodology in multiple scenarios and describes techniques accounting for higher-order dependences among tasks. 

\section{Determining uncertainty sets from sample sets of evolving tasks}\label{sec:evolution}

This section presents techniques to recursively obtain uncertainty sets as in~\eqref{eq:us} accounting for the evolution of the tasks' distribution and leveraging information from the tasks in the sequence. Specifically, the proposed techniques recursively obtain mean and \ac{MSE} vectors for each $j$-th task leveraging information from preceding tasks (forward learning) and, reciprocally, obtain mean and \ac{MSE} for preceding tasks leveraging information from the succeeding tasks in the sequence (backward learning). 

\subsection{Forward learning}

This section presents the recursions that allow us to leverage information from preceding tasks. For a sequence of $k$ tasks, the proposed techniques first obtain for each $j$-th task with \mbox{$j = \{1, 2, \ldots, k\}$} forward mean and \ac{MSE} vectors leveraging information up to the $j$-th task. Let $\B{\tau}_j^{j}$ and $\B{s}_j^{j}$ denote the mean and MSE vectors for forward learning corresponding to the $j$-th task. The following recursions allow us to obtain $\B{\tau}_j^{j}, \B{s}_j^{j} \in \mathbb{R}^m$ for each $j$-th task using those for the preceding task $\B{\tau}_{j-1}^{j-1}, \B{s}_{j-1}^{j-1}$ together with a smoothing gain $\B{\eta}_j^j$ as follows
\begin{alignat}{5}
\label{eq:tau_general}
\B{\tau}_j^{j} & = \B{\tau}_{j-1}^{j-1} + \B{\eta}_j^{j} \left({\B{\tau}}_j - \B{\tau}_{j-1}^{j-1}\right)\\
\label{eq:s_general}
\B{s}_j^{j} & = \B{\eta}_j^{j} {\B{s}_j}\\
\label{eq:gain_general}
\B{\eta}_j^{j} & =  \frac{\B{s}_{j-1}^{j-1}+ \B{d}_j}{\B{s}_j + \B{s}_{j-1}^{j-1}+ \B{d}_j}
\end{alignat}
where non-linear operators acting on vectors of the same dimension denote component-wise operations, the vectors t$\B{\tau}_j, \B{s}_j$ are given by~\eqref{eq:tau1} and $\B{\tau}_1^1 = \B{\tau}_1$, $\B{s}_1^1 = \B{s}_1$. Mean and \ac{MSE} vectors $\B{\tau}_j^{j}$ and $\B{s}_j^{j}$ are obtained leveraging information up to the $j$-th task from sample sets $D_1, D_2, \ldots, D_{j}$. Specifically, for each $j$-th task, the vectors $\B{\tau}_j^{j}$ and $\B{s}_j^{j}$ are obtained by acquiring information from sample set $D_{j}$ through mean and \ac{MSE} vectors $\B{\tau}_{j}$ and $\B{s}_{j}$ and retaining information from sample sets $D_1, D_2, \ldots, D_{j-1}$ through mean and \ac{MSE} vectors $\B{\tau}_{j-1}^{j-1}$ and $\B{s}_{j-1}^{j-1}$. 

Recursions in~\eqref{eq:tau_general}-\eqref{eq:s_general} can be applied to multiple supervised learning scenarios as detailed in Section~\ref{sec:scenarios}. For instance, the proposed methodology applied to \ac{CL} obtains forward mean vector $\B{\tau}_k^k$ for the $k$-th task leveraging information from preceding tasks using the forward mean vector $\B{\tau}_{k-1}^{k-1}$, the sample average $\B{\tau}_k$, and the expected quadratic change between consecutive tasks $\B{d}_{k}$. Specifically, the mean vector $\B{\tau}_k^k$ is obtained by adding a correction to the sample average ${\B{\tau}}_k$. This correction is proportional to the difference between ${\B{\tau}}_k$ and $\B{\tau}_{k-1}^{k-1}$ multiplied by the smoothing gain $\B{\eta}_{k}^{k}$ that depends on the \ac{MSE} vectors $\B{s}_k,  \B{s}_{k-1}^{k-1}$ and the expected quadratic change $\B{d}_{k}$. In particular, if $\B{s}_k \ll \B{s}_{k-1}^{k-1} +\B{d}_k$, the mean vector is given by the sample average as in single-task learning, and if $\B{s}_k \gg \B{s}_{k-1}^{k-1} +\B{d}_k$, the mean vector is given by that of the preceding task.

The result below shows that using exact values for variances and expected quadratic changes, the previous recursions obtain the unbiased linear estimates of the mean vector $\B{\tau}_j^\infty = \mathbb{E}_{\up{p}_j}\{\Phi(x, y)\}$ with minimum \ac{MSE}.

\begin{teorema}\label{th:kalman1}
Let $\B{\sigma}_j^2$ and $\B{d}_j$ in recursions \eqref{eq:tau_general}-\eqref{eq:gain_general} be the actual variance and the actual expected quadratic change between tasks, that is $\B{\sigma}_j^2=[\mathbb{V}\text{ar}_{\up{p}_j}\{\Phi^{(1)}(x,y)\}, \mathbb{V}\text{ar}_{\up{p}_j}\{\Phi^{(2)}(x,y)\}, \ldots, \mathbb{V}\text{ar}_{\up{p}_j}\{\Phi^{(m)}(x,y)\}]^\top$ and \mbox{$\B{d}_j  = \mathbb{E}\{\B{w}_j^2\} = \mathbb{E}\{(\B{\tau}_j^\infty - \B{\tau}_{j-1}^\infty)^2\}$}. If the tasks' distributions satisfy \eqref{eq:td}, then we have that $\B{\tau}_j^{j}$ given by \eqref{eq:tau_general} is the unbiased linear estimator of the mean vector ${\B{\tau}}_j^\infty$ based on $D_1, D_2, \ldots, D_{j}$ that has the minimum \ac{MSE}, and $\B{s}_j^{j}$ given by \eqref{eq:s_general} is its \ac{MSE}. 
 
 \vspace{0.5cm}
\begin{proof}
The mean vectors evolve over tasks through the linear dynamical system (state-space model with white noise processes)  
\begin{align}
\label{eq:w1}
{\B{\tau}}_j^\infty & = {\B{\tau}}_{j-1}^\infty + \B{w}_{j}
\end{align}
where vectors $\B{w}_{j}$ for $j = 2, 3, \ldots$ are independent and zero-mean. Equation~\eqref{eq:w1} is obtained because the sequence of probability distribution satisfy the~\eqref{eq:td} assumption, that is 
\mbox{$\up{p}_{j} = \up{p}_{j-1} + \varepsilon_{j}$} where $\varepsilon_{j}$ for $j = 2, 3, \ldots$ are independent and zero-mean. In addition, each state variable $\B{\tau}_j^\infty$ is observed at each step $j$ through $\B{\tau}_j$ so that we have
\begin{align}
\label{eq:v1}
\B{\tau}_j & = {\B{\tau}}_j^\infty + \B{v}_{j}
\end{align}
where vectors $\B{v}_j$ for $j \in \{2, 3, \ldots, k\}$ are independent and zero-mean because $\B{\tau}_j$ is the sample average of i.i.d. samples. For the above dynamical systems, the Kalman filter recursions provide the unbiased linear estimator with minimum MSE based on samples corresponding to preceding steps $D_1, D_2, \ldots, D_j$. Then, equations~\eqref{eq:tau_general} and \eqref{eq:s_general} are obtained after some algebra from the Kalman filter recursions (see details in Appendix~\ref{app:rec}).
\end{proof}
\end{teorema}

The above theorem shows that equations in~\eqref{eq:tau_general}-\eqref{eq:gain_general} enable to recursively obtain the mean vector estimate for each task as well as its \ac{MSE} vector leveraging information from preceding tasks.

The vectors  $\B{d}_j$ and $\B{\sigma}_j^2$ in the above recursions can be estimated for each $j$-th task using sample sets. As described in Section~\ref{sec:gf}, the vector $\B{d}_j$ assesses the expected quadratic change between consecutive tasks $\mathbb{E}\{\B{w}_j^2\} = \mathbb{E}\{(\B{\tau}_j^\infty - \B{\tau}_{j-1}^\infty)^2\}$. Such expectation can be estimated by the sample average of the most recent samples as
\begin{equation} 
 \label{eq:d}
\B{d}_j = \frac{1}{W} \sum_{l = 1}^{W}\left(\B{\tau}_{j_{l}} - \B{\tau}_{j_{l-1}}\right)^2
 \end{equation}
where $j_0, j_1, \ldots, j_{W}$ are the $W+1$ closest indexes to $j$ and sample average $\B{\tau}_j$ is given by~\eqref{eq:tau1}. The vector $\B{\sigma}_{j}^{2}$ assesses the variance of the samples of the $j$-th task $\mathbb{V}\text{ar}_{\up{p}_{j}}\{\Phi(x, y)\}$. Such variance can be estimated by the sample variance of sample set $D_j$. As shown in Section~\ref{sec:ess} below, the usage of approximated values for $\B{d}_j$ and $\B{\sigma}_{j}^{2}$ does not significantly affect the performance of the methods proposed.

\subsection{Forward and backward learning}

This section presents the recursions that allow us to obtain mean and \ac{MSE} vectors for each task leveraging information from preceding and succeeding tasks. The proposed techniques obtain for each $j$-th task with \mbox{$j = \{1, 2, \ldots, k\}$} forward and backward mean and \ac{MSE} vectors leveraging information from the $k$ tasks in the sequence. Let $\B{\tau}_{j}^{k}$ and $\B{s}_{j}^{k}$ denote the mean and \ac{MSE} vectors for forward and backward learning corresponding to the $j$-th task. The following recursions allow us to obtain $\B{\tau}_{j}^{k}, \B{s}_{j}^{k}\in \mathbb{R}^m$ using forward mean and \ac{MSE} vectors $\B{\tau}_{j}^{j}, \B{s}_{j}^{j}$ together with a smoothing gain $\B{\eta}_{j}^{k}$
as follows
\begin{align}
\label{eq:tau_general_back}
      \B{\tau}_{j}^{k}& =  \B{\tau}_{j+1}^k + \B{\eta}_j^{k} \left(\B{\tau}_{j}^{j} - \B{\tau}_{j+1}^{k}\right) \\
      \label{eq:s_general_back}
       \B{s}_{j}^{k} & = \B{s}_{j+1}^{k} + \B{\eta}_j^k (\B{s}_{j}^{j} - 2 \B{s}_{j+1}^{k} + \B{\eta}_j^k \B{s}_{j+1}^{k})\\
                     \label{eq:gain_general_back}
\B{\eta}_j^{k} & =  \frac{\B{d}_{j+1}} {\B{s}_{j}^{j} + \B{d}_{j+1}}
\end{align}
for $j = k-1, k-2, \ldots, 1$ with $\B{\tau}_j^j, \B{s}_j^j$ given by~\eqref{eq:tau_general}-\eqref{eq:s_general}. Mean and \ac{MSE} vectors $\B{\tau}_j^k, \B{s}_j^k$ are obtained leveraging information from all the sample sets $D_1, D_2, \ldots, D_k$. Specifically, for each $j$-th task, the vectors $\B{\tau}_j^k, \B{s}_j^k$ are obtained acquiring information from sample sets $D_{j+1}, D_{j+2}, \ldots, D_k$ through mean and \ac{MSE} vectors $\B{\tau}_{j+1}^k, \B{s}_{j+1}^k$ and retaining information from sample sets $D_1, D_2, \ldots, D_j$ through mean and \ac{MSE} vectors $\B{\tau}_j^j, \B{s}_j^j$.

Similarly as the forward learning recursions in~\eqref{eq:tau_general}-\eqref{eq:s_general}, forward and backward learning recursions in~\eqref{eq:tau_general_back}-\eqref{eq:s_general_back} can be applied to multiple supervised learning scenarios as detailed in Section~\eqref{sec:scenarios}. At each step $k$, the proposed methodology applied to \ac{CL} obtains the mean and \ac{MSE} vectors for the $j$-th task by retaining information from preceding tasks and acquiring information from the new task. First, we obtain mean and \ac{MSE} vectors $\B{\tau}_k^k$ and $\B{s}_k^k$, in that case forward and backward learning recursions in~\eqref{eq:tau_general_back}-\eqref{eq:s_general_back} coincide with forward learning recursions in~\eqref{eq:tau_general}-\eqref{eq:s_general}. Then, we obtain mean and \ac{MSE} vectors $\B{\tau}_j^{k}$ and $\B{s}_j^{k}$ that retain information from preceding tasks through mean and \ac{MSE} vectors $\B{\tau}_j^{j}$ and $\B{s}_j^{j}$ and acquire information from the $k$-th task through mean and \ac{MSE} vectors  $\B{\tau}_{j+1}^{k}$ and $\B{s}_{j+1}^{k}$.

Recursions in~\eqref{eq:tau_general_back}-\eqref{eq:s_general_back} obtain the mean vector $\B{\tau}_{j}^{k}$ by adding a correction to the mean vector of the succeeding task $\B{\tau}_{j+1}^{k}$ obtained for forward and backward learning. This correction is proportional to the difference between $\B{\tau}_{j}^{j}$ and $\B{\tau}_{j+1}^{k}$ multiplied by the smoothing gain $\B{\eta}_{j}^{k}$ that depends on the \ac{MSE} vector $\B{s}_{j}^{j}$ and the expected quadratic change $\B{d}_{j+1}$. In particular, if $\B{s}_{j}^{j} \gg \B{d}_{j+1}$, the mean vector is given by that of the corresponding task for forward learning, while if $\B{s}_{j}^{j} \ll \B{d}_{j+1}$, the mean vector is given by that of the next task for backward learning. 

The result below shows that using exact values for variances and expected quadratic changes, the previous recursions obtain the unbiased linear estimates of the mean vector $\B{\tau}_j^\infty = \mathbb{E}_{\up{p}_j}\{\Phi(x, y)\}$ with minimum \ac{MSE}.

\begin{teorema}\label{th:kalman}
Let $\B{\sigma}_j^2$ and $\B{d}_j$ in recursions \eqref{eq:tau_general_back}-\eqref{eq:gain_general_back} be the actual variance and the actual expected quadratic change between tasks, that is $\B{\sigma}_j^2=[\mathbb{V}\text{ar}_{\up{p}_j}\{\Phi^{(1)}(x,y)\}, \mathbb{V}\text{ar}_{\up{p}_j}\{\Phi^{(2)}(x,y)\}, \ldots, \mathbb{V}\text{ar}_{\up{p}_j}\{\Phi^{(m)}(x,y)\}]^\top$ and \mbox{$\B{d}_j  = \mathbb{E}\{\B{w}_j^2\} = \mathbb{E}\{(\B{\tau}_j^\infty - \B{\tau}_{j-1}^\infty)^2\}$}. If the tasks' distributions satisfy \eqref{eq:td}, then we have that $\B{\tau}_j^{k}$ given by \eqref{eq:tau_general_back} is the unbiased linear estimator of the mean vector ${\B{\tau}}_j^\infty$ based on $D_1, D_2, \ldots, D_j, \ldots, D_k$  that has the minimum \ac{MSE}, and $\B{s}_j^{k}$ given by \eqref{eq:s_general_back} is its \ac{MSE}.

\vspace{0.5cm}
\begin{proof}
Since the sequence of probability distribution satisfy the~\eqref{eq:td} assumption, we have that the mean vectors evolve over tasks through the linear dynamical system in~\eqref{eq:w1}-\eqref{eq:v1}  where vectors $\B{w}_j$ and $\B{v}_j$ for $j \in \{2, 3, \ldots, k\}$ are independent and zero-mean. For such systems, the Rauch-Tung-Striebel recursions provide the unbiased linear estimator with minimum MSE based on samples sets $D_1, D_2, \ldots, D_k$. Then, equations~\eqref{eq:tau_general} and \eqref{eq:s_general} are obtained from the Rauch-Tung-Striebel recursions (see details in Appendix~\ref{app:rec}).
\end{proof}
\end{teorema}

The above theorem shows that equations in~\eqref{eq:tau_general_back}-\eqref{eq:gain_general_back} enable to recursively obtain the mean vector estimate for each task as well as its \ac{MSE} vector leveraging all the available information.

Recursions in~\eqref{eq:tau_general}-\eqref{eq:s_general} and recursions in~\eqref{eq:tau_general_back}-\eqref{eq:s_general_back} adapt to multidimensional tasks' changes by accounting for the change of all the mean vector components. Specifically, for each $j$-th task and each $i$-th component for $i = 1, 2, \ldots, m$, the $i$-th component of the mean vector estimate is updated by using the corresponding component of the expected quadratic change, the estimate for the consecutive task, and the most recent sample average. Such update accounts for the specific evolution of each $i$-th component of the mean vector through the recursions in equations~ \eqref{eq:tau_general} and~\eqref{eq:tau_general_back} and smoothing gains ($\B{\eta}_{j}^{j}$, $\B{\eta}_{j}^{k}$) in equations~\eqref{eq:gain_general} and~\eqref{eq:gain_general_back}. In particular, updates for mean vector components with a low smoothing gain slightly change the estimate for the consecutive task (previous task in~\eqref{eq:tau_general} and next task in~\eqref{eq:tau_general_back}), while those updates for components and tasks with a high smoothing gain increase the relevance of the information of the corresponding task (sample average in~\eqref{eq:tau_general} and forward mean vector in~\eqref{eq:tau_general_back}). 

\section{Effective sample sizes and performance guarantees}\label{sec:ess}

This section analytically characterizes the \ac{ess} increase obtained by using both forward and backward learning. Then, we compare the \ac{ess} of the proposed methodology with conventional techniques that adapt to evolving tasks. 

The \ac{ess} commonly quantifies the performance improvement of an algorithm in terms of the number of samples the baseline method would require to achieve the same performance.
 In the baseline approach of single-task learning, \acp{MRC} provide bounds for the minimax risk in terms of the smallest minimax risk as described in~\cite{mazuelas:2020, mazuelas:2022, MazRomGru:22}. The smallest minimax risk corresponds with the ideal case of knowing mean vectors exactly, that is, the minimax risk corresponding with the uncertainty set $$\set{U}_j^\infty=\{\up{p} \in \Delta(\set{X}~\times~\set{Y}):\mathbb{E}_{\up{p}}\{\Phi(x, y)\} = \B{\tau}_j^\infty\}$$ given by the expectation $\B{\tau}_j^\infty = \mathbb{E}_{\up{p}_j}\{\Phi(x, y)\}$. The minimum worst-case error probability over distributions in $\set{U}_j^\infty$ is given by 
$$R_j^\infty = \min_{\B{\mu}} 1 - {\B{\tau}_j^\infty}^\top \B{\mu} + \varphi(\B{\mu}) = 1 - {\B{\tau}_j^\infty}^\top \B{\mu}_j^\infty +  \varphi(\B{\mu}_j^\infty)$$
corresponding with the rule given by parameters $\B{\mu}_j^\infty$. Such classification rule is referred to as optimal minimax rule because for any uncertainty set $\set{U}_j^k$ given by~\eqref{eq:us} that contains the underlying distribution, we have that $\set{U}_j^\infty \subseteq \set{U}_j^k$ and hence $R_j^\infty \leq R(\set{U}_j)$. 

The baseline approach of single-task learning obtains uncertainty sets given by~\eqref{eq:us} using the mean and confidence vector provided by~\eqref{eq:tau1}. In that case, with probability at least $1 - \delta$ we have that 
\begin{equation}
\label{eq:bound_singletask}
R(\set{U}_j) \leq R_j^\infty + \frac{M (\kappa \sqrt{2 \log(2 m/\delta)}+\lambda_{0})}{\sqrt{n_j}}\|\B{\mu}_j^\infty\|_1
\end{equation}
where $M$ is a bound for the feature mapping, i.e., $\|\Phi(x, y)\|_\infty \leq M$  for any $(x,y)\in\set{X}\times\set{Y}$.   
Coefficient $\kappa > 0$ is a constant that  describes how the bounds are affected by the inaccuracies of the values for $\B{\sigma}_{j}$. This coefficient shows the relationship between the values used for $\B{\sigma}_j$ and the sub-Gaussian parameters of $\Phi_j$ as \mbox{$subG({\Phi}_j) \preceq \kappa \B{\sigma}_j$} where $\Phi_j$ denotes the random variable given by the feature mapping of samples from the $j$-th task, and $subG(\V{z})$ denotes the vector composed by the sub-Gaussian parameters of each component of random vector $\V{z}$ (i.e., $\mathbb{E}\{e^{t(\V{z} - \mathbb{E}\{\V{z}\})}\}\preceq e^{subG(\V{z})^2t^2/2}$).  The sub-Gaussian condition is necessary to obtain high-probability bounds and is satisfied with wide generality. In particular, in this paper we consider feature mappings that are bounded, which ensures that variable $\Phi_j$ is also bounded and hence sub-Gaussian (see e.g., \citet{wainwright2019high}). Inequality~\eqref{eq:bound_singletask} is obtained using the bounds in~\cite{MazRomGru:22} together with the Chernoff bound (see e.g., equation (2.9) in Section 2.1.2 in \citet{wainwright2019high}) for sample averages of sub-Gaussian variables.

\subsection{Effective sample sizes using forward learning}

The following result provides bounds for the minimax risk for each task using forward learning. Specifically, the theorem below provides bounds for the minimax risk of each $j$-th task with respect to the smallest minimax risk using forward mean and \ac{MSE} vectors $\B{\tau}_j^{j}$ and $\B{s}_j^{j}$ obtained as in~\eqref{eq:tau_general} and~\eqref{eq:s_general}, respectively.

\begin{teorema} \label{th:ess_recursion}
Let $\Phi(x, y)$ be a feature mapping bounded by $M$, and $\set{U}_j^j$ be the uncertainty set given by~\eqref{eq:us} using the mean and confidence vectors  $\B{\tau}_j^j$ and $\B{\lambda}_j^j = \lambda_{0} \sqrt{\B{s}_j^j}$ provided by~\eqref{eq:tau_general} and~\eqref{eq:s_general}. If the vectors $\B{\sigma}_j^{2}$ and $\B{d}_j$ utilized in recursions~\eqref{eq:tau_general}-\eqref{eq:gain_general} satisfy that $subG({\Phi}_j) \preceq \kappa \B{\sigma}_{j}$ and $subG \left(\B{w}_{j}\right)\preceq \kappa \sqrt{\B{d}_{j}}$ for $\kappa>0$ and $j = 1, 2, \ldots, k$. 
Then, under the evolving task assumption in~\eqref{eq:td}, with probability at least $1 - \delta$  we have that
\begin{align}
\label{eq:ess}
R(\mathcal{U}_j^{j}) \leq R_j^{\infty} + \frac{M (\kappa \sqrt{2 \log(2 m /\delta)} + \lambda_{0})}{\sqrt{n_j^{j}}} \left\|\B{\mu}_j^{\infty}\right\|_1
\end{align}
with 
\begin{equation}
\label{eq:ess_preceding_current}
n_j^j \geq n_j + n_{j-1}^{j-1} \frac{\|\B{\sigma}_{j}^2\|_\infty}{\|\B{\sigma}_{j}^2\|_\infty +  \|\B{d}_j\|_\infty n_{j-1}^{j-1}}
\end{equation}
 for $j = 2, 3, \ldots k$ and $n_1^1 = n_1$.
 \begin{proof}
See Appendix~\ref{eq:al}.
\end{proof}
\end{teorema}

Theorem~\ref{th:ess_recursion} shows that the value $n_{j}^{j}$ in~\eqref{eq:ess} is the \ac{ess} of the proposed methods since the bound in~\eqref{eq:ess} coincides with that of single-task learning in~\eqref{eq:bound_singletask} if the sample size for the $j$-th task is $n_{j}^{j}$. In particular, single-task learning would require $n_{j}^{j} > n_{j}$ samples to achieve the same performance bound than forward learning using $n_{j}$ samples. 
As shown in~\eqref{eq:ess_preceding_current}, the \ac{ess} of each task is obtained by adding a fraction of the \ac{ess} for the preceding task to the sample size. In particular, if $\B{d}_j$ is large, the \ac{ess} is given by the sample size, while if $\B{d}_j$ is small, the \ac{ess} is given by the sum of the sample size and the \ac{ess} of the preceding task. 

The bound in~\eqref{eq:ess} shows that recursions in~\eqref{eq:tau_general} do not need to use very accurate values for $\B{\sigma}_j$ and $\B{d}_j$. Specifically, the coefficient $\kappa$ in~\eqref{eq:ess} can be taken to be small as long as the values used for $\B{\sigma}_j$ and $\B{d}_j$ are not much lower than the sub-Gaussian parameters of $\Phi_j$ and $\B{w}_j$, respectively. In particular, $\kappa$ is smaller than the maximum of $M/\min_{j, i}\{\sigma_j^{(i)}\}$ and $2M/\min_{j, i} \{\sqrt{d_j^{(i)}}\}$ due to the bound for the sub-Gaussian parameter of bounded random variables (see e.g., \citet{wainwright2019high}).

The above theorem shows the~\ac{ess} obtained by forward learning techniques in terms of the \ac{ess} for the preceding task. The following result allows us to directly quantify the \ac{ess} in terms of the sample size, the number of tasks, and the expected quadratic change.

\begin{teorema} \label{th:forward}
Let $d$, $\B{\sigma}_j$, and $n$ be such that $d \geq \|\B{d}_j\|_\infty$, $\|\B{\sigma}_j^2\|_\infty \leq 1$, and $n \leq {n}_j$ for $j = 1, 2, \ldots, k$. Then, we have that the \ac{ess} in~\eqref{eq:ess} satisfies
\begin{equation}
\label{eq:N_ineq}
n_j^{j}  \geq n\left(1 + \left(\frac{1 + \alpha}{\alpha}\right)\frac{(1+\alpha)^{2 j-2} -1}{(1+\alpha)^{2 j-1}+1}\right) \; \text{ with } \; \alpha = \frac{2}{\sqrt{1+\frac{4}{nd}} - 1}.
\end{equation}
In particular, for $j > 1$, we have that 
\begin{alignat*}{5}
 n_j^{j} & \geq  n\left(1 + \frac{j-1}{3}\right) && \text{ if } && n d < \frac{1}{j^2}\\
n_j^{j} & \geq n\left(1 + \frac{1}{5 \sqrt{n d}}\right) \; \; \; \; && \text{ if } \; \; \; \; & \frac{1}{j^2} \leq \; & n d < 1\\
n_j^{j} & \geq n\left(1 + \frac{1}{3n d}\right) &&\text{ if } && n d \geq 1.
\end{alignat*}
\begin{proof}
See Appendix \ref{ap:proofthbound}.
\end{proof}
\end{teorema}

\begin{figure}
     \begin{center}
     \psfrag{AMRC j = 50abcdefghijklm}[l][l][0.7]{AMRCs $j = 50$}
     \psfrag{W = 5}[l][l][0.7]{Window $W_c = 5$}
     \psfrag{W = 25}[l][l][0.7]{Window $W_c = 25$}
      \psfrag{W = 45}[l][l][0.7]{Window $W_c = 45$}
     \psfrag{Window optimaabcdefghijklmnopqrstuv}[l][l][0.7]{Window $W_c=\left\lfloor\sqrt{\frac{3}{n d} + \frac{1}{2}}\right\rfloor$}
     \psfrag{ESS/n}[c][][0.7]{$\text{ESS}/n$}
     \psfrag{Number of tasks}[][][0.7]{Number of tasks}
     \psfrag{0.6}[][][0.5]{$0.6$}
     \psfrag{0.4}[][][0.5]{$0.4$}
     \psfrag{0.3}[][][0.5]{$0.3$}
     \psfrag{0.2}[][][0.5]{$0.2$}
          \psfrag{10(-4)abcdefghijkl}[l][l][0.7]{$n d = 10^{-4}$}
           \psfrag{data3}[l][l][0.7]{$n d = 10^{-3}$}
          \psfrag{data2}[l][l][0.7]{$n d = 10^{-2}$}
          \psfrag{data1}[l][l][0.7]{$n d = 10^{-1}$}
          \psfrag{0.001}[][][0.5]{}
      \psfrag{25}[][][0.5]{$25$}
      \psfrag{5}[][][0.5]{$5$}
      \psfrag{1}[][][0.5]{$1$}
       \psfrag{0}[][][0.5]{$0$}
      \psfrag{10}[][][0.5]{$10$}
       \psfrag{20}[][][0.5]{$20$}
        \psfrag{40}[][][0.5]{$40$}
         \psfrag{60}[][][0.5]{$60$}
          \psfrag{80}[][][0.5]{$80$}
       \psfrag{50}[][][0.5]{$50$}
       \psfrag{100}[][][0.5]{$100$}
     \includegraphics[width=0.5\textwidth]{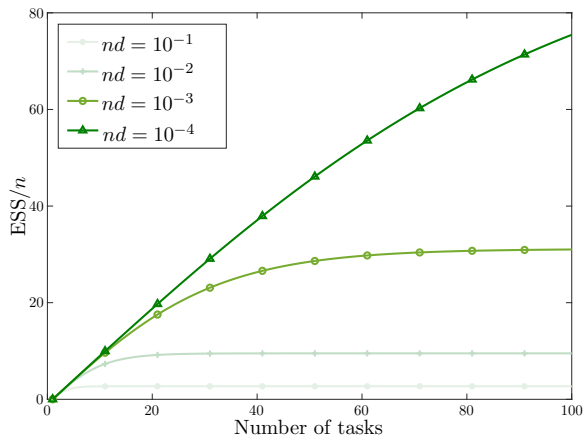}
   \end{center}
    \caption{The ESS provided by forward learning significantly increases with the number of tasks especially for small values for the sample size $n$ and the expected quadratic change $d$.}
       \label{fig:ess_cd}
\end{figure}

The above theorem characterizes the increase in the ESS using forward learning in comparison with single-task learning $(\alpha>0$ and $(1+\alpha)^{2j-2}>1 \text{ for }j > 1)$. For $j = 1$, the ESS equals the sample size $n$ because forward learning coincides with single-task learning. For $j > 1$, the ESS is significantly larger using forward learning in comparison with single-task learning for reduced samples sizes $n$ and small tasks’ expected quadratic change $d$ ($\alpha$ significantly larger than 0).   In addition, the ESS in~\eqref{eq:N_ineq} grows monotonically with the number of tasks $j$ and becomes proportional to $j$ when the expected quadratic change is smaller than $1/(j^2n)$.  $\mbox{Figure}$~\ref{fig:ess_cd} further illustrates the increase in \ac{ess} with respect to the sample size due to forward learning in terms of the number of tasks, the sample size, and the expected quadratic change between consecutive tasks. This figure describes the increase in \ac{ess} with respect to the sample size for $nd = 10^{-4}$, $10^{-3}$, $10^{-2}$, and $10^{-1}$, and a varying number of tasks. In particular, the figure shows that the \ac{ess} of the proposed methodology significantly increases with the number of tasks especially for small sample sizes.  

Existing methods for learning a sequence of tasks provide comparable performance bounds that decrease with the number of tasks $j$ and the sample size $n$ \citep{denevi2019learning, pentina2015lifelong, denevi2019online, khodak2019adaptive, balcan2019provable}. For instance, in scenarios where tasks follow the~\eqref{eq:iid} assumption the bounds depend on \(n\) and \(j\) as $\mathcal{O}(\frac{1}{\sqrt{j}}) + \mathcal{O}(\frac{D_{\text{Var}}}{\sqrt{n}})$ in \citet{denevi2019learning}, where $D_{\text{Var}}$ is given by the ``relatedness among tasks'' and quantifies the tasks variability. In scenarios where tasks are evolving, the bounds depend on $n$ an $j$ as \(\mathcal{O}(\frac{1}{\sqrt{j}}) + \mathcal{O}(\frac{D_{KL}}{\sqrt{n}})\) in \citet{pentina2015lifelong}, where \(D_{KL}\) is the KL-divergence describing the difference between consecutive tasks.  As shown in Theorems~\ref{th:ess_recursion} and~\ref{th:forward}, the presented bounds depend on \(n\) and \(j\) as \(\mathcal{O}(\frac{1}{\sqrt{\text{ESS}(n,d,j)}})\), where $d$ is the expected quadratic change between consecutive tasks.  Similarly to previous works, our bounds decrease when $n$ and $j$ increase and increase when $d$ increases, as shown in~\eqref{eq:N_ineq} and \eqref{eq:N_bineq}. In particular, if $nd< \frac{1}{j^2}$, Theorem~\ref{th:forward} shows that the proposed methodology provides bounds that depend on $n$ and $j$ as $\mathcal{O}(\frac{1}{\sqrt{nj}})$. Differently from other techniques, the proposed methodology provides bounds that decrease to zero increasing the number of samples for any number of tasks $j$. In particular, as shown in Theorem 4, we have that ESS$(n,d,j)= {\Omega}(n)$ for any $d$ and $j$. The next section, shows how the usage of backward learning results in even larger \acp{ess}.

\subsection{Effective sample sizes using forward and backward learning}

The following result provides bounds for the minimax risk for each task using forward and backward learning

\begin{teorema} \label{th:ess_recursion_continual} 
Let $n_j^j$ and $\kappa$ be as in Theorem~\ref{th:ess_recursion} for any $j \in \{1, 2, ..., k\}$. Then, under the evolving task assumption in~\eqref{eq:td}, with probability at least $1 - \delta$ we have  that
\begin{align}
\label{eq:ess_continual}
R(\set{U}_j^{k}) \leq R^\infty_j +  \frac{M (\kappa \sqrt{2 \log(2 m /\delta)} + \lambda_{0})}{\sqrt{n_j^{k}}} \left\|\B{\mu}_j^{\infty}\right\|_1
\end{align}
$$\hspace{-4cm}\text{with }\hspace{4cm}n_{j}^{k} \geq n_j^{j} + n_{j+1}^{-k} \frac{\|\B{\sigma}_{j}^2\|_\infty}{\|\B{\sigma}_{j}^2\|_\infty + n_{j+1}^{-k} \|\B{d}_{j+1}\|_\infty}$$
for $j < k$, where $n_{j+1}^{-k}$ satisfies $n_{k}^{-k} = n_{k}$ and $n_{j}^{-k} \geq n_{j} + n_{j+1}^{-k}\frac{\|\B{\sigma}_{j}^2\|_\infty}{\|\B{\sigma}_{j}^2\|_\infty + n_{j+1}^{-k} \|\B{d}_{j+1}\|_\infty}$.
    \begin{proof}
See Appendix~\ref{ap:th4}.
\end{proof}
 \vspace{-0.7cm}
\end{teorema}

Theorem~\ref{th:ess_recursion_continual} shows that the value $n_{j}^{k}$ in~\eqref{eq:ess_continual} is the \ac{ess} of the proposed methods since the bound in~\eqref{eq:ess_continual} coincides with that of single-task learning in~\eqref{eq:bound_singletask} if the sample size for the $j$-th task is $n_{j}^{k}$. That is, $n_{j}^{k}$ is the \ac{ess} of the proposed methodology using forward and backward learning since single-task learning would require $n_{j}^{k} > n_{j}$ samples to achieve the same performance than forward and backward learning using $n_{j}$ samples. 
In addition, Theorem~\ref{th:ess_recursion_continual} shows that the methods proposed can increase the \ac{ess} of each task by leveraging information from all the tasks in the sequence. In particular, the bounds provided by inequality~\eqref{eq:ess_continual} are lower than those in Theorem~\ref{th:ess_recursion}. The \ac{ess} of each $j$-th task is obtained by adding a fraction of the \ac{ess} for the next task to the \ac{ess} of the corresponding task leveraging information from preceding tasks. In particular, if $\B{d}_j$ is large, the \ac{ess} is given by that leveraging information from $j$ tasks, while if $\B{d}_j$ is small, the \ac{ess} is given by the sum of the \ac{ess} leveraging information from $j$ tasks and the \ac{ess} of the next task. 
 
The above theorem shows the increase in \ac{ess} in terms of the \ac{ess} for consecutive tasks. The following result allows to directly quantify the \ac{ess} in terms of the sample size and the expected quadratic change.
\begin{teorema} \label{th:backward}
Let $d$, $\B{\sigma}_j$, and $n$ be such that $d \geq \|\B{d}_j\|_\infty$, $\|\B{\sigma}_j^2\|_\infty \leq 1$, and $n \leq {n}_j$ for $j = 1, 2, \ldots, k$. For any $j \in \{1, 2, \ldots, k\}$, we have that the \ac{ess} in~\eqref{eq:ess_continual} satisfies
\begin{equation}
\label{eq:N_bineq}
n_j^{k} \geq n\left(1 + \left(\frac{1+\alpha}{\alpha}\right)\frac{(1+\alpha)^{2 j-2} -1}{(1+\alpha)^{2 j-1}+1}+  \left(\frac{1+\alpha}{\alpha}\right) \frac{(1+\alpha)^{2 (k-j)}- 1}{(1+\alpha)^{2 (k-j)+1}+1}\right)
\end{equation}
with ${\alpha} = \frac{2}{\sqrt{1+\frac{4}{nd}} - 1}$. In particular, for $j > 1$, we have that 
\begin{alignat*}{5}
n_j^{k}& \geq n_j^j + n \frac{j(k-j)}{j+ 2 {(k-j)}} && \geq n\left(1+ \frac{ j-1}{3} + \frac{j(k-j)}{j+ 2 {(k-j)}} \right)  \; \; \; \;&& \text{ if } \; \; \; \; && n d < \frac{1}{j^2}\\
n_j^{k} & \geq n_j^j + \frac{1}{5} \sqrt{\frac{n}{d}}&&\geq n\left(1+ \frac{2}{5 \sqrt{n d}}\right) && \text{ if }& \frac{1}{j^2} \; \leq \;& nd < 1\\
n_j^{k} & \geq n_j^j + \frac{1}{3 d}&& \geq n\left(1+ \frac{2}{3n d}\right) && \text{ if }&& n d \geq 1
\end{alignat*}
where $n_j^j$ satisfies~\eqref{eq:N_ineq}.
\begin{proof}
See Appendix \ref{ap:proofthbackward}.
\end{proof}
\end{teorema}

\begin{figure}
  \begin{center}
    \psfrag{ess/n}[][][0.8]{ESS/$n$}
     \psfrag{100}[r][r][0.6]{$100$}
     \psfrag{0.0001}[r][r][0.6]{$100^{-2}$}
      \psfrag{0.1}[][][0.6]{$0.1$}
      \psfrag{10}[r][r][0.6]{$10$}
     \psfrag{5}[r][r][0.6]{$5$}
      \psfrag{2}[r][r][0.6]{$2$}
      \psfrag{20}[r][r][0.6]{$20$}
       \psfrag{1}[l][l][0.6]{$1$}
         \psfrag{0.01}[][][0.6]{$100^{-1}$}
           \psfrag{0}[][][0.6]{0}
             \psfrag{4}[][][0.6]{4}
         \psfrag{Forw j = 10}[l][l][0.8]{$n_{10}^{10}/n$}
         \psfrag{Forw k = 13}[l][l][0.8]{$n_{10}^{13}/n$}
         \psfrag{Forw k = 20}[l][l][0.8]{$n_{10}^{20}/n$}
         \psfrag{Forw k = 100}[l][l][0.8]{$n_{100}^{100}/n$}
                  \psfrag{Forw k = 103}[l][l][0.8]{$n_{100}^{103}/n$}
         \psfrag{Forw k = 110}[l][l][0.8]{$n_{100}^{110}/n$}
                                \psfrag{nu/n}[][][0.8]{$n d$}
         \includegraphics[width=0.5\textwidth]{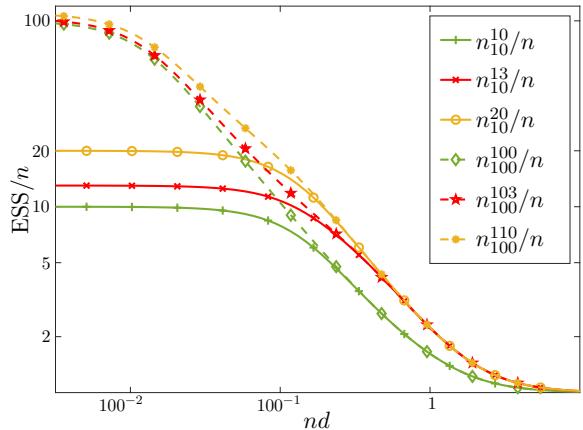}
  \end{center}
              \caption{The ESS provided by forward and backward learning (red and orange lines) significantly increases in comparison with forward learning (green line) especially for small values for the sample size $n$ and the expected quadratic change $d$.}
                       \label{fig:ess_ll_2}
\end{figure}

The above theorem characterizes the increase in the ESS using forward and backward learning in comparison with single-task learning and forward learning. 
For $j = k$, the ESS equals the ESS using forward learning in Theorem 4 because forward and backward learning coincides with forward learning. For $j < k$, the ESS is significantly higher for reduced sample sizes and small tasks' expected quadratic change. In addition, the ESS in~\eqref{eq:N_bineq} grows monotonically with the number of preceding tasks ($j$) and with the number of succeeding tasks ($k-j$). It becomes proportional to the total number of tasks $k$ when the expected quadratic change is smaller than $1/(j^2n)$ and $j\geq k/2$.  
$\mbox{Figure}$~5 further illustrates the increase in \ac{ess} due to forward and backward learning in comparison with forward learning. As the figure shows, the increase in the ESS can be classified into three regimes depending on the sample size $n$ and the expected quadratic change $d$, as described in Theorems~4 and~6. The ESS is only marginally larger than the sample size for sizeble values of $nd$ (large samples sizes or drastic changes in the distribution); the ESS quickly increases when $nd$ becomes small (reduced sample sizes and moderate changes in the distribution); and the ESS becomes proportional to the total number of tasks $k$ if $nd$ is rather small (very small sample sizes and very slow changes in the distribution).

The next section numerically shows the increase in the \ac{ess} due to forward and backward learning in comparison with conventional techniques designed for evolving tasks. 

\subsection{Numerically assessment of the \acp{ess}}

This section compares the ESS of the proposed methodology with the baseline approach that utilizes sliding windows. One difference between the sliding window approach and the proposed methodology is that sliding windows abruptly discard older data and assign the same relevance to all the tasks in the window. On the other hand, smoothing strategies, such as the one used in this paper, increase the relevance of the most recent tasks. These strategies have been shown to allow for a more gradual adaptation to temporal changes in the data  \citep{muth1960optimal}. In addition, the methods proposed can provide a multidimensional adaptation to changes in the sequence of tasks. For instance, the smoothing gains $\B{\eta}_{j}^{j}$ and $\B{\eta}_{j}^{k}$ in~\eqref{eq:gain_general} and~\eqref{eq:gain_general_back} are vectors that allow us to estimate each component of the mean vectors in a different manner.

\begin{figure}
 \begin{subfigure}[t]{0.48\textwidth}
     \begin{center}
     \psfrag{AMRC j = 50abc}[l][l][0.8]{Proposed}
     \psfrag{W = 5}[l][l][0.8]{$\overline{W} = 5$}
     \psfrag{W = 25}[l][l][0.8]{$\overline{W} = 25$}
      \psfrag{W = 45}[l][l][0.8]{$\overline{W} = 45$}
     \psfrag{Window optimaabcdefghijklmnopqrstuv}[l][l][0.7]{Window $W_c=\left\lfloor\sqrt{\frac{3}{n d} + \frac{1}{2}}\right\rfloor$}
     \psfrag{ess/n}[][][0.8]{ESS/$n$}
     \psfrag{nd}[][][0.8]{$n d$}
     \psfrag{100}[][][0.5]{$100$}
     \psfrag{50}[][][0.5]{$50$}
     \psfrag{5}[][][0.5]{$5$}
          \psfrag{1e-04}[][][0.5]{$100^{-2}$}
          \psfrag{0.01}[][][0.5]{$100^{-1}$}
          \psfrag{1}[][][0.5]{$1$}
          \psfrag{0.001}[][][0.5]{}
      \psfrag{25}[][][0.5]{$25$}
      \psfrag{45}[][][0.5]{$45$}
       \psfrag{0}[][][0.5]{$0$}
       \psfrag{10}[][][0.5]{$10$}
     \includegraphics[width=\textwidth]{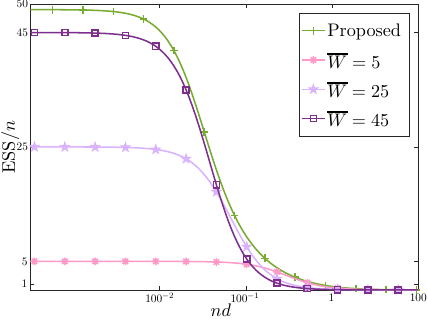}
   \end{center}
    \caption{\ac{ess} of the $j = 50$-th task leveraging information from preceding tasks.}
       \label{fig:ess_ventanas}
       \end{subfigure}
\hfill
      \begin{subfigure}[t]{0.48\textwidth}
      \begin{center}
     \psfrag{AMRC j = 50abc}[l][l][0.8]{Proposed}
     \psfrag{W = 5}[l][l][0.8]{$\overline{W} = 5$}
     \psfrag{W = 50}[l][l][0.8]{$\overline{W} = 50$}
      \psfrag{W = 95}[l][l][0.8]{$\overline{W} = 95$}
     \psfrag{Window optimaabcdefghijklmnopqrstuv}[l][l][0.8]{Window $W_c=\left\lfloor\sqrt{\frac{3}{n d} + \frac{1}{2}}\right\rfloor$}
     \psfrag{ess/n}[][][0.8]{ESS/$n$}
     \psfrag{nd}[][][0.8]{$n d$}
     \psfrag{100}[][][0.5]{$100$}
     \psfrag{50}[][][0.5]{$50$}
     \psfrag{5}[][][0.5]{$5$}
          \psfrag{1e-04}[][][0.5]{$100^{-2}$}
          \psfrag{0.01}[][][0.5]{$100^{-1}$}
          \psfrag{1}[][][0.5]{$1$}
          \psfrag{0.001}[][][0.5]{}
      \psfrag{95}[][][0.5]{$95$}
       \psfrag{0}[][][0.5]{$0$}
     \includegraphics[width=\textwidth]{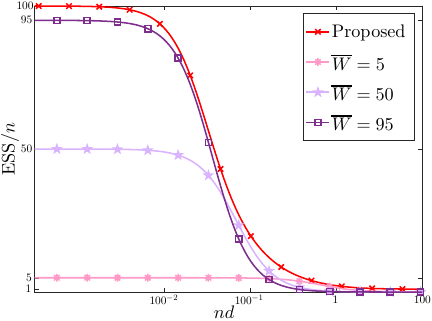}
   \end{center}
    \caption{For a sequence of $k = 100$ tasks, \ac{ess} of the $j = 50$-th task leveraging information from preceding and succeeding tasks.}
       \label{fig:ess_ll}     
       \end{subfigure}
                       \caption{The ESS provided by the proposed methodology is grater than the ESS of the baseline approach that utilizes sliding windows for multiple windows sizes. }
                       \label{fig:ess_succeeding}
\end{figure}

Figure~\ref{fig:ess_succeeding} shows the \ac{ess} increase obtained by the proposed methodology in comparison with the baseline approach that utilizes sliding windows. Techniques based on sliding windows are commonly used to adapt to evolving tasks \citep{zhang2016sliding, tahmasbi2021driftsurf, mazzetto2024adaptive}. 
 Such techniques obtain classification rules for each task using the sample sets corresponding with the $\overline{W}$ closest tasks. Large values of window size adapt to gradual changes in the tasks' distributions, while small values of window size adapt to abrupt changes in the tasks' distributions. Appendix~\ref{app:sliding_windows} shows the expressions corresponding to the above \acp{ess} obtained using sliding windows.

Figure~\ref{fig:ess_ventanas} illustrates the increase in the \ac{ess} due to forward learning in comparison with techniques based on sliding windows of preceding tasks, and Figure~\ref{fig:ess_ll} illustrates the increase in \ac{ess} due to forward and backward learning in comparison with techniques based on sliding windows of preceding and succeeding tasks. Specifically, Figure~\ref{fig:ess_ventanas} and Figure~\ref{fig:ess_ll} show the \ac{ess} of the proposed methods and sliding windows with $\overline{W} = 5, 25,$ and $45$ leveraging information from preceding tasks, and $\overline{W} = 5, 50,$ and $95$ leveraging information from the same number of preceding and succeeding tasks. Small values of window size yield to low \ac{ess} when $nd$ decreases, while large values of window size yield to low \ac{ess} when $nd$ increases. The \ac{ess} provided by the proposed methodology is greater than that provided by techniques based on sliding windows for different number of tasks, sample sizes, and changes between consecutive tasks.
 
\section{Implementation in multiple scenarios with evolving tasks} \label{sec:scenarios}%

This section describes the implementation and the computational and memory complexity of the proposed methodology applied to batch learning scenarios and online learning scenarios.

Multiple supervised learning paradigms obtain classification rules leveraging information only from preceding tasks. For instance, \ac{MDA} methods often obtain classification rules without samples from the target task \citep{Stojanov2020, Li2022}, while \ac{SCD} methods obtain classification rules before receiving samples from the target task  \citep{elwell2011incremental, brzezinski2013reacting}. In these cases, we obtain mean and \ac{MSE} vectors $\B{\tau}_j^{j-1}$ and $\B{s}_j^{j-1}$ for each $j$-th task using information from sample sets $D_1, D_2, \ldots, D_{j-1}$. Then, the sample set corresponding with each $j$-th task arrives after obtaining mean and \ac{MSE} vectors for the $j$-th task. 
Mean and \ac{MSE} vectors $\B{\tau}_j^{j-1}$ and $\B{s}_j^{j-1}$ can be obtained from forward mean and \ac{MSE} vectors $\B{\tau}_{j-1}^{j-1}$ and $\B{s}_{j-1}^{j-1}$ given by~\eqref{eq:tau_general} and \eqref{eq:s_general} as \begin{equation}
\label{eq:concept_drift}
\B{\tau}_j^{j-1} = \B{\tau}_{j-1}^{j-1}, \; \; \; \B{s}_j^{j-1} = \B{s}_{j-1}^{j-1} + \B{d}_j.\end{equation} 

The proposed methodology obtains at each step the classifier parameters and minimax risks for each task using an \ac{ASM} that solves the convex optimization problem~\eqref{eq:mrc} using Nesterov extrapolation strategy \citep{nesterov2015quasi, tao:2019}. In addition, we propose efficient algorithms that use a warm-start for the \ac{ASM} iterations.

The proposed techniques efficiently obtain classifier parameter $\B{\mu}$ and minimax risk $R(\set{U})$ for each task from updated mean vector estimate ${\B{\tau}}$ and confidence vector $\B{\lambda}$. The \ac{ASM} algorithm applied to optimization~\eqref{eq:mrc} obtains classifier parameters using the iterations for {$l~=~1, 2, \ldots, K$}
\begin{align}
&\bar{\B{\mu}}^{l+1)} = \B{\mu}^{l)} + a_l \left({\B{\tau}} - \partial \varphi({\B{\mu}^{l)}}) - \B{\lambda}\text{sign}(\B{\mu}^{l)})\right) \nonumber\\
\label{eq:sgdrule}
&\B{\mu}^{l+1)} = \bar{\B{\mu}}^{l+1)} + \theta_{l+1}(\theta_{l}^{-1} - 1)\left({\B{\mu}}^{l)} - \bar{\B{\mu}}^{l)}\right)\end{align}
where $\B{\mu}^{l)}$ is the $l$-th iterate for $\B{\mu}$, $\theta_l = 2/(l+1)$ and $a_l~=~1/(l+1)^{3/2}$ are step sizes and $\partial \varphi(\B{\mu}^{l)})$ denotes a subgradient of $\varphi(\cdot)$ at $\B{\mu}^{l)}$.

The proposed algorithm reduces the number of \ac{ASM} iterations by using a warm-start. Specifically, for each task, the proposed algorithm initializes the parameters in~\eqref{eq:sgdrule} with the solution obtained for the closest task (see Alg.~\ref{alg:upmu} in Appendix~\ref{sec:lear}).

\subsection{Batch learning scenarios}~\label{sec:batch_learning_scenario}

\begin{algorithm}
\captionsetup{labelfont={bf}}
\caption{\ac{MDA}}
\label{alg:transfer_learning}
\begin{algorithmic}
\State \textbf{Input:} \hspace{0.3cm}$D_1, D_2, \ldots, D_{k}$
\State \textbf{Output:}\hspace{0.18cm}$\B{\mu}_k^k, R(\mathcal{U}_k^k)$ if $|D_k| > 0$ and $\B{\mu}_k^{k-1}, R(\mathcal{U}_k^{k-1})$ if $|D_k| = 0$.
 \For{$j = 1, 2, ..., k-1$}
\State Obtain forward mean and \ac{MSE} vectors $\B{\tau}_{j}^{j}, \B{s}_{j}^{j}$ as in \eqref{eq:tau_general}-\eqref{eq:s_general}
\EndFor
\If{$|D_k| > 0$}
\State Obtain mean and \ac{MSE} vectors $\B{\tau}_{k}^{k}, \B{s}_{k}^{k}$ as in \eqref{eq:tau_general}-\eqref{eq:s_general}
\State Take confidence vector $\B{\lambda}_k^k$ as $\lambda_{0} \sqrt{\B{s}_k^k}$
 \State Obtain classifier parameter and minimax risk $\B{\mu}_k^k$ and $R(\mathcal{U}_k^k)$ solving~\eqref{eq:mrc}
  \Else
\State Obtain mean and \ac{MSE} vectors $\B{\tau}_{k}^{k-1}, \B{s}_{k}^{k-1}$ as in \eqref{eq:concept_drift}
\State Take confidence vector $\B{\lambda}_k^{k-1}$ as $\lambda_{0} \sqrt{\B{s}_k^{k-1}}$
 \State Obtain classifier parameter and minimax risk $\B{\mu}_k^{k-1}$ and $R(\mathcal{U}_k^{k-1})$ solving~\eqref{eq:mrc}
 \EndIf
 \end{algorithmic}
\end{algorithm}

In \ac{MDA} information from preceding tasks can be used to improve the performance of the last task. Specifically, \ac{MDA} methods use $k$ sample sets corresponding with different source tasks to obtain a classification rule for the $k$-th target task with $|D_k| > 0$  \citep{Reeve2021, Tripuraneni2020} or $|D_k| = 0$ \citep{Stojanov2020, Li2022}. Algorithm~\ref{alg:transfer_learning} details the implementation of the proposed methodology applied to \ac{MDA} that first obtains forward mean and \ac{MSE} vectors $\B{\tau}_{j}^{j}$ and $\B{s}_{j}^{j}$ as in~\eqref{eq:tau_general}-\eqref{eq:s_general} for $j = 1, 2, \ldots, k-1$. Then, we obtain mean and \ac{MSE} vectors $\B{\tau}_k^k$ and $\B{s}_k^k$ as in~\eqref{eq:tau_general}-\eqref{eq:s_general} if $|D_k| > 0$ and mean and \ac{MSE} vectors $\B{\tau}_k^{k-1}$ and $\B{s}_k^{k-1}$ as in~\eqref{eq:concept_drift} if $|D_k| = 0$. We take the confidence vector as in~\eqref{eq:tau1} and obtain the classifier parameter and the minimax risk for the $k$-th task solving~\eqref{eq:mrc} (see Alg.~\ref{alg:upmu} in Appendix~\ref{sec:lear}). Algorithm~\ref{alg:transfer_learning} has computational complexity $O(km + n 2^{|\set{Y}|} Km)$ and memory complexity $O(m)$ where $m$ is the length of the feature mapping, $n$ is the sample size, and $K$ is the number of iterations of the optimization step.

In \ac{MTL}, at each time step, information from preceding tasks can be used to improve the performance of the last task and, reciprocally, the information from the last task can be used to improve the performance of the preceding tasks. Specifically, \ac{MTL} methods use $k$ sample sets corresponding with different tasks to obtain classification rules $\up{h}_1^k, \up{h}_2^k, \ldots, \up{h}_k^k$ for the $k$ tasks  \citep{Zhang2018, Lin2020}. Algorithm~\ref{alg:multitask_learning} details the implementation of the proposed methodology applied to \ac{MTL} that first obtains forward mean and \ac{MSE} vectors $\B{\tau}_1^1, \B{\tau}_2^2, \ldots, \B{\tau}_k^k$ and $\B{s}_1^{1}, \B{s}_2^{2}, \ldots, \B{s}_k^{k}$ as in~\eqref{eq:tau_general}-\eqref{eq:s_general}. Then, we obtain forward and backward mean and~\ac{MSE} vectors $\B{\tau}_1^k, \B{\tau}_2^k, \ldots, \B{\tau}_k^k$ and $\B{s}_1^{k}, \B{s}_2^{k}, \ldots, \B{s}_k^{k}$ as in~\eqref{eq:tau_general_back}-\eqref{eq:s_general_back}, take the confidence vectors $\B{\lambda}_1^{k}, \B{\lambda}_2^{k}, \ldots, \B{\lambda}_k^{k}$ as in~\eqref{eq:tau1}, and obtain the classifier parameters $\B{\mu}_1^{k}, \B{\mu}_2^{k}, \ldots, \B{\mu}_k^{k}$ and the minimax risks $R(\set{U}_1^{k}), R(\set{U}_2^{k}), \ldots, R(\set{U}_k^{k})$ for the $k$ tasks in the sequence solving~\eqref{eq:mrc} (see Alg.~\ref{alg:upmu} in Appendix~\ref{sec:lear}). Algorithm~\ref{alg:multitask_learning} has computational complexity $\set{O}(mk + n2^{|\set{Y}|} Kmk)$ and memory complexity $\set{O}(mk + k)$.
\begin{algorithm}
\captionsetup{labelfont={bf}}
\caption{\ac{MTL}}
\label{alg:multitask_learning}
\begin{algorithmic}
\State \textbf{Input:} \hspace{0.3cm}$D_1, D_2, \ldots, D_{k}$
\State \textbf{Output:}\hspace{0.18cm}$\{\B{\mu}_j^{k}\}_{1 \leq j \leq k}$ and $\{R(\mathcal{U}_j^{k})\}_{1 \leq j \leq k}$
\For{$j = 1, 2, ..., k$}
\State Obtain forward mean and \ac{MSE} vectors $\B{\tau}_{j}^{j}, \B{s}_{j}^{j}$ as in \eqref{eq:tau_general}-\eqref{eq:s_general}
\EndFor
\For{$j = k-1, k-2, ..., 1$}
\State Obtain forward and backward mean and \ac{MSE} vectors $\B{\tau}_{j}^{k}, \B{s}_{j}^{k}$ as in \eqref{eq:tau_general_back}-\eqref{eq:s_general_back}
\State Take confidence vector $\B{\lambda}_j^{k}$ as $\lambda_{0} \sqrt{\B{s}_{j}^{k}}$
 \State Obtain classifier parameter and minimax risk $\B{\mu}_j^{k}$ and $R(\mathcal{U}_j^{k})$ solving~\eqref{eq:mrc}
\EndFor
\end{algorithmic}
\end{algorithm}

\subsection{Online learning scenarios}~\label{sec:online_learning_scenario}

In \ac{SCD}, sample sets corresponding with different tasks arrive over time and at each time step, information from preceding tasks can be used to improve the performance of the last task. \ac{SCD} methods use for each $k$-th task the most recent sample set $D_{k-1}$ and information retained from preceding tasks to obtain the classification rule $\up{h}_{k}^{k-1}$ \citep{elwell2011incremental, brzezinski2013reacting}. 
Algorithm~\ref{alg:amrc} details the implementation of the proposed methodology applied to \ac{SCD} that first obtains forward mean and \ac{MSE} vectors $\B{\tau}_{k-1}^{k-1}$ and $\B{s}_{k-1}^{k-1}$ as in~\eqref{eq:tau_general}-\eqref{eq:s_general} from those for the preceding task $\B{\tau}_{k-2}^{k-2}$ and $\B{s}_{k-2}^{k-2}$ together with sample set $D_{k-1}$. Then, we obtain mean and \ac{MSE} vectors $\B{\tau}_{k}^{k-1}$ and $\B{s}_{k}^{k-1}$ as in~\eqref{eq:concept_drift}. We take the confidence vector $\B{\lambda}_k^{k-1}$ as in~\eqref{eq:tau1} and obtain the classifier parameter $\B{\mu}_k^{k-1}$ and the minimax risk $R(\set{U}_k^{k-1})$ for the $k$-th task solving~\eqref{eq:mrc} (see Alg.~\ref{alg:upmu} in Appendix~\ref{sec:lear}).  Algorithm~\ref{alg:amrc} has at each step $k$ computational complexity $O(m+n 2^{|\set{Y}|} Km)$ and memory complexity $O(m)$.
\begin{algorithm}
\captionsetup{labelfont={bf}}
\caption{\ac{SCD} at step $k$}
\label{alg:amrc}
\begin{algorithmic}
\State \textbf{Input:} \hspace{0.3cm}$D_{k-1}$, $\B{\tau}_{k-2}^{k-2}, \B{s}_{k-2}^{k-2}$, and $\B{\mu}_{k-1}^{k-2}$
\State \textbf{Output:}\hspace{0.18cm}$\B{\tau}_{k-1}^{k-1}, \B{s}_{k-1}^{k-1}, \B{\mu}_k^{k-1}$, and $R(\mathcal{U}_k^{k-1})$
\State Obtain forward mean and \ac{MSE} vectors $\B{\tau}_{k-1}^{k-1}, \B{s}_{k-1}^{k-1}$ as in \eqref{eq:tau_general}-\eqref{eq:s_general}
\State Obtain mean and \ac{MSE} vectors $\B{\tau}_k^{k-1}, \B{s}_k^{k-1}$ as in \eqref{eq:concept_drift}
\State Take confidence vector $\B{\lambda}_k^{k-1}$ as $\lambda_{0} \sqrt{\B{s}_k^{k-1}}$
 \State Obtain classifier parameter and minimax risk $\B{\mu}_k^{k-1}$ and $R(\mathcal{U}_k^{k-1})$ solving~\eqref{eq:mrc}
\end{algorithmic}
\end{algorithm}

In \ac{CL}, sample sets corresponding with different tasks arrive over time and at each time step, information from preceding tasks can be used to improve the performance of the last task and, reciprocally, the information from the last task can be used to improve the performance of the preceding tasks. \ac{CL} methods use at each step $k$ the sample set $D_{k}$ and information retained from preceding tasks to obtain classification rules $\up{h}_1^k, \up{h}_2^k, \ldots, \up{h}_k^k$ for the sequence of tasks \citep{henning2021posterior, hurtado2021optimizing}. Algorithm~\ref{alg:continual_learning} details the implementation of the proposed methodology applied \ac{CL} that first obtains at each step $k$ forward mean and \ac{MSE} vectors $\B{\tau}_{k}^{k}$ and $\B{s}_{k}^{k}$ as in~\eqref{eq:tau_general}-\eqref{eq:s_general}. Then, we obtain forward and backward mean and \ac{MSE} vectors  $\B{\tau}_{k-b}^{k}, \B{\tau}_{k-b+1}^{k}, \ldots, \B{\tau}_{k-1}^{k}$ and $\B{s}_{k-b}^{k}, \B{s}_{k-b+1}^{k}, \ldots, \B{s}_{k-1}^{k}$ as in~\eqref{eq:tau_general_back}-\eqref{eq:s_general_back} for $b = k-j$ backward steps. In particular, if $b = 0$, the proposed methodology carries out only forward learning. Then, we take confidence vectors $\B{\lambda}_{k-b}^k, \B{\lambda}_{k-b+1}^k, \ldots, \B{\lambda}_{k}^k$ as in~\eqref{eq:tau1} and obtain classifier parameters $\B{\mu}_{k-b}^k, \B{\mu}_{k-b+1}^k, \ldots, \B{\mu}_{k}^k$ and minimax risks $R(\set{U}_{k-b}^k), R(\set{U}_{k-b+1}^k), \ldots, R(\set{U}_{k}^k)$ solving~\eqref{eq:mrc} (see Alg.~\ref{alg:upmu} in Appendix~\ref{sec:lear}). Algorithm~\ref{alg:continual_learning} has at each step $k$ computational complexity $\set{O}((b + 1)mk + b n2^{|\set{Y}|}Km)$ and memory complexity $\set{O}((b + k)m + n2^{|\set{Y}|} m b)$. The number of backward steps $b = k-j$ can be taken to be rather small since the benefits of learning from succeeding tasks are achieved using only $b = 3$ backward steps in most of the situations (see Fig.~\ref{fig:ess_succeeding} in Section~\ref{sec:ess}). The proposed techniques can be extended to situations in which a new sample set can correspond with a precedingly learned task by updating the sample average and the \ac{MSE} vector in~\eqref{eq:tau1} with the new sample set (see Appendix~\ref{app:cl_extension}). 

\begin{algorithm}
\caption{\ac{CL} at step $k$}
\label{alg:continual_learning}
\begin{algorithmic}
\State \textbf{Input:} \hspace{0.2cm}$D_k, \B{\tau}_{j}^{j}, \B{s}_{j}^{j}, \B{\mu}_j^j$ for $k-b \leq j < k$
\State \textbf{Output:} $\B{\mu}_j^k$ for $k-b+1 \leq j \leq k$, $\B{\tau}_k^{k}, \B{s}_k^k$ 
\State Obtain forward mean and \ac{MSE} vectors $\B{\tau}_k^k, \B{s}_k^k$ using \eqref{eq:tau_general}-\eqref{eq:s_general} 
\State Take confidence vector $\B{\lambda}_k^k = \lambda_{0} \sqrt{\B{s}_k^k}$ 
\State Obtain classifier parameter $\B{\mu}_k^k$ and minimax risk $R(\set{U}_k^k)$ solving~\eqref{eq:mrc} 
\For{$j = k-1, k-2, \ldots, k-b$} 
\State Obtain forward and backward mean and \ac{MSE} vectors $\B{\tau}_j^{k}, \B{s}_j^{k}$ using \eqref{eq:tau_general_back}-\eqref{eq:s_general_back}
\State Take confidence vector $\B{\lambda}_j^{k}$ as $\lambda_{0} \sqrt{\B{s}_j^{k}}$
\State Obtain classifier parameter $\B{\mu}_j^k$ and minimax risk $R(\set{U}_j^k)$ solving~\eqref{eq:mrc} 
\EndFor
\end{algorithmic}
\end{algorithm}

 \section{Extension to higher-order dependences}\label{subsec:time_increment}
  
 This section describes algorithms that account for higher-order dependences among tasks. Such algorithms
utilize methods that are commonly used for target tracking to describe target trajectories using kinematic
models (see e.g., \citet{bar2004estimation}). 

For each $j$-th task, each component of the mean vector $\B{\tau}_j^\infty = \mathbb{E}_{\up{p}_{j}} \{\Phi(x, y)\} \in \mathbb{R}^m$ is estimated leveraging information from $k$ tasks and accounting for high-order dependences between consecutive tasks. Let ${{\tau}_j^\infty}^{(i)}$ denote the $i$-th component of the mean vector for $i = 1, 2, \ldots, m$. We assume that ${{\tau}_{j}^\infty}^{(i)}$ is $p$ times differentiable with respect to time and denote by $\B{\gamma}_{j, i}^\infty \in \mathbb{R}^{p+1}$ the vector composed by the $i$-th component of the mean vector ${\tau_{j}^\infty}^{(i)}$ and its successive derivatives up to order $p$. If $p = 0$, the state vector $\gamma_{j, i}^{\infty}$ is the mean vector component ${\tau_{j}^{\infty}}^{(i)}$ and the recursions below coincide with the recursions in Section~\ref{sec:evolution}.
 
 As is usually done for target tracking \citet{bar2004estimation}, we model the evolution of the state vector $\B{\gamma}_{j, i}^\infty$ using the partially-observed linear dynamical system 
\begin{equation}
\begin{split}
\label{eq:kalman}
\B{\gamma}_{j, i}^\infty & = \V{T}_j \B{\gamma}_{j-1, i}^\infty + \B{\overline{w}}_{j, i}\\
\tau_{j}^{(i)} & = {{\tau}_{j}^\infty}^{(i)} + v_{j, i}
\end{split}
\end{equation}
with transition matrix $\V{T}_j = \V{I} + \sum_{s=1}^p {\Delta_j^s}\V{U}_s/s!$ where $\V{U}_s$ is the $(p+1)\times(p+1)$ matrix with ones on the $s$-th upper diagonal and zeros in the rest of components, $\Delta_j$ is the time increment between the $j$-th and the $(j-1)$-th tasks, $i = 1, 2, \ldots, m$, and $j = 1, 2, \ldots$. 
The variables $\B{\overline{w}}_{j, i}$ and $v_{j, i}$ represent uncorrelated zero-mean noise processes with variance $\V{D}_{j, i} =  \B{g}_j\B{g}_j^\top \bar{d_{j}}^{(i)}$ and $s_j^{(i)} = {\sigma_{j}^2}^{(i)}/n_j$, respectively, where $\B{g}_{j}$ is given by $\B{g}_{j} = [\Delta_j^{p+1}/(p+1)!, \Delta_j^{p}/p!, \ldots, \Delta_{j}]^\top$. Variances $\V{D}_{j, i}$ and $s_{j}^{(i)}$ can be estimated online using methods such as those proposed in~\citet{odelson2006new, akhlaghi2017adaptive}. Dynamical systems as that given by~\eqref{eq:kalman} are known in target tracking as kinematic state models and can be derived using the $p$-th order Taylor expansion of ${{\tau}_{j}^\infty}^{(i)}$.

Analogously to Section~\ref{sec:evolution}, for a sequence of $k$ tasks, the proposed techniques first obtain for each $j$-th task with $j = \{1, 2, \ldots, k\}$ forward state vector and \ac{MSE} matrix $\B{\gamma}_{j, i}^{j}, \B{\Sigma}_{j, i}^{j}$ as
\begin{alignat}{5}
\label{eq:gamma_c}
\B{\gamma}_{j, i}^j & = \V{T}_j \B{\gamma}_{j-1, i}^{j-1} + \B{\eta}_{j, i}^j \left({\tau}_j^{(i)} - \V{e}_1^\top \V{T}_j \B{\gamma}_{j-1, i}^{j-1}\right)\\
\label{eq:Sigma_c}
\B{\Sigma}_{j, i}^j & = \left(\V{I} - \B{\eta}_{j, i}^j \V{e}_1^\top\right) \left(\V{T}_j \B{\Sigma}_{j-1, i}^{j-1} \V{T}_j^\top + \V{D}_{j, i}\right)\\
\label{eq:gain_c}
\B{\eta}_{j, i}^j & =  \frac{\left(\V{T}_j \B{\Sigma}_{j-1, i}^{j-1} \V{T}_j^\top + \V{D}_{j, i}\right)\V{e}_1}{\V{e}_1^{\top} \left(\V{T}_j \B{\Sigma}_{j-1, i}^{j-1} \V{T}_j^\top + \V{D}_{j, i}\right) \V{e}_1 + s_j^{(i)}}
\end{alignat}
for $j = 2, 3, \ldots, k$ and $i = 1, 2, \ldots, m$ with $\B{\tau}_j, \B{s}_j$ given by~\eqref{eq:tau1}, $\B{\gamma}_{1, i}^1 = \tau_1^{(i)} \V{e}_{1}$, and \mbox{$\B{\Sigma}_{1, i}^1 = s_1^{(i)} \V{e}_{1} \V{e}_{1}^{\top}$}. State vector and \ac{MSE} matrix $\B{\gamma}_{j, i}^{j}, \B{\Sigma}_{j, i}^{j}$ are obtained leveraging information up to the $j$-th task from sample sets $D_1, D_2, \ldots, D_{j}$. Specifically, for each $j$-th task, those vectors and matrices are obtained by acquiring information from sample set $D_{j}$ through ${\tau}_{j}^{(i)}, {s}_{j}^{(i)}$ and retaining information from sample sets $D_1, D_2, \ldots, D_{j-1}$ through state vector and \ac{MSE} matrix $\B{\gamma}_{j-1, i}^{j-1}, \B{\Sigma}_{j-1, i}^{j-1}$. 

The proposed techniques obtain for each $j$-th task with \mbox{$j = \{1, 2, \ldots, k\}$} forward and backward state vector and \ac{MSE} matrix leveraging information from the $k$ tasks in the sequence. Let $\B{\gamma}_{j, i}^{k}$ and $\B{\Sigma}_{j, i}^{k}$ denote the state vector and \ac{MSE} matrix for forward and backward learning corresponding to the $j$-th task for \mbox{$j = \{1, 2, \ldots, k\}$} and the $i$-th component. The following recursions allow us to obtain $\B{\gamma}_{j ,i}^{k}, \B{\Sigma}_{j, i}^{k}$ using forward state vector and \ac{MSE} matrix $\B{\gamma}_{j, i}^{j}, \B{\Sigma}_{j, i}^{j}$ as 
\begin{align}
\label{eq:gamma_back}
      \B{\gamma}_{j, i}^{k}& =  \B{\gamma}_{j, i}^{j} + \V{H}_{j, i}^{k} \left( \B{\gamma}_{j+1, i}^{k}  - \B{\gamma}_{j, i}^j\right)\\
      \label{eq:Sigma_back}
       \B{\Sigma}_{j, i}^{k} & = \B{\Sigma}_{j, i}^{j} + \V{H}_{j, i}^{k}(\B{\Sigma}_{j+1, i}^{k} - \V{T}_{j+1} \B{\Sigma}_{j, i}^{j}\V{T}_{j+1}^\top - \V{D}_{j+1, i}){\V{H}_{j, i}^{k}}^\top\\
           \label{eq:H_back}
 \V{H}_{j, i}^{k} & =  \B{\Sigma}_{j, i}^{j}\V{T}_j^\top\left(\V{T}_{j+1} \B{\Sigma}_{j, i}^{j}\V{T}_{j+1}^\top + \V{D}_{j+1, i}\right)^{-1}
       \end{align}
for $j = 1, 2, \ldots k-1$ and $i = 1, 2, \ldots, m$ with $\B{\gamma}_{j, i}^j$, $\B{\Sigma}_{j, i}^j$ given by~\eqref{eq:gamma_c}-\eqref{eq:Sigma_c}. State vectors and \ac{MSE} matrices $\B{\gamma}_{j, i}^k, \B{\Sigma}_{j, i}^k$ are obtained leveraging information from sample sets $D_1, D_2, \ldots, D_k$. Specifically, for each $j$-th task and $i = 1, 2, \ldots, m$, state vectors and \ac{MSE} matrices are obtained acquiring information from sample sets $D_{j+1}, D_{j+2}, \ldots, D_k$ through state vector and \ac{MSE} matrix $\B{\gamma}_{j+1, i}^k, \B{\Sigma}_{j+1, i}^k$ and retaining information from sample sets $D_1, D_2, \ldots, D_j$ through state vector and \ac{MSE} matrix $\B{\gamma}_{j, i}^j, \B{\Sigma}_{j, i}^j$.
       
The above recursions allow us to obtain state vectors as well as the \ac{MSE} matrices for each task accounting for high-order dependences among tasks, accounting for multidimensional tasks' changes, and leveraging information from all tasks in the sequence. Such recursions adapt to multidimensional tasks' changes by accounting for the change between consecutive mean vector components. Specifically, for each $j$-th task and each $i$-th component with $i = 1, 2, \ldots, m$, the $i$-th component of the mean vector estimate is updated by using the corresponding component of the expected quadratic change, the estimate for the consecutive task, and the most recent sample average. 

The result below shows that the state vectors obtained above are the estimates of the state vectors with minimum \ac{MSE}.
       
       \begin{teorema}\label{th:high_order}
       Let ${\sigma_j^2}^{(i)}$ and $\bar{d_j}^{(i)}$ in recursions~\eqref{eq:gamma_c}-\eqref{eq:H_back} be ${\sigma_j^2}^{(i)} =\mathbb{V}\text{ar}_{\up{p}_j}\{\Phi\left(x, y\right)^{(i)}\}$ and $\bar{d_j}^{(i)} = {\tau_{j}^{\infty}}^{(p+1)}$. If the tasks' distributions satisfy the dynamical system in~\eqref{eq:kalman}, then we have that
\begin{enumerate}
\item $\B{\gamma}_{j, i}^{j}$ given by~\eqref{eq:gamma_c} is the unbiased linear estimator of the state vector ${\B{\gamma}}_{j, i}^\infty$ based on $D_1, D_2, \ldots, D_{j}$ that has the minimum \ac{MSE} and $\B{\Sigma}_{j, i}^{j}$ given by~\eqref{eq:Sigma_c} is its \ac{MSE}. 
\item $\B{\gamma}_{j, i}^{k}$ given by~\eqref{eq:gamma_back} is the unbiased linear estimator of the state vector ${\B{\gamma}}_{j, i}^\infty$ based on $D_1, D_2, \ldots, D_k$ that has the minimum \ac{MSE} and $\B{\Sigma}_{j, i}^{k}$ given by~\eqref{eq:Sigma_back} is its \ac{MSE}.
\end{enumerate}
\begin{proof}
See Appendix~\ref{app:rec}.
\end{proof}
\end{teorema}
The above theorem shows that equations in~\eqref{eq:gamma_c}-\eqref{eq:H_back} enable to recursively obtain, for each $j$-th task, the state vector estimate as well as its \ac{MSE} matrix leveraging all the information at each step.

The proposed methodology extended to higher-order dependences can be applied to the supervised learning scenarios described in Sections~\ref{sec:batch_learning_scenario} and~\ref{sec:online_learning_scenario} above. The application of the extension for high-order time dependences in these scenarios is similar to the implementation described in Algorithms~\ref{alg:transfer_learning}-\ref{alg:continual_learning} by changing the estimate of mean and \ac{MSE} vectors to the estimate of state vector and \ac{MSE} matrix.

\section{Numerical results} \label{sec:nr}

This section evaluates the performance of the proposed methods in comparison with the presented performance guarantees and the state-of-the-art. We show the reliability of the performance guarantees; show the improvement of the multidimensional adaptation; evaluate the performance in batch and in online learning scenarios; and quantify the performance of the extension presented. The methods presented can be implemented using MRCpy library~\citep{bondugula2021mrcpy} and the specific code used in the experimental results is provided on the web \url{https://github.com/MachineLearningBCAM/Supervised-learning-evolving-task-JMLR-2025}.

We utilize $13$ public datasets that have been often used as benchmark for tasks that are in a sequence (see Table~\ref{tab:datasets} in Appendix~\ref{app:res}). The tabular datasets (7) are divided in segments of $300$ samples corresponding to consecutive times where each task corresponds to each of those segments; while the rest of the datasets (6) are composed by evolving tasks (images with characteristics/quality/realism that change over time). The samples in each task are randomly splitted in $100$ samples for test and the rest of samples for training. In each repetition, the samples used for training are randomly sampled from the pool of training samples for each task.

    \begin{figure}
         \centering
                   \begin{subfigure}[t]{0.48\textwidth}
         \centering
                  \psfrag{Errorabcdefghijklmnopqrstuvwxyzabcdefghijklmnopqrstuvwxyzabcde}[l][l][0.6]{Error probability $R(\up{h}_j^{j-1})$}
                  \psfrag{Bound}[l][l][0.6]{Our bound $R(\set{U}_j^{j-1})$}
\psfrag{Bound2}[l][l][0.6]{Our bound $R(\set{U}_j^{j-1})+ \left(|\B{\tau}_k^\infty - \B{\tau}_j^{j-1}| - \B{\lambda}_j^{j-1}\right)^{\top} \left|\B{\mu}_j^{j-1}\right|$}
                  \psfrag{Task}[][][0.6]{Task $j$}
                  \psfrag{Error probability}[][][0.6]{Error probability}
                  \psfrag{AMRC}[l][l][0.6]{Error probability $R(\up{h}_j^{j-1})$}
                     \psfrag{10}[][][0.5]{$10$}
                  \psfrag{30}[][][0.5]{$30$}
                  \psfrag{50}[][][0.5]{$50$}
                  \psfrag{70}[][][0.5]{$70$}
                  \psfrag{90}[][][0.5]{$90$}
                  \psfrag{0.45}[][][0.5]{$0.45$}
                  \psfrag{0.4}[][][0.5]{$0.4$}
                  \psfrag{0.3}[][][0.5]{$0.3$}
                  \psfrag{0.35}[][][0.5]{$0.35$}
                  \psfrag{0.1}[][][0.5]{$0.1$}
         \includegraphics[width=\textwidth]{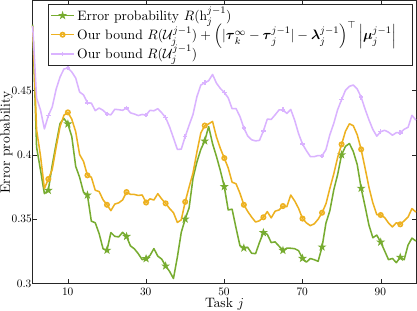}
                                \caption{Bounds for error probabilities of each $j$-th task leveraging information from the $j-1$ preceding tasks.}   
                                \label{fig:error_probability_10}
\end{subfigure}
\hfill
                   \begin{subfigure}[t]{0.48\textwidth}
         \centering
                  \psfrag{Errorabcdefghijklmnopqrstuvwxyzabcdefghijklmnopqrstuv}[l][l][0.6]{Error probability $R(\up{h}_j^{k})$}
                  \psfrag{Bound}[l][l][0.6]{Our bound $R(\set{U}_j^{k})$}
                                     \psfrag{Bound2}[l][l][0.6]{Our bound $R(\set{U}_j^{k})+ \left(|\B{\tau}_j^\infty - \B{\tau}_j^{k}| - \B{\lambda}_j^{k}\right)^{\top} \left|\B{\mu}_j^{k}\right|$}
                  \psfrag{Task}[][][0.6]{Task $j$}
                  \psfrag{Error probability}[][][0.6]{Error probability}
                  \psfrag{LMRC}[l][l][0.6]{LMRC $R(\up{h}_j^{k})$}
                  \psfrag{10}[][][0.5]{$10$}
                  \psfrag{30}[][][0.5]{$30$}
                  \psfrag{50}[][][0.5]{$50$}
                  \psfrag{70}[][][0.5]{$70$}
                  \psfrag{90}[][][0.5]{$90$}
                  \psfrag{0.5}[][][0.5]{$0.5$}
                  \psfrag{0.2}[][][0.5]{$0.2$}
                  \psfrag{0.35}[][][0.5]{$0.35$}
                  \psfrag{0.3}[][][0.5]{$0.3$}
                  \psfrag{0.4}[][][0.5]{$0.4$}
                  \psfrag{0.1}[][][0.5]{$0.1$}
                  \psfrag{0}[][][0.5]{$0$}
         \includegraphics[width=\textwidth]{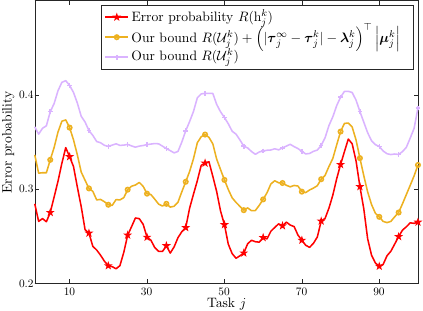}
                                \caption{Bounds for error probabilities of each $j$-th task leveraging information from the $k$ tasks.}   
                        \label{fig:error_probability_continual}
\end{subfigure}
              \caption{Results on synthetic data show the evolution over tasks of performance bounds and error probabilities.}
                       \label{fig:Bounds_100}
\end{figure}

The choice of the feature mapping has a significant impact on the final performance. The results for the proposed methods are obtained using a feature mapping defined by multiple features over instances together with one-hot encoding of labels as described in~\eqref{eq:feature_mapping}. The map $\Psi$ in~\eqref{eq:feature_mapping} that represents instances as real vectors is given by the pixel values for the ``Rotated MNIST'' dataset, by the last layer of the ResNet18 pre-trained network for the rest of image datasets, and by \acp{RFF} \citep{shen:2019} with $200$ Gaussian weights and covariance matrix given by $\sigma^{2} = 10$ for the tabular datasets. 
The confidence vector $\B{\lambda}$ in equation~\eqref{eq:tau1} is obtained with $\lambda_0 = 0.7$, vector $\B{\sigma}_j^2$ in equation~\eqref{eq:tau1} is given by the variance of $n_j$ samples, vector $\B{d}_j$ in equation~\eqref{eq:d} is estimated using $W = 2$, variance $\B{\overline{d}}_{j}$ of the noise process $\B{\overline{w}}_{j}$ in~\eqref{eq:kalman} is estimated using the recursive approach presented in \citet{akhlaghi2017adaptive}; and the proposed methodology applied to CL in Section~\ref{sec:scenarios} is implemented using $b = 3$ backward steps. 
 Value for hyper-parameter $\lambda_0$ can be selected by methods such as cross-validation over a grid of possible values. For simplicity, we select the value by inspection over the synthetic data. We use the same value for all the numerical results for fair comparison with the state-of-the-art and to show that the methodology presented does not heavily rely on its value.

\subsection{Tightness of the performance guarantees}
  
In this section, we show the tightness of the presented bounds for error probabilities using the synthetic data since the error probability cannot be computed using real datasets. These numerical results are obtained averaging the classification errors and the bounds achieved with $10000$ random instantiations of data samples in the synthetic data. Such data comprises a rotating hyperplane in $2$ dimensional space where each coefficient of the hyperplane rotates $5$ degrees between consecutive tasks. 

Figures~\ref{fig:error_probability_10} and~\ref{fig:error_probability_continual} show the averaged bounds for error probabilities corresponding to inequality~\eqref{eq:prob_error_continual} and the minimax risk in comparison with the true error probabilities using $n = 10$ samples per task. Figure~\ref{fig:error_probability_10} shows bounds for error probabilities of each $j$-th task  $R(\up{h}_j^{j-1})$ obtained leveraging information from sample sets $D_1, D_2, \ldots, D_{j-1}$ (forward learning) and Figure~\ref{fig:error_probability_continual} shows bounds for error probabilities of each $j$-th task $R(\up{h}_j^k)$ obtained leveraging information from sample sets $D_1, D_2, \ldots, D_k$ (forward and backward learning). Such figures show that the bounds $R(\set{U}_j^{j-1})$ and $R(\set{U}_j^{k})$ can offer, for each $j$-th task, tight upper bounds for error probabilities $R(\up{h}_j^{j-1})$ and $R(\up{h}_j^k)$, respectively. 

   \begin{figure}
         \centering
                   \begin{subfigure}[t]{0.48\textwidth}
         \centering
                  \psfrag{data1}[l][l][0.6]{MRCs multi. changes}
                   \psfrag{Iguales}[l][l][0.6]{MRCs uni. changes}
                  \psfrag{Condor}[l][l][0.6]{Condor multi. changes}
\psfrag{Condor igualesabcdefghijklmn}[l][l][0.6]{Condor uni. changes}
                  \psfrag{Classification error}[][][0.6]{Classification error}
                  \psfrag{Variance}[][][0.6]{Variance of the change between hyperplanes $\sigma_w^2$}
                  \psfrag{AMRC}[l][l][0.6]{$R(\up{h}_j^{j-1})$}
                     \psfrag{1}[][][0.5]{$0$}
                  \psfrag{3}[][][0.5]{$0.1$}
                  \psfrag{5}[][][0.5]{$0.2$}
                  \psfrag{7}[][][0.5]{$0.3$}
                  \psfrag{90}[][][0.5]{$90$}
                  \psfrag{0.5}[][][0.5]{$0.5$}
                  \psfrag{0.4}[][][0.5]{$0.4$}
                  \psfrag{0.3}[][][0.5]{$0.3$}
                  \psfrag{0.2}[][][0.5]{$0.2$}
                  \psfrag{0.1}[][][0.5]{$0.1$}
         \includegraphics[width=\textwidth]{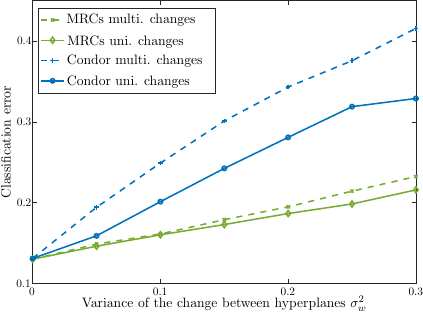}
                                \caption{Classification error leveraging information from the $j-1$ preceding tasks.}   
                                \label{fig:error_multidimensional_n100}
\end{subfigure}
\hfill
                   \begin{subfigure}[t]{0.48\textwidth}
         \centering
                  \psfrag{MRCs}[l][l][0.6]{MRCs multi. changes}
                  \psfrag{MRCs mismo cambio}[l][l][0.6]{MRCs uni. changes}
                                     \psfrag{GEM}[l][l][0.6]{GEM multi. changes}
                                     \psfrag{GEM mismo cambioabcdefgh}[l][l][0.6]{GEM uni. changes}
                  \psfrag{Task}[l][l][0.6]{Tasks $j$}
                  \psfrag{Classification error}[][][0.6]{Classification error}
                              \psfrag{Change between tasks}[][][0.6]{Variance of the change between hyperplanes $\sigma^2$}
                  \psfrag{n = 100}[l][l][0.6]{MRCs}
                     \psfrag{1}[][][0.5]{$0$}
                  \psfrag{3}[][][0.5]{$0.1$}
                  \psfrag{5}[][][0.5]{$0.2$}
                  \psfrag{7}[][][0.5]{$0.3$}
                  \psfrag{0.5}[][][0.5]{$0.5$}
                  \psfrag{0.4}[][][0.5]{$0.4$}
                  \psfrag{0.3}[][][0.5]{$0.3$}
                  \psfrag{0.2}[][][0.5]{$0.2$}
                  \psfrag{0.1}[][][0.5]{$0.1$}
                  \psfrag{0}[][][0.5]{$0$}
         \includegraphics[width=\textwidth]{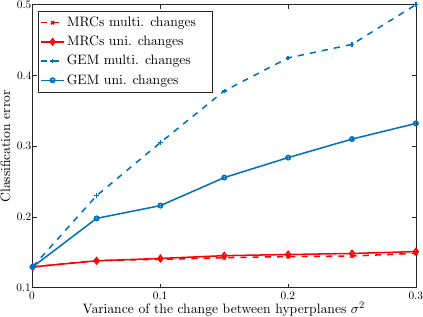}
                                \caption{Classification error leveraging information from the $k$ tasks.}   
                                \label{fig:multidimensional_continual}
\end{subfigure}
              \caption{Results on synthetic data show the multidimensional adaptation to tasks' changes of the proposed methodology.}
                                     \label{fig:Multidimensional_100}
\end{figure}

\subsection{Improvement of multidimensional adaptation}

This section shows the improvement of multidimensional adaptation using the synthetic data since the tasks' changes cannot be modified in real datasets. These numerical results are obtained averaging the classification errors achieved with $100$ random instantiations of data samples in the synthetic data with $n = 100$ samples per task. Such data comprises a rotating hyperplane in $5$ dimensional space where each coefficient of the hyperplane changes between consecutive tasks. Specifically, for each $j$-th task, we obtain the coefficients of the hyperplane $\V{w}_j$ by adding a Gaussian random variable to the coefficients of the preceding task as $\V{w}_j = \V{w}_{j-1} + N(\V{0}, \sigma_w^2 \V{I})$ for multidimensional (multi.) changes and as \mbox{$\V{w}_j = \V{w}_{j-1} + \V{1} N(0, \sigma_w^2)$} for unidimensional (uni.) changes where $\sigma_w^2$ denotes the variance of the change between consecutive hyperplanes. The results of the proposed methodology are compared with state-of-the-art techniques: Condor \citep{zhao2020handling} that leverages information from preceding tasks and \ac{GEM} \citep{lopez2017gradient} that leverages information from all the tasks in the sequence. 

 \begin{figure}
         \centering
 \begin{subfigure}[t]{0.48\textwidth}
         \centering
                  \psfrag{k = 3}[l][l][0.6]{Proposed $k = 3$}
                  \psfrag{k = 7}[l][l][0.6]{Proposed $k = 7$}
                   \psfrag{Singleabcdefghijklmnopq}[l][l][0.6]{Single-task}
                  \psfrag{All k = 3}[l][l][0.6]{Joint $k = 3$}
                  \psfrag{All k = 7}[l][l][0.6]{Joint $k = 7$}
                  \psfrag{Classification error}[][][0.6]{Classification error}
                  \psfrag{Batch size}[][][0.6]{Sample size $n$}
                  \psfrag{30}[][][0.5]{$30$}
                  \psfrag{50}[][][0.5]{$50$}
                  \psfrag{10}[][][0.5]{$10$}
                  \psfrag{70}[][][0.5]{$70$}
                  \psfrag{0.05}[][][0.5]{$0.05$}
                  \psfrag{0.15}[][][0.5]{$0.15$}
                  \psfrag{0.25}[][][0.5]{$0.25$}
                  \psfrag{90}[][][0.5]{$90$}
                  \psfrag{0.1}[][][0.5]{$0.1$}
                  \psfrag{110}[][][0.5]{$110$}
         \includegraphics[width=\textwidth]{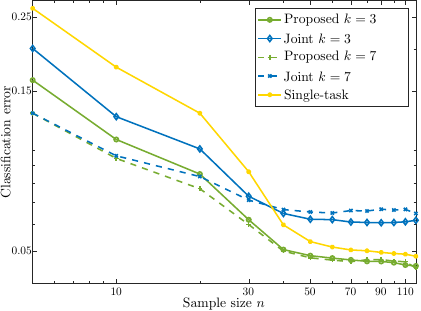}
                                \caption{Classification error for \ac{MDA} per sample size.}   
                                \label{fig:domain_batch}
\end{subfigure}
                                             \begin{subfigure}[t]{0.48\textwidth}
         \centering
                  \psfrag{n = 50}[l][l][0.6]{Proposed $n = 50$}
                  \psfrag{n = 20}[l][l][0.6]{Proposed $n = 20$}
                   \psfrag{Single n = 20}[l][l][0.6]{Single-task $n = 20$}
                   \psfrag{Single n = 50}[l][l][0.6]{Single-task $n = 50$}
                  \psfrag{Joint n = 50abcdefghijk}[l][l][0.6]{Joint $n = 50$}
                  \psfrag{Joint  n = 20}[l][l][0.6]{Joint $n = 20$}
                  \psfrag{Classification error}[][][0.6]{Classification error}
                  \psfrag{Number of tasks}[][][0.6]{Number of tasks $k$}
                  \psfrag{1}[][][0.5]{$1$}
                  \psfrag{3}[][][0.5]{$3$}
                  \psfrag{5}[][][0.5]{$5$}
                  \psfrag{7}[][][0.5]{$7$}
                   \psfrag{9}[][][0.5]{$9$}
                  \psfrag{0.8}[][][0.5]{$0.8$}
                  \psfrag{1.2}[][][0.5]{$1.2$}
                  \psfrag{0.3}[][][0.5]{$0.3$}
                  \psfrag{0.2}[][][0.5]{$0.2$}
                  \psfrag{0.1}[][][0.5]{$0.1$}
                  \psfrag{0}[][][0.5]{$0$}
                  \psfrag{0.05}[][][0.5]{$0.05$}
                  \psfrag{0.06}[][][0.5]{$0.06$}
                  \psfrag{0.07}[][][0.5]{$0.07$}
                  \psfrag{0.08}[][][0.5]{$0.08$}
         \includegraphics[width=\textwidth]{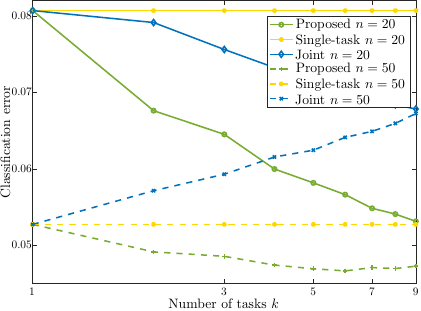}
                                \caption{Classification error for \ac{MDA} per number of tasks.}   
                                \label{fig:domain_ntasks}
\end{subfigure}
              \caption{Results on ``Yearbook'' dataset show the relationship among classification error, number of tasks, and sample size for \ac{MDA}.}
                       \label{fig:batch_scenarios}
\end{figure}

Figures~\ref{fig:error_multidimensional_n100} and~\ref{fig:multidimensional_continual} show the classification error of the proposed methodology and state-of-the-art techniques increasing the variance of the change between consecutive tasks. Such figures show the performance improvement due to the multidimensional adaptation in comparison with state-of-the-art techniques leveraging information from preceding tasks (forward learning) and from all the tasks in the sequence (forward and backward learning). Figure~\ref{fig:Multidimensional_100} shows that the proposed methodology better account for multidimensional tasks' changes than state-of-the-art techniques. 

\subsection{Classification error in batch learning scenarios}

In the following we show the relationship among classification error, number of tasks, and sample size in batch learning scenarios using real datasets. These numerical results are obtained averaging, for each number of tasks and sample size, the classification errors achieved with $10$ random instantiations of data samples in "Yearbook'' dataset. The performance improvement of the proposed methodology is compared with relevant baselines for learning from a sequence of tasks: joint learning (also known as offline learning)  \citep{mai2022online} 
and single-task learning (also known as independent learning) \citep{riemer2018learning}. Joint learning and single-task learning obtain classification rules as in standard supervised classification using the samples from all the tasks and using samples only from the corresponding task, respectively. 

 \begin{figure}
         \centering
                                    \begin{subfigure}[t]{0.48\textwidth}
         \centering
                  \psfrag{k = 3}[l][l][0.6]{Proposed $k = 3$}
                  \psfrag{k = 7}[l][l][0.6]{Proposed $k = 7$}
                   \psfrag{Single k = 10}[l][l][0.6]{Single-task $k = 10$}
                  \psfrag{Singleabcdefghijklmnopq}[l][l][0.6]{Single-task}
                  \psfrag{All k = 3}[l][l][0.6]{Joint $k = 3$}
                  \psfrag{All k = 7}[l][l][0.6]{Joint $k = 7$}
                  \psfrag{Classification error}[][][0.6]{Classification error}
                  \psfrag{Batch size}[][][0.6]{Sample size $n$}
                  \psfrag{30}[][][0.5]{$30$}
                  \psfrag{50}[][][0.5]{$50$}
                  \psfrag{10}[][][0.5]{$10$}
                  \psfrag{70}[][][0.5]{$70$}
                  \psfrag{90}[][][0.5]{$90$}
                  \psfrag{110}[][][0.5]{$110$}
                  \psfrag{0.5}[][][0.5]{$0.5$}
                  \psfrag{0.4}[][][0.5]{$0.4$}
                  \psfrag{0.3}[][][0.5]{$0.3$}
                  \psfrag{0.2}[][][0.5]{$0.2$}
                  \psfrag{0.1}[][][0.5]{$0.1$}
                  \psfrag{0}[][][0.5]{$0$}
         \includegraphics[width=\textwidth]{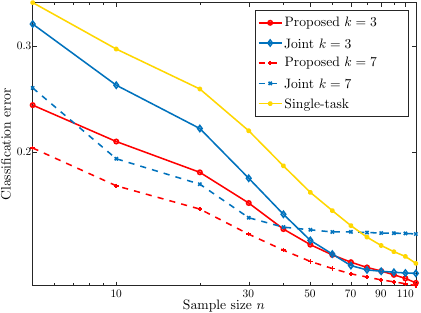}
                                \caption{Classification error for \ac{MTL} per sample size.}   
                                \label{fig:multitask_batch}
\end{subfigure}
                            \begin{subfigure}[t]{0.48\textwidth}
         \centering
                  \psfrag{n = 100}[l][l][0.6]{Proposed $n = 100$}
                  \psfrag{n = 10}[l][l][0.6]{Proposed $n = 10$}
                   \psfrag{Single n = 10}[l][l][0.6]{Single-task $n = 10$}
                   \psfrag{Single n = 100}[l][l][0.6]{Single-task $n = 100$}
                  \psfrag{All n = 100abcdefghijklm}[l][l][0.6]{Joint $n = 100$}
                  \psfrag{All n = 10}[l][l][0.6]{Joint $n = 10$}
                  \psfrag{Classification error}[][][0.6]{Classification error}
                  \psfrag{Number of tasks}[][][0.6]{Number of tasks $k$}
                  \psfrag{1}[][][0.5]{$1$}
                  \psfrag{3}[][][0.5]{$3$}
                  \psfrag{5}[][][0.5]{$5$}
                  \psfrag{7}[][][0.5]{$7$}
                  \psfrag{0.7}[][][0.5]{$0.7$}
                  \psfrag{0.9}[][][0.5]{$0.9$}
                  \psfrag{0.3}[][][0.5]{$0.3$}
                  \psfrag{0.2}[][][0.5]{$0.2$}
                  \psfrag{0.1}[][][0.5]{$0.1$}
                  \psfrag{0}[][][0.5]{$0$}
         \includegraphics[width=\textwidth]{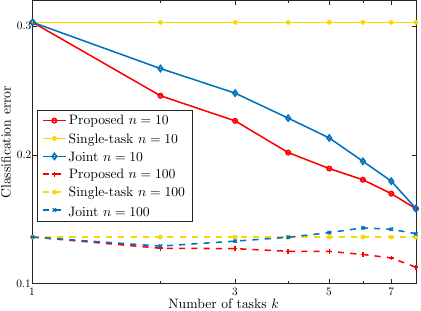}
                                \caption{Classification error for \ac{MTL} per number of tasks.}   
                                \label{fig:multitask_ntasks}
\end{subfigure}
              \caption{Results on ``Yearbook'' dataset show the relationship among classification error, number of tasks, and sample size for \ac{MTL}.}
                       \label{fig:batch_scenarios}
\end{figure}

Figures~\ref{fig:domain_batch} and~\ref{fig:domain_ntasks} show the classification error for \ac{MDA} for different sample sizes and number of tasks. Such figures show that the proposed learning methodology achieves significantly better results than joint learning and single-task learning. In particular, Figure~\ref{fig:domain_batch} shows that for $n = 20$ samples per tasks joint learning requieres $k = 7$ tasks to achieve similar results to the proposed methodology with $k = 3$; and shows that single-task learning requieres $n = 30$ samples to achieve similar results to the proposed methodology with $k = 3$ tasks and $n = 20$ samples per task. In addition, Figure~\ref{fig:domain_ntasks} shows that the proposed methodology improves performance increasing the number of tasks, while the performance of single-task learning remains constant increasing the number of tasks and the performance of joint learning decreases with $n = 50$ samples per task. 

Figures~\ref{fig:multitask_batch} and~\ref{fig:multitask_ntasks} show the classification error for \ac{MTL} for different sample sizes and number of tasks. Such figures show that the proposed learning methodology achieves significantly better results than joint learning and single-task learning. In particular, Figure~\ref{fig:multitask_batch} shows that with $n = 40$ samples per task, joint learning requieres $k = 7$ tasks to achieve similar results to the proposed methodology with $k = 3$; and shows that single-task learning requieres $n = 70$ samples to achieve similar results to the proposed methodology with $k = 3$ tasks and $n = 40$ samples per task. In addition, Figure~\ref{fig:multitask_ntasks} shows that the proposed methodology can improve performance as tasks arrive. The methods proposed can effectively adapt to tasks' changes that improves classification performance in all the experimental results.

\begin{table}[]
\centering
\setstretch{1.2}
\caption{Classification error and standard deviation of the proposed methodology in comparison with the state-of-the-art techniques \ac{SCD}.}
\label{tab:error_cd}
\begin{adjustbox}{width=\textwidth,center}
\begin{tabular}{crrrrrrrr}
\hline
Dataset                  & \multicolumn{2}{c}{Condor}                       & \multicolumn{2}{c}{Drift Surf}                   & \multicolumn{2}{c}{AUE}                          & \multicolumn{2}{c}{Proposed methods}                         \\
\hline
\multicolumn{1}{c}{$n$} & \multicolumn{1}{l}{10} & \multicolumn{1}{l}{100} & \multicolumn{1}{l}{10} & \multicolumn{1}{l}{100} & \multicolumn{1}{l}{10} & \multicolumn{1}{l}{100} & \multicolumn{1}{l}{10} & \multicolumn{1}{l}{100} \\
\hline
BAF                     & 1.11  $\pm$ 0.00                 & 1.12 $\pm$ 0.00                   & 1.08     $\pm$ 0.00                  & 0.96       $\pm$ 0.00                 & 1.06  $\pm$ 0.00                     & 1.05          $\pm$ 0.00              & \textbf{0.59}       $\pm$ 0.00                & \textbf{0.59}   $\pm$ 0.00                     \\
Elec2                   & 38.90 $\pm$ 0.45                 & 40.10         $\pm$ 0.30          & 42.90    $\pm$ 0.61              & 43.66            $\pm$ 0.43       & 42.38      $\pm$ 0.55            & 43.34          $\pm$ 0.37         & \textbf{38.83} $\pm$ 0.66       & \textbf{38.29} $\pm$ 0.21        \\
Airlines                & 43.32   $\pm$ 0.20               & 43.42   $\pm$ 0.19                & 44.41           $\pm$ 0.00         & 45.66    $\pm$ 0.00                 & 44.54      $\pm$ 0.00              & 45.74    $\pm$ 0.00                 & \textbf{39.07} $\pm$ 0.00       & \textbf{38.74} $\pm$ 0.00        \\
USPS                    & 48.52      $\pm$ 5.58            & 48.01      $\pm$ 5.50             & \textbf{38.00} $\pm$ 6.30                 & 38.00         $\pm$ 5.25     & 43.60     $\pm$ 6.46             & 38.80         $\pm$ 5.74          & 40.61 $\pm$ 4.43       & \textbf{34.66} $\pm$ 1.89        \\
Spam                    & \textbf{24.72}       $\pm$ 1.59           & 26.25    $\pm$ 1.93               & 33.68     $\pm$ 3.85             & 32.17     $\pm$ 2.17              & 27.38          $\pm$ 3.02         & 29.70           $\pm$ 1.59         & 26.23 $\pm$ 1.53       & \textbf{20.80} $\pm$ 0.91        \\
Power supply            & \textbf{34.33}     $\pm$ 0.45              & 33.10           $\pm$ 0.24        & 46.27     $\pm$ 0.95             & 43.32      $\pm$ 0.43             & 46.27       $\pm$ 0.95                      & 43.32           $\pm$ 0.43                & 40.26 $\pm$ 0.80       & \textbf{28.99} $\pm$ 0.29       \\
\hline
\end{tabular}
\end{adjustbox}
\end{table}

\subsection{Classification error in online learning scenarios}

In this section, we compare the performance of the proposed methodology applied to online learning scenarios with the state-of-the-art techniques for $n = 10$ and $n = 100$ samples per task. These numerical results are obtained computing the average classification error over all the tasks in $100$ random instantiations of data samples. The proposed methodology is compared with $3$ state-of-the-art \ac{SCD} techniques: Condor \citep{zhao2020handling}, DriftSurf \citep{tahmasbi2021driftsurf}, and \ac{AUE} \citep{brzezinski2013reacting}; and $3$ state-of-the-art \ac{CL} techniques: \ac{GEM} \citep{lopez2017gradient}, \ac{MER} \citep{riemer2018learning}, and \ac{EWC} \citep{kirkpatrick2017overcoming}. The hyper-parameters in these methods are set to the default values provided by the authors.

Tables~\ref{tab:error_cd} and~\ref{tab:error_continual} show the classification error and the standard deviation of the state-of-the-art techniques using $12$ real datasets. \footnote{Bold numbers indicate the top result for $n = 10$ and $n = 100$ samples per task.} Such tables show that the proposed methodology applied to \ac{SCD} and \ac{CL} offers an overall improved performance compared to existing techniques. The proposed methodology can significantly improve performance using evolving tasks with respect to the state-of-the-art techniques. 

\begin{table}[]
\centering
\setstretch{1.2}
\caption{Classification error and standard deviation of the proposed methodology in comparison with the state-of-the-art techniques for \ac{CL}.}
\label{tab:error_continual}
\begin{adjustbox}{width=\textwidth,center}
\begin{tabular}{crrrrrrrr}
\hline
Dataset        & \multicolumn{2}{c}{GEM}                          & \multicolumn{2}{c}{MER}                          & \multicolumn{2}{c}{EWC}                          & \multicolumn{2}{c}{Proposed methods}                         \\
\hline
$n$            & \multicolumn{1}{c}{10} & \multicolumn{1}{c}{100} & \multicolumn{1}{c}{10} & \multicolumn{1}{c}{100} & \multicolumn{1}{c}{10} & \multicolumn{1}{c}{100} & \multicolumn{1}{c}{10} & \multicolumn{1}{c}{100} \\
\hline
Yearbook       & 43.53 $\pm$ 10.70      & 23.45 $\pm$  6.98       & 38.62 $\pm$11.92       & 19.37 $\pm$ 7.05        & 48.94 $\pm$ 6.06       & 37.17 $\pm$ 8.27        &  \textbf{14.18} $\pm$ 1.51       &  \textbf{9.42} $\pm$ 0.57         \\
ImageNet noise & 39.09 $\pm$ 7.71       & 13.78 $\pm$ 7.31       & 27.25 $\pm$ 8.39       & 12.71 $\pm$ 5.15        & 45.75 $\pm$ 6.69       & 30.68 $\pm$ 8.12        &  \textbf{14.48} $\pm$ 1.55       &  \textbf{8.68} $\pm$ 0.54         \\
DomainNet      & 69.78 $\pm$ 5.06       & 53.60 $\pm$ 10.79       & 47.58 $\pm$ 9.71       & 30.26 $\pm$ 11.01       & 50.03 $\pm$ 6.31       & 41.17 $\pm$ 2.92        &  \textbf{33.81} $\pm$ 2.31       & \textbf{24.53} $\pm$ 1.21        \\
UTKFaces       & 12.20 $\pm$ 0.00       & 12.10 $\pm$ 0.00        & 12.13 $\pm$ 0.00       & 12.13 $\pm$ 0.00        & 12.13 $\pm$ 0.00       & 12.13 $\pm$ 0.00        &  \textbf{10.32} $\pm$ 0.18       &  \textbf{10.32} $\pm$ 0.00        \\
CLEAR          & 56.60 $\pm$ 10.35      & 8.60 $\pm$ 2.02        & 20.53 $\pm$ 9.75       & 7.40 $\pm$ 3.27         & 62.73 $\pm$ 4.28       & 38.53 $\pm$ 7.11        &  \textbf{8.22} $\pm$ 1.74        &  \textbf{4.06} $\pm$ 0.73        \\
Rotated MNIST  & 45.28 $\pm$ 4.48      & 32.02 $\pm$ 2.64       &  \textbf{36.34} $\pm$ 3.79       & 34.54 $\pm$ 4.47        & 47.05 $ \pm$ 1.69      & 39.93 $\pm$ 1.53        & 37.02 $\pm$ 0.87       &  \textbf{21.04} $\pm$ 0.39        \\
\hline
Rotated MNIST iid & 37.43 $\pm$ 3.84 & 25.97 $\pm$ 1.13 & \textbf{19.34} $\pm$ 0.93 & \textbf{18.30} $\pm$ 0.15 & 43.41 $\pm$ 0.02 &33.13 $\pm$ 1.08 &44.05 $\pm$ 0.01 & 24.65 $\pm$ 0.01 \\
\hline
\end{tabular}
\end{adjustbox}
\end{table}

 \begin{figure}
         \centering
                            \begin{subfigure}[t]{0.48\textwidth}
         \centering
                  \psfrag{Order p = 0abc}[l][l][0.6]{Order $p = 0$}
                  \psfrag{Order p = 1}[l][l][0.6]{Order $p = 1$}
                  \psfrag{Order p = 2}[l][l][0.6]{Order $p = 2$}
                  \psfrag{Delta}[][][0.6]{Time increment between tasks $\Delta$}
                  \psfrag{Classification error}[][][0.6]{Classification error}
                  \psfrag{1}[][][0.5]{$2$}
                  \psfrag{2}[][][0.5]{$3$}
                  \psfrag{3}[][][0.5]{$4$}
                  \psfrag{4}[][][0.5]{$5$}
                  \psfrag{5}[][][0.5]{$6$}
                  \psfrag{0.23}[][][0.5]{$0.23$}
                  \psfrag{0.25}[][][0.5]{$0.25$}
                  \psfrag{0.27}[][][0.5]{$0.27$}
                  \psfrag{0.29}[][][0.5]{$0.29$}
         \includegraphics[width=\textwidth]{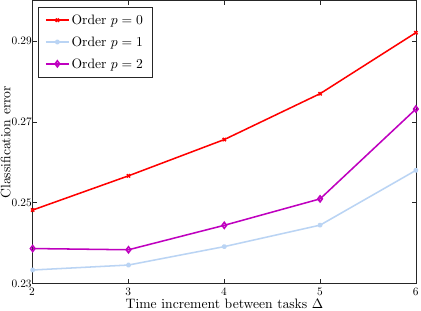}
                                \caption{Classification error per time increment.}   
                                \label{fig:order}
\end{subfigure}
                            \begin{subfigure}[t]{0.48\textwidth}
         \centering
                  \psfrag{Order p = 0abc}[l][l][0.6]{Order $p = 0$}
                  \psfrag{Order p = 1}[l][l][0.6]{Order $p = 1$}
                  \psfrag{Order p = 2}[l][l][0.6]{Order $p = 2$}
                  \psfrag{Sample size}[][][0.6]{Sample size $n$}
                  \psfrag{Classification error}[][][0.6]{Classification error}
                  \psfrag{10}[][][0.5]{$10$}
                  \psfrag{30}[][][0.5]{$30$}
                  \psfrag{50}[][][0.5]{$50$}
                   \psfrag{70}[][][0.5]{$70$}
                    \psfrag{90}[][][0.5]{$90$}
                     \psfrag{110}[][][0.5]{$110$}
                  \psfrag{0.5}[][][0.5]{$0.5$}
                  \psfrag{0.4}[][][0.5]{$0.4$}
                  \psfrag{0.3}[][][0.5]{$0.3$}
                  \psfrag{0.2}[][][0.5]{$0.2$}
                  \psfrag{0.1}[][][0.5]{$0.1$}
                  \psfrag{0}[][][0.5]{$0$}
         \includegraphics[width=\textwidth]{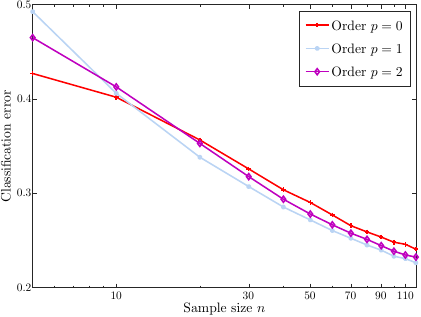}
                                \caption{Classification error per sample size.}   
                                \label{fig:order}
\end{subfigure}
              \caption{Results on ``Rotated MNIST'' dataset show the improvement of accounting for high-order time dependences.}
                       \label{fig:order2}
\end{figure}

In order to more clearly assess the performance improvement of the proposed method using datasets with evolving tasks, we also use ``Rotated MNIST i.i.d.'' dataset that is composed by non-evolving tasks. This dataset satisfies the (i.i.d.-A) assumption since it contains rotated images with angles uniformly sampled from $[0, 360)$ over tasks. Table 3 shows that the proposed methodology outperforms existing techniques in cases with evolving tasks; while existing techniques obtain improved performance in cases with non-evolving tasks. In particular, the performance of existing methods is significantly better in ``Rotated MNIST i.i.d.'' than in ``Rotated MINST'' while the performance of the proposed method is significantly better in ``Rotated MNIST'' than in ``Rotated MINST i.i.d.'' As shown in Table 3, the presented methodology can enable significant performance improvements using evolving tasks but not in datasets with non-evolving tasks.

\subsection{High-order dependences}

This section shows the classification performance of the proposed methodology accounting for high-order dependences among tasks.  These numerical results are obtained averaging the classification errors for \ac{CL} scenarios achieved with $100$ random instantiations of data samples in "Yearbook'' dataset using $n = 10$ samples per task. Figure~\ref{fig:order2} shows the relationship between classification error, order, and time increment.  Such figure shows that order $p = 1$ can result in an improved overall performance at the expenses of worst initial performance. In addition, Figure~\ref{fig:order2} shows that order $p = 2$ may perform worse than order $p = 1$ because the number of model parameters increases with higher order. 

The experimental results show that the proposed methodology achieves better performance than existing techniques. In addition, the numerical results show that the proposed
performance guarantees can offer tight upper and lower bounds
for error probabilities of each task.

\section{Conclusion} \label{sec:conc}

The paper proposes a learning methodology for evolving tasks that is applicable to multiple supervised learning scenarios and provides computable performance guarantees. The proposed methodology accounts for multidimensional adaptation to changes by estimating multiple statistical characteristics of the underlying distribution. In addition, the paper analytically characterizes the increase in \ac{ess} achieved by the proposed methodology in terms of the expected quadratic change and the number of tasks. The numerical results assess the reliability of the performance guarantees presented and show the performance improvement in multiple supervised learning scenarios using multiple datasets, sample sizes, and number of tasks. The methodology proposed can lead to efficient methods for multiple learning scenarios and provide performance guarantees.

\acks{Funding in direct support of this work has been provided by projects PID2022-137063NBI00,
PID2022-137442NB-I00, CNS2022-135203, and \mbox{CEX2021-001142-S} funded by
MCIN/AEI/10.13039/501100011033 and the European Union “NextGenerationEU”/PRTR,
BCAM Severo Ochoa accreditation \mbox{CEX2021-001142-S/}MICIN/AEI/10.13039/501100011033
funded by the Ministry of Science and Innovation, and programes ELKARTEK, IT1504-22, and
BERC-2022-2025 funded by the Basque Government.}

\newpage
\appendix 

\section{Proof of Theorems~\ref{th:kalman1},~\ref{th:kalman}, and~\ref{th:high_order}} \label{app:rec}
\begin{proof}
To prove Theorems~\ref{th:kalman1} and~\ref{th:kalman}, we use that the recursions in~\eqref{eq:tau_general},~\eqref{eq:s_general}, and recursions~\eqref{eq:tau_general_back},~\eqref{eq:s_general_back} correspond to those for filtering in linear dynamical systems that obtain the unbiased linear estimator with minimum \ac{MSE}.

The mean vectors $\B{\tau}_j^\infty = \mathbb{E}_{\up{p}_j}\{\Phi(x, y)\}$ evolve over tasks through the linear dynamical system
\begin{align}
\label{eq:w}{\B{\tau}}_{j+1}^\infty & = {\B{\tau}}_{j}^\infty + \B{w}_{j+1}
\end{align}
because 
\begin{align}
\B{\tau}_{j+1}^\infty & = \mathbb{E}_{\up{p}_{j+1}}\{\Phi(x, y)\} = \int \Phi(x,y) d \up{p}_{j+1}(x,y)\nonumber\\
\label{eq:p_tau}
&  = \int \Phi(x,y) d (\up{p}_{j} + \varepsilon_{j+1})(x,y) = \B{\tau}_{j}^\infty + \B{w}_{j+1}
\end{align}
with $ \B{w}_{j} =  \int \Phi(x,y) d \varepsilon_{j}(x,y)$. Equality~\eqref{eq:p_tau} follows since the sequence of probability distributions satisfies the~\eqref{eq:td} assumption, that is 
\begin{align*}
\up{p}_{j+1} & = \up{p}_{j} + \varepsilon_{j+1}.
\end{align*}
So that the random vectors $\B{w}_{j}$ are independent and zero-mean because they are a linear function of the independent zero-mean random measures $\varepsilon_{j+1}$ (see e.g., \citet{kallenberg2017random}). 

Each state variable $\B{\tau}_j^\infty$ is observed at each step $j$ through $\B{\tau}_j$ that is the sample average of i.i.d. samples from $\up{p}_j$, so that we have
\begin{align}
\label{eq:v}
\B{\tau}_j & = {\B{\tau}}_j^\infty + \B{v}_{j}
\end{align}
where $\B{v}_j$ for $j\in \{1, 2, \ldots, k\}$ are independent and zero-mean, and independent of $\B{w}_j$ for $j \in \{1, 2, \ldots, k\}$. Therefore, equations~\eqref{eq:w} and~\eqref{eq:v} above describe a linear dynamical system (state-space model with white noise processes) \citep{bishop2006pattern, anderson2012optimal}. For such systems, the Kalman filter recursions provide the unbiased linear estimator with minimum \acs{MSE} based on samples corresponding to preceding steps $D_1, D_2, \ldots, D_j$ \citep{bishop2006pattern, anderson2012optimal}. The Kalman filter recursions are given by
\begin{align}
\B{\tau}_j^{j-1} & = \B{\tau}_{j-1}^{j-1}\\
\B{s}_j^{j-1} & = \B{s}_{j-1}^{j-1} + \B{d}_j\\
\label{eq:kalman1}
\B{\tau}_j^{j} & = \B{\tau}_j^{j-1} + \B{\eta}_j^j(\B{\tau}_j - \B{\tau}_{j}^{j-1})\\
\label{eq:kalman2}
\B{s}_j^j & = \B{s}_j^{j-1} + \B{\eta}_j^j \B{s}_j^{j-1}\\
\label{eq:kalman3}
\B{\eta}_j^j & = \frac{\B{s}_j^{j-1}}{\B{s}_j^{j-1} + \B{s}_j}.
\end{align}
that lead to~\eqref{eq:tau_general}-\eqref{eq:gain_general} substituting $\B{\tau}_j^{j-1}$ and $\B{s}_j^{j-1}$ in equations~\eqref{eq:kalman1}-\eqref{eq:kalman3}.

The Rauch-Tung-Striebel smoother recursions provide the unbiased linear estimator with minimum \acs{MSE} based on samples corresponding to preceding and succeeding steps $D_1, D_2, \ldots, D_k$ \citep{bishop2006pattern, anderson2012optimal}. The Rauch-Tung-Striebel smoother recursions are given by
\begin{align}
\label{eq:rauch1}
    \B{\tau}_j^k & = \B{\tau}_j^j + \B{\eta}_j^j(\B{\tau}_{j+1}^k - \B{\tau}_{j+1}^j)\\
    \label{eq:rauch2}
    \B{s}_j^k & = \b{s}_j^j + \B{\eta}_j^j(\B{s}_{j+1}^k-\B{s}_{j+1}^j)\B{\eta}_j^j\\
    \label{eq:rauch3}
    \B{\eta}_j^j & = \frac{\B{s}_j^j}{\B{s}_{j+1}^j}
\end{align}
Substituting $\B{\tau}_{j+1}^j$ and $\B{s}_{j+1}^j$ in equations~\eqref{eq:rauch1}-\eqref{eq:rauch3}, we have that 
\begin{align}
    \B{\tau}_j^k & = \B{\tau}_j^j + \B{\eta}_j^j(\B{\tau}_{j+1}^k - \B{\tau}_{j}^j)\\
    \B{s}_j^k & = \B{s}_j^j + \B{\eta}_j^j(\B{s}_{j+1}^k-\B{s}_{j}^{j} + \B{d}_{j+1})\B{\eta}_j^j\\
    \B{\eta}_j^j & = \frac{\B{s}_j^j}{\B{s}_{j}^{j} + \B{d}_{j+1}}
\end{align}
that lead to~\eqref{eq:tau_general_back}-\eqref{eq:gain_general_back} by defining $\B{\eta}_j^k = 1 - \B{\eta}_j^j = \frac{\B{d}_{j+1}}{\B{s}_j^j + \B{d}_{j+1}}$. 
Then, equations~\eqref{eq:tau_general} and \eqref{eq:s_general}, and equations~\eqref{eq:tau_general_back} and \eqref{eq:s_general_back} are obtained after some algebra from the Kalman filter recursions and Rauch-Tung-Striebel smoother recursions.

Similarly as proof of Theorems~\ref{th:kalman1} and~\ref{th:kalman}, we prove Theorem~\ref{th:high_order} using that recursions~\eqref{eq:gamma_c},~\eqref{eq:Sigma_c}, and recursions~\eqref{eq:gamma_back},~\eqref{eq:Sigma_back} are obtained after some algebra
from the Kalman recursions for updated state vector and updated MSE matrix and from the Rauch-Tung-Striebel smoother recursions. The unbiased linear estimator with minimum \ac{MSE}
for a dynamical system such as~\eqref{eq:kalman} is given by the Kalman filter recursions (see e.g., \citet{bishop2006pattern, anderson2012optimal}). 

Let ${\tau_j^{\infty}}^{(i)}$ denote the $i$-th component of the mean vector ${\tau_j^{\infty}}$. Dynamical systems as that given by~\eqref{eq:kalman} can be derived using the $p$-th order Taylor expansion of the expectations ${{\tau}_{j}^{\infty}}^{(i)}$ with $i = 1, 2, \ldots, m$. The Taylor expansions of ${{\tau}_{j}^{\infty}}^{(i)}$ and its successive derivatives up to order $p$ are given by
\begin{align*}
{{\tau}_{j}^{\infty}}^{(i)} &  \approx {{\tau}_{j-1}^{\infty}}^{(i)} +  \Delta_j {{{\tau}_{j}^{\infty}}^{(i)}}' +  \frac{\Delta_j^2}{2} {{{\tau}_{j}^{\infty}}^{(i)}}'' + ... + \frac{\Delta_j^p}{p!} {{{\tau}_{j-1}^{\infty}}^{(i)}}^{p)}\\ 
{{{\tau}_{j}^{\infty}}^{(i)}}' & \approx {{{\tau}_{j-1}^{\infty}}^{(i)}}' + \Delta_j {{{\tau}_{j-1}^{\infty}}^{(i)}}'' + ... + \frac{\Delta_j^{p-1}}{(p)!} {{{\tau}_{j-1}^{\infty}}^{(i)}}^{p-1)}\\
\vdots & \\
{{{\tau}_{j}^{\infty}}^{(i)}}^{p)}& \approx {{{\tau}_{j-1}^{\infty}}^{(i)}}^{p)}
\end{align*}
where $\Delta_j$ is the time increment between the $j$-th and the $j-1$-th tasks and ${{{\tau}_{j}^{\infty}}^{(i)}}^{p)}$ denotes the $p$ derivative of ${{{\tau}_{j}^{\infty}}^{(i)}}$. The above equations lead to
\begin{align}
\label{eq:e}
\B{\gamma}_{j, i}^{\infty} &  \approx \begin{pmatrix} 1 & \Delta_j &  \frac{\Delta_j^2}{2} & ... &  \frac{\Delta_j^p}{p!} \\
0 & 1 & \Delta_j & ... &  \frac{\Delta_j^{p-1}}{(p-1)!}\\
\vdots & \vdots & \vdots &&\vdots\\
0 & 0 & 0 & ... & 1
\end{pmatrix} \B{\gamma}_{j-1, i}^{\infty}
 \end{align}
since $\B{\gamma}_{j, i}^{\infty}$ is composed by ${{\tau}_{j}^{\infty}}^{(i)}$ and its derivatives up to order $p$. In addition, ${{\tau}_{j}^{\infty}}^{(i)}$ is observed at each time $t$ through the sample set $D_{j}$, so that we have 
\begin{equation}
\label{eq:m}
\frac{1}{n_{j}}\sum_{{l = 1}}^{n_{j}}\Phi(x_{j, l}, y_{l})^{(i)} \approx {{\tau}_{j}^{\infty}}^{(i)}
\end{equation}
with $i = 1, 2, \ldots, m$. Equations~\eqref{eq:e} and~\eqref{eq:m} above lead to the dynamical system in~\eqref{eq:kalman}.
\end{proof}

\section{Proof of Theorem~\ref{th:ess_recursion}} \label{eq:al}
 \begin{proof}
 To obtain bound in~\eqref{eq:ess} we first prove that the mean vector estimate and the \acs{MSE} vector given by~\eqref{eq:tau_general} and~\eqref{eq:s_general}, respectively, satisfy  
 \begin{align}
\label{eq:pr}
& \mathbb{P}\left\{|{{\tau}_j^\infty}^{(i)} -{{\tau}_{j}^{j}}^{(i)}| \leq \kappa \sqrt{2 {s_{j}^{j}}^{(i)} \log\left(\frac{2m}{\delta}\right)}\right\}  \geq (1 - \delta)
\end{align}
for any component $i = 1, 2, \ldots, m$. Then, we prove that $\|\sqrt{\B{s}_j^j}\|_\infty \leq M/\sqrt{n_j^j}$ for $j\in\{1,2,\ldots,k\}$, where the \acsp{ess} satisfy $n_1^1 = n_1$ and $n_{j}^{j} \geq n_j + n_{j-1}^{j-1} \frac{\|\B{\sigma}_{j}^2\|_\infty}{\|\B{\sigma}_{j}^2\|_\infty +  \|\B{d}_j\|_\infty n_{j-1}^{j-1}}$ for $j \geq 2$.

To obtain inequality~\eqref{eq:pr}, we prove by induction that each component $i = 1, 2, \ldots, m$ of the error in the mean vector estimate ${{z}_{j}^{j}}^{(i)} = {{\tau}_j^\infty}^{(i)}- {{\tau}_{j}^{j}}^{(i)}$ is sub-Gaussian with parameter ${{\rho}_{j}^{j}}^{(i)} \leq \kappa \sqrt{{{s}_{j}^{j}}^{(i)}}$. Firstly, for $j = 1$, we have that
\begin{align*}
{{z}_1^1}^{(i)} & = {\tau_1^\infty}^{(i)} -{\tau_1^1}^{(i)} = {\tau_1^\infty}^{(i)} -\tau_1^{(i)}.
\end{align*}
Since the bounded random variable $\Phi^{(i)}_1$ is sub-Gaussian with parameter $\sigma(\Phi^{(i)}_1)$, then the error in the mean vector estimate ${{z}_1^1}^{(i)}$ is sub-Gaussian with parameter that satisfies
$$\left({\rho_1^1}^{(i)}\right)^2 = \frac{\sigma\left(\Phi^{(i)}_1\right)^2}{n_1} \leq \frac{\kappa^2 {\sigma_1^2}^{(i)}}{n_1} = \kappa^2 {s_1^{(i)}}.$$

If ${{z}_{j-1}^{j-1}}^{(i)} = {{\tau}_{j-1}^\infty}^{(i)} - {\tau_{j-1}^{j-1}}^{(i)}$ is sub-Gaussian with parameter ${{\rho}_{j-1}^{j-1}}^{(i)} \leq \kappa \sqrt{{s_{j-1}^{j-1}}^{(i)}}$ for any $i~=~1, 2, \ldots, m$, then using the recursions~\eqref{eq:tau_general} and~\eqref{eq:s_general} we have that 
\begin{align}
{{z}_{j}^{j}}^{(i)} & = {\tau_j^\infty}^{(i)} -{\tau_{j}^{j}}^{(i)} = {\tau_{j-1}^\infty}^{(i)} + {w}_j^{(i)} - \tau_j^{(i)} -  \frac{{s}_j^{(i)}}{{s_{j-1}^{j-1}}^{(i)} +{s}_j^{(i)}+ {d_j^2}^{(i)}}\left({\tau_{j-1}^{j-1}}^{(i)} - \tau_j^{(i)}\right) \nonumber\\
& = {\tau_{j-1}^\infty}^{(i)} + {w}_j^{(i)} - {\tau_{j-1}^{j-1}}^{(i)}  + \left(1 - \frac{{s}_j^{(i)}}{{s_{j-1}^{j-1}}^{(i)} +{s}_j^{(i)}+ {d_j^2}^{(i)}}\right)\left({\tau_{j-1}^{j-1}}^{(i)}  - \tau_j^{(i)} \right) \nonumber\\
& = {\tau_{j-1}^\infty}^{(i)} + {w}_j^{(i)} - {\tau_{j-1}^{j-1}}^{(i)}  - \frac{{s_j^j}^{(i)}}{{s}_j^{(i)}}\left( {\tau}_j^{(i)} - {\tau_{j-1}^{j-1}}^{(i)}\right) \nonumber
\end{align}
since $\B{w}_j = \B{\tau}_j^\infty - \B{\tau}_{j-1}^\infty$. If $\B{v}_j = {\B{\tau}}_j - \B{\tau}_j^\infty$, the error in the mean vector estimate is given by
\begin{align}
{{z}_{j}^{j}}^{(i)} & ={\tau_{j-1}^\infty}^{(i)} + {w}_j^{(i)} - {\tau_{j-1}^{j-1}}^{(i)} - \frac{{s_j^j}^{(i)}}{{s}_j^{(i)}}\left({\tau_j^\infty}^{(i)} + {v}_j^{(i)} - {\tau_{j-1}^{j-1}}^{(i)}\right) \nonumber\\
& = {\tau_{j-1}^\infty}^{(i)} + {w}_j^{(i)} - {\tau_{j-1}^{j-1}}^{(i)} - \frac{{s_j^j}^{(i)}}{{s}_j^{(i)}}\left({\tau_{j-1}^\infty}^{(i)} + {w}_j^{(i)} + {v}_j^{(i)} - {\tau_{j-1}^{j-1}}^{(i)}\right) \nonumber\\
& = \left(1 - \frac{{s_j^j}^{(i)}}{{s}_j^{(i)}}\right)  {z_{j-1}^{j-1}}^{(i)}  +  \left(1 - \frac{{s_j^j}^{(i)}}{{s}_j^{(i)}}\right) \left({w}_{j}^{(i)} \right) -\frac{{s_j^j}^{(i)}}{{s}_j^{(i)}} {v}_j^{(i)}\nonumber
\end{align}
where ${w}_j^{(i)}$ and ${v}_j^{(i)}$ are sub-Gaussian with parameter $\sigma(w_{j}^{(i)})$ and ${\sigma\left(\Phi^{(i)}_j\right)}/{\sqrt{n_j}}$, respectively.  Therefore, we have that ${z_{j}^{j}}^{(i)}$ is sub-Gaussian with parameter ${\rho_{j}^{j}}^{(i)}$ that satisfies
\begin{align}
\left({\rho_{j}^{j}}^{(i)}\right)^2  = & \left(1 - \frac{{s_j^j}^{(i)}}{{s}_j^{(i)}}\right)^2 \left({\rho_{j-1}^{j-1}}^{(i)}\right)^2 +  \left(1 - \frac{{s_j^j}^{(i)}}{{s}_j^{(i)}}\right)^2   \sigma\left(w_{j}^{(i)}\right)^2 +  \left(\frac{{s_j^j}^{(i)}}{{s}_j^{(i)}}\right)^2 \frac{\sigma\left(\Phi^{(i)}_j\right)^2}{n_j} \nonumber
\end{align}
since $\B{z}_{j-1}^{j-1}$, $\B{w}_j$, and $\B{v}_j$ are independent. Using that ${\rho_{j-1}^{j-1}}^{(i)} \leq {\kappa} \sqrt{{s_{j-1}^{j-1}}^{(i)}}$ and the definition of $\kappa$, we have that
\begin{align}
\left({\rho_{j}^{j}}^{(i)}\right)^2  
 & \leq  \left(1 -\frac{{s_j^j}^{(i)}}{{s}_j^{(i)}}\right)^2 {\kappa^2} {s_{j-1}^{j-1}}^{(i)} +  \left(1 - \frac{{s_j^j}^{(i)}}{{s}_j^{(i)}}\right)^2  \kappa^2  {d_j^2}^{(i)} + \left(\frac{{s_j^j}^{(i)}}{{s}_j^{(i)}}\right)^2 \kappa^2  \frac{{\sigma_j^2}^{(i)}}{n_j} \nonumber\\
    \label{eq:errores_clambda}
    & \leq \left(1 - \frac{{s_j^j}^{(i)}}{{s}_j^{(i)}}\right)^2 \kappa^2 \left( \left( \frac{1}{{{s_j^j}}^{(i)}} - \frac{1}{{{s_{j}}^{(i)}}}\right)^{-1} + {d_j^2}^{(i)}\right) +\frac{\left({s_j^j}^{(i)}\right)^2}{{s}_j^{(i)}} \kappa^2\\
  & = \left(1 - \frac{{s_j^j}^{(i)}}{{s}_j^{(i)}}\right) {\kappa^2} {s_{j}^{j}}^{(i)} + \kappa^2\frac{\left({s_j^j}^{(i)}\right)^2}{{s}_j^{(i)}}\nonumber
\end{align}
where~\eqref{eq:errores_clambda} is obtained using~\eqref{eq:s_general}. 

The inequality in~\eqref{eq:pr} is obtained using the union bound together with the Chernoff bound (concentration inequality) (see e.g., \citet{wainwright2019high}) for the random variables  ${z_{j}^{j}}^{(i)}$ that are sub-Gaussian with parameter ${\rho_{j}^{j}}^{(i)}$.

Now, we prove by induction that, for any $j$, $\|\sqrt{\B{s}_j^j}\|_\infty \leq M /\sqrt{n_j^j}$ where the \acsp{ess} satisfy $n_1^1 = n_1$ and $$n_{j}^{j} \geq n_j + n_{j-1}^{j-1} \frac{\|\B{\sigma}_{j}^2\|_\infty}{\|\B{\sigma}_{j}^2\|_\infty +  \|\B{d}_j\|_\infty n_{j-1}^{j-1}}$$ for $j \geq 2$. For $j = 1$, using the definition of $\B{s}_j^j$ in equation~\eqref{eq:s_general}, we have that for any component $i$
\begin{align*}
\left({s_1^1}^{(i)}\right)^{-1} = \left(s_1^{(i)}\right)^{-1} = \frac{n_1}{{\sigma_1^2}^{(i)}} \geq \frac{n_1}{M^2}.
\end{align*}
Then, vector $\B{s}_1^1$ satisfies $$\|\sqrt{\B{s}_1^1}\|_\infty \leq  \frac{M}{\sqrt{n_1}} = \frac{M}{\sqrt{n_1^1}}.$$

 If $\|\sqrt{\B{s}_{j-1}^{j-1}}\|_\infty \leq M /\sqrt{n_{j-1}^{j-1}}$, then we have that  for any component $i$
\begin{align*}
\left({s_{j}^{j}}^{(i)}\right)^{-1} & = \frac{1}{ s_{j}^{(i)}} + \frac{1}{{{{s}_{j-1}^{j-1}}^{(i)} + {{d}_{j}}^{(i)}}} \geq \frac{1}{ s_{j}^{(i)}}  + \frac{1}{{ \frac{M^2}{n_{j-1}^{j-1}}+  {{d}_{j}}^{(i)}}} \geq \frac{1}{M^2} \left(n_j  + \frac{1}{{ \frac{1}{ n_{j-1}^{j-1}}+  \frac{{{d}_{j}}^{(i)}}{M^2}}} \right)\\
& \geq \frac{1}{M^2} \left(n_j  + \frac{1}{{ \frac{1}{ n_{j-1}^{j-1}}+  \frac{\|{\B{d}_{j}}\|_\infty}{\|\B{\sigma}_j^2\|_\infty}}} \right)
\end{align*}
by using the recursion~\eqref{eq:s_general} and the induction hypothesis. Then, vector $\B{s}_j^j$ satisfies
\begin{equation*}
\left\|\sqrt{\B{s}_j^j}\right\|_\infty \leq \frac{M}{\sqrt{n_j + n_{j-1}^{j-1} \frac{\|\B{\sigma}_{j}^2\|_\infty}{\|\B{\sigma}_{j}^2\|_\infty +  \|\B{d}_j\|_\infty n_{j-1}^{j-1}}}}.
\end{equation*}

The inequality in~\eqref{eq:ess} is obtained because the minimax risk is bounded by the smallest minimax risk as shown in  \citep{mazuelas:2020, mazuelas:2022, MazRomGru:22} so that
\begin{align*}
R(\set{U}_j^{j}) \leq R_k^\infty + \left(\|\B{\tau}_k^\infty - \B{\tau}_j^{j}\|_\infty + \|\B{\lambda}_j^{j}\|_\infty \right)\left\|\B{\mu}_k^\infty\right\|_1
\end{align*}
that leads to~\eqref{eq:ess} using~\eqref{eq:pr} and $\|\sqrt{\B{s}_j^{j}}\|_\infty \leq M/\sqrt{n_j^{j}}$.
\end{proof}

\section{Proof of Theorem~\ref{th:forward}} \label{ap:proofthbound}

\begin{proof}
To obtain bound in~\eqref{eq:N_ineq}, we proceed by induction. For $j = 1$, using the expression for the \acs{ess} in~\eqref{eq:ess}, we have that
\begin{align*}
n_1^1 & = n_1 \geq n.
\end{align*}

 If~\eqref{eq:N_ineq} holds for the $(j-1)$-task, then for the $j$-th task, we have that 
\begin{align*}
 n_{j}^{j} & \geq {n_j} + n_{j-1}^{j-1} \frac{\|\B{\sigma}_j^2\|_\infty }{\|\B{\sigma}_j^2\|_\infty + {n_{j-1}^{j-1}}\|\B{d}_j\|_\infty } \geq n + n_{j-1}^{j-1} \frac{1}{1 + {n_{j-1}^{j-1}} d} = n\left(1 +  \frac{1}{\frac{n}{n_{j-1}^{j-1}} + n d} \right)
   \end{align*}
  where the second inequality is obtained because $n_j \geq n$, $\|\B{\sigma}_j^2\|_\infty \leq 1$, and $\|\B{d}_j\|_\infty  \leq d$. 
Using that $ n_{j-1}^{j-1} \geq n \left(1+ \frac{(1+\alpha)^{2 j-3} - 1 - \alpha}{\alpha(1+\alpha)^{2 j-3}+\alpha}\right)$, the \acs{ess} of the $j$-th task satisfies
\begin{align}
 n_{j}^{j}& \geq {n}\left(1 + \frac{1}{{ \frac{n}{n \left(1+ \frac{(1+\alpha)^{2 j-3} - 1 - \alpha}{\alpha(1+\alpha)^{2 j-3}+\alpha}\right)}+ n d}}\right) = n \left(1 + \frac{1}{{ \frac{\alpha(1+\alpha)^{2 j-3}+\alpha}{(1+\alpha)^{2 j-2}-1}+ n d}}\right) \nonumber\\
 \label{eq:proof_defalpha}
 & = n \left(1 + \frac{1}{{ \frac{\alpha(1+\alpha)^{2 j-3}+\alpha}{(1+\alpha)^{2 j-2}-1}+\frac{\alpha^2}{\alpha+1}}}\right) \\
 & = n \left(1 + \frac{(1+\alpha)^{2 j-1} -1-\alpha}{{\alpha(1+\alpha)^{2 j-2}+\alpha(\alpha+1)+\alpha^2(1+\alpha)^{2 j-2}- \alpha^2}}\right) \nonumber
 \end{align}
 where~\eqref{eq:proof_defalpha} is obtained because $n d = \frac{\alpha^2}{\alpha+1}$ since $\alpha = \frac{ n d}{2} \left(\sqrt{1+\frac{4}{nd}} + 1\right)$. 
 
Now, we obtain bounds for the \acs{ess} if $n d < \frac{1}{j^2}$. In the following, the constant $\phi$ represents the golden ratio $\phi = 1.618\ldots$.
\begin{enumerate}
\item If $n d < \frac{1}{j^2} \Rightarrow \sqrt{n d} \leq \alpha \leq \sqrt{n d} \phi \leq  \frac{\phi}{j} \leq 1$ similarly as in the preceding case, then we have that $n_{j}^{j}$ satisfies
\begin{align*}
n_{j}^{j} & \geq  n\frac{1}{\alpha} \frac{\alpha(2 j-2) }{2+\alpha(2 j-1)} =  n\frac{2 j-2}{2+\alpha(2 j-1)}\end{align*}
where the first inequality follows because $(1+\alpha)^{2 j-2} \geq 1 + \alpha(2j-2)$. Using $\alpha  \leq \frac{\phi}{j}$, we have that
\begin{align*}
n_{j}^{j} 
& \geq n \frac{2 j-2}{2+\frac{\phi}{j} (2 j-1)} \geq n\frac{2 j-2}{2+ 2\phi - \frac{\phi}{j}} \geq n \frac{ j-1}{1+ \phi}. 
\end{align*}
\item If $\frac{1}{j^2} \leq n d < 1 \Rightarrow \frac{1}{j}< \sqrt{n d} < \alpha < \sqrt{n d} \phi$ because ${\alpha} = \frac{n d}{2} \left(\sqrt{1 + \frac{4}{n d}} + 1\right) = \sqrt{n d}\frac{\sqrt{n d + 4} +  \sqrt{n d}}{2}$, then we have that $n_{j}^{j}$ satisfies
\begin{align*}
n_{j}^{j} 
& \geq n\frac{1}{\sqrt{n d}} \frac{1}{\phi} \frac{(1+\alpha)^{2 j-2} -1}{(1+\alpha)^{2 j-2}+1} \geq n \frac{1}{\sqrt{n d}} \frac{1}{\phi} \frac{(1+\frac{1}{j})^{2 j-2} -1}{(1+\frac{1}{j})^{2 j-2}+1}
\end{align*}
where the first inequality follows because $\alpha < \sqrt{n d} \phi$ and the second inequality follows because the expression is monotonically increasing for $\alpha$ and $\frac{1}{j}< \alpha$. Since $(1+\frac{1}{j})^{2 j-2} \geq 1 + \frac{2j-2}{j}$, we have that
\begin{align*}
n_{j}^{j} & \geq n \frac{1}{\sqrt{n d}} \frac{1}{\phi} \frac{\frac{{2 j-2}}{j}}{2+\frac{{2 j-2}}{j}} \geq n \frac{1}{\sqrt{n d}} \frac{1}{\phi} \frac{1}{3} \; \; \; \text{because $j \geq 2$.}
\end{align*}
\item If $n d \geq 1 \Rightarrow 1 \leq n d \leq \alpha \leq n d \phi$ because ${\alpha} = \frac{n d}{2} \left(\sqrt{1 + \frac{4}{n d}} + 1\right)$, then we have that $n_{j}^{j}$ satisfies
\begin{align*}
n_{j}^{j} &\geq n \frac{1}{n d \phi} \frac{(1+\alpha)^{2 j-1} -1 - \alpha}{(1+\alpha)^{2 j-1}+1}\geq  n \frac{1}{n d\phi} \frac{(1+\alpha)^{2 j-2} -1}{(1+\alpha)^{2 j-2}+1} 
\end{align*}
where the first inequality follows because $\alpha \leq n d \phi$ and the second inequality follows multiplying and dividing by $1+\alpha$ and because $1/(1+\alpha)<1$. Since the above expression is monotonically increasing for $\alpha$ and $\alpha \geq 1$, we have that 
\begin{align*}
n_{j}^{j} & \geq  n \frac{1}{n d \phi} \frac{2^{2 j-2} -1}{2^{2 j-2}+1} \geq n\frac{1}{n d \phi} \frac{3}{5} \; \; \; \text{because $j \geq 2$.}
\end{align*}
\end{enumerate}
\end{proof}

\section{Proof of Theorem~\ref{th:ess_recursion_continual}}\label{ap:th4}

 \begin{proof}
 To obtain bound in~\eqref{eq:ess_continual} we first prove that the mean vector estimate and the \acs{MSE} vector given by~\eqref{eq:tau_general_back},~\eqref{eq:s_general_back} satisfy  
 \begin{align}
\label{eq:pr2}
& \mathbb{P}\left\{| {\tau_k^\infty}^{(i)} - {{{\tau}}_j^{k}}^{(i)}| \leq \kappa \sqrt{2{s_j^{k}}^{(i)} \log\left(\frac{2m}{\delta}\right)}\right\}  \geq (1 - \delta)
\end{align}
for any component $i = 1, 2, \ldots, m$. Then, we prove that $\|\B{s}_j^{ k}\|_\infty \leq M/\sqrt{n_j^{ k}}$ for $j \in \{1, 2, \ldots, k\}$, where the \acsp{ess} satisfy $n_k^{ k} = n_k^k$ and $n_j^{ k} \geq n_j^j + n_{j+1}^{- k} \frac{\|\B{\sigma}_j^2\|_\infty }{\|\B{\sigma}_j^2\|_\infty  + n_{j+1}^{- k}\|\B{d}_{j+1}^2\|_\infty }$ for $j \geq 2$.

As described in Appendix~\ref{app:rec}, recursions in~\eqref{eq:tau_general_back} and~\eqref{eq:s_general_back} for $\B{\tau}_{j}^{k}$ and $\B{s}_{j}^{k}$ correspond to the Rauch-Tung-Striebel smoother recursions. An alternative manner to obtain such mean and \ac{MSE} vectors is using the fixed-lag smoother recursions that become
\begin{align}
\label{eq:tau_back_2}
      \B{\tau}_{j}^{k} & = \B{\tau}_{j}^{j} +  \frac{\B{s}_j^{j}}{\B{s}_j^{j}+ \B{s}_{j+1}^{- k} + \B{d}_{j+1}} \left( \B{\tau}_{j+1}^{- k} - \B{\tau}_{j}^{j}\right), \; \;
      \B{s}_j^{k} =\left(\frac{1}{ \B{s}_j^j} + \frac{1}{\B{s}_{j+1}^{- k} + \B{d}_{j+1}}\right)^{-1}
\end{align}
where backward mean and \ac{MSE} vectors $\B{\tau}_{j}^{- k}, \B{s}_{j}^{- k}$ are obtained using recursions in~\eqref{eq:tau_general} and~\eqref{eq:s_general} in retrodiction. Specifically, vectors $\B{\tau}_{j}^{- k}$ and $\B{s}_{j}^{- k}$ are obtained using the same recursion as for $\B{\tau}_j^j$ and $\B{s}_j^j$ in~\eqref{eq:tau_general} and~\eqref{eq:s_general} with $\B{s}_{j+1}^{- k}, \B{d}_{j+1}$, and $\B{\tau}_{j+1}^{- k}$ instead of $\B{s}_{j-1}^j, \B{d}_j$, and $\B{\tau}_{j-1}^{j-1}$.

To obtain inequality~\eqref{eq:pr2}, we prove that each component $i = 1, 2, \ldots, m$ of the error in the mean vector estimate ${{z}_{j}^{ k}}^{(i)} = {{\tau}_j^\infty}^{(i)}- {{\tau}_j^{ k}}^{(i)}$ is sub-Gaussian with parameter \mbox{${{\eta}_{j}^{ k}}^{(i)} \leq \kappa \sqrt{{{s}_{j}^{ k}}^{(i)}}$}. Analogously to the proof of Theorem~\ref{th:ess_recursion}, it is proven that each component in the error of the backward mean vector $\B{\tau}_{j+1}^{ - k}$ is sub-Gaussian with parameters satisfying $\B{\eta}_{j+1}^{ - k} \preceq {\kappa} \sqrt{\B{s}_{j+1}^{ - k}}$. The error in the forward and backward mean vector estimate is given by
\begin{align*}
{z_j^{ k}}^{(i)} & = {\tau_j^\infty}^{(i)} - {{{\tau}}_j^{ k}}^{(i)} =  {\tau_j^\infty}^{(i)} - {\tau_{j}^{j}}^{(i)} - \frac{{s_{j}^{j}}^{(i)}}{{s_{j}^{j}}^{(i)}+ {s_{j+1}^{ - k}}^{(i)} + {d_{j+1}}^{(i
)}}  \left({\tau_{j+1}^{ - k}}^{(i)} - {\tau_{j}^{j}}^{(i)}\right)
\end{align*}
where the second equality is obtained using the recursion for ${\tau_{j}^{ k}}^{(i)}$ in~\eqref{eq:tau_back_2}. Adding and subtracting $ \frac{{s_{j}^{j}}^{(i)}}{{s_{j}^{j}}^{(i)}+ {s_{j+1}^{ - k}}^{(i)} + {d_{j+1}}^{(i
)}} {\tau_{j+1}^\infty}^{(i)}$, we have that 
\begin{align*}
{z_j^{ k}}^{(i)} & = {z_{j}^{j}}^{(i)} - \frac{{s_{j}^{j}}^{(i)}}{{s_{j}^{j}}^{(i)}+ {s_{j+1}^{ - k}}^{(i)} + {d_{j+1}}^{(i
)}}   \left({\tau_{j+1}^\infty}^{(i)} -{\tau_{j+1}^\infty}^{(i)}+ {\tau_{j+1}^{ - k}}^{(i)} - {\tau_{j}^{j}}^{(i)}\right) \\
& = {z_{j}^{j}}^{(i)} - \frac{{s_{j}^{j}}^{(i)}}{{s_{j}^{j}}^{(i)}+ {s_{j+1}^{ - k}}^{(i)} + {d_{j+1}}^{(i
)}}  \left({\tau_{j}^\infty}^{(i)} + {w}_{j+1}^{(i)} - {z_{j+1}^{ - k}}^{(i)} - {\tau_{j}^{j}}^{(i)}\right)
\end{align*}
since $\B{w}_j = \B{\tau}_j^\infty - \B{\tau}_{j-1}^\infty$ and ${z_{j}^{j}}^{(i)} = {\tau_j^\infty}^{(i)} -{\tau_{j}^{j}}^{(i)}$. Then, we have that 
\begin{align}
\label{eq:backward_error}
& {z_j^{ k}}^{(i)} = {z_{j}^{j}}^{(i)} - \frac{{s_{j}^{j}}^{(i)}}{{s_{j}^{j}}^{(i)}+ {s_{j+1}^{ - k}}^{(i)} + {d_{j+1}}^{(i
)}} \left({z_{j}^{j}}^{(i)} + {w}_{j+1}^{(i)} - {z_{j+1}^{ - k}}^{(i)} \right) \\
& =  \left(1 - \frac{{s_{j}^{j}}^{(i)}}{{s_{j}^{j}}^{(i)}+ {s_{j+1}^{ - k}}^{(i)} + {d_{j+1}^2}^{(i
)}}  \right) {z_{j}^{j}}^{(i)} -  \frac{{s_{j}^{j}}^{(i)}}{{s_{j}^{j}}^{(i)}+ {s_{j+1}^{ - k}}^{(i)} + {d_{j+1}}^{(i
)}}  \left({w}_{j+1}^{(i)} - {z_{j+1}^{ - k}}^{(i)}\right) \nonumber
\end{align}
where ${z_{j}^{j}}^{(i)}$, ${z_{j+1}^{ - k}}^{(i)}$, and ${w}_{j+1}^{(i)}$ are sub-Gaussian with parameters ${\eta_{j}^{j}}^{(i)} \leq {\kappa} \sqrt{{s_{j}^{j}}^{(i)}}$, \mbox{${\eta_{j+1}^{ - k}}^{(i)} \leq {\kappa} \sqrt{{s_{j+1}^{ - k}}^{(i)}}$}, and $\sigma(w_{j}^{(i)})$, respectively. Since $\B{z}_{j}^{j}$, $\B{z}_{j+1}^{ - k}$, and $\B{w}_{j+1}$ are independent, we have that ${z_j^{ k}}^{(i)}$ given by~\eqref{eq:backward_error} 
is sub-Gaussian with parameter that satisfies
\begin{align*}
\left({\eta_j^{ k}}^{(i)}\right)^2 = & \left(1 - \frac{{s_{j}^{j}}^{(i)}}{{s_{j}^{j}}^{(i)}+ {s_{j+1}^{ - k}}^{(i)} + {d_{j+1}}^{(i)}}  \right)^2 \left({\eta_{j}^{j}}^{(i)}\right)^2 \\
&+\left(\frac{{s_{j}^{j}}^{(i)}}{{s_{j}^{j}}^{(i)}+ {s_{j+1}^{ - k}}^{(i)} + {d_{j+1}}^{(i)}}  \right)^2 \left(\sigma\left(w_{j}^{(i)}\right)^2 + \left({\eta_{j+1}^{ - k}}^{(i)}\right)^2\right)
\end{align*}
\begin{align*}
\left({\eta_j^{ k}}^{(i)}\right)^2 \leq & \left(1 - \frac{{s_{j}^{j}}^{(i)}}{{s_{j}^{j}}^{(i)}+ {s_{j+1}^{ - k}}^{(i)} + {d_{j+1}}^{(i)}}  \right)^2{\kappa^2}{s_{j}^{j}}^{(i)} \\
& +\left(\frac{{s_{j}^{j}}^{(i)}}{{s_{j}^{j}}^{(i)}+ {s_{j+1}^{ - k}}^{(i)} + {d_{j+1}}^{(i)}}  \right)^2  {\kappa^2} \left({d_{j+1}}^{(i)} + {s_{j+1}^{ - k}}^{(i)}\right).
\end{align*}
The sub-Gaussian parameter satisfies
\begin{align}
\left({\eta_j^{ k}}^{(i)}\right)^2
\leq & \left(1 - \frac{{s_j^{ k}}^{(i)}}{{s_{j+1}^{ - k}}^{(i)} + {d_{j+1}}^{(i)}}\right)^2 {\kappa^2}\left(\frac{1}{ {{s_j^{ k}}^{(i)}}} - \frac{1}{{{s_{j+1}^{ - k^2}}^{(i)}} + {d_{j+1}}^{(i)}}\right)^{-1}+ \frac{\left( {s_j^{ k}}^{(i)}\right)^2}{ {s_{j+1}^{ - k}}^{(i)} + {d_{j+1}}^{(i)}} \kappa^2 \nonumber \\
= & \left(\frac{ {s_{j+1}^{ - k}}^{(i)} + {d_{j+1}}^{(i)} - {s_j^{ k}}^{(i)}}{{s_{j+1}^{ - k}}^{(i)} + {d_{j+1}}^{(i)}}  \right) {\kappa^2} {s_j^{ k}}^{(i)} + \frac{\left({s_j^{ k}}^{(i)}\right)^2}{{s_{j+1}^{ - k}}^{(i)} + {d_{j+1}}^{(i)}} \kappa^2 = {\kappa^2} {s_j^{ k}}^{(i)}.\nonumber
\end{align}

The inequality in~\eqref{eq:pr2} is obtained using the union bound together with the Chernoff bound (concentration inequality)~(see e.g., \citet{wainwright2019high}) for the random variables ${z_j^{ k}}^{(i)}$ that are sub-Gaussian with parameter ${\eta_j^{ k}}^{(i)}$.

Now, we prove that, for any $j$, $\|\sqrt{\B{s}_j^{ k}}\| \leq M/\sqrt{n_j^{ k}}$ where the \acsp{ess} satisfy \mbox{$n_j^{ k} \geq n_j^j + n_{j+1}^{ - k} \frac{\|\B{\sigma}_j^2\|_\infty }{\|\B{\sigma}_j^2\|_\infty + n_{j+1}^{ - k}\|\B{d}_{j+1}\|_\infty }$} for $j \geq 2$. Analogously to the proof of Theorem~\ref{th:ess_recursion}, we prove that the backward \acs{MSE} vector $\B{s}_{j+1}^{ - k}$ satisfies $\|\sqrt{\B{s}_{j+1}^{ - k}}\|_\infty \leq M/\sqrt{n_{j+1}^{ - k}}$. Then, using that $\|\sqrt{\B{s}_{j+1}^{ - k}}\|_\infty \leq M/\sqrt{n_{j+1}^{ - k}}$, we have that  for every component $i$
\begin{align*}
  \left({{s}_j^{ k}}^{(i)}\right)^{-1} & = \frac{1}{ {{s}_j^j}^{(i)}} + \frac{1}{{{s}_{j+1}^{ - k}}^{(i)} + {{d}_{j+1}}^{(i)}} \geq \frac{ {n}_j^j }{{{\sigma_j^2}}^{(i)}}+ \frac{1}{\frac{M^2}{{n}_{j+1}^{ - k} } + {{d}_{j+1}}^{(i)}}\\
  & \geq \frac{1}{M^2}\left({n}_j^j + \frac{1}{\frac{1}{{n}_{j+1}^{ - k} } + \frac{{d}_{j+1}}{M^2}}\right) \geq  \frac{1}{M^2}\left({n}_j^j + \frac{1}{\frac{1}{{n}_{j+1}^{ - k} } + \frac{\|{d}_{j+1}\|_\infty}{\|\B{\sigma}_j^2\|_\infty}}\right).
\end{align*}
Then, we obtain
\begin{equation}
\label{eq:19}
\|\sqrt{\B{s}_j^{ k}}\|_\infty \leq \frac{M}{\sqrt{{n}_j^j + \frac{1}{\frac{1}{{n}_{j+1}^{ - k} } + \frac{\|\B{d}_{j+1}\|_\infty }{\|\B{\sigma}_j^2\|_\infty}}}}.
\end{equation}

The inequality in~\eqref{eq:ess_continual} is obtained because the minimax risk is bounded by the smallest minimax risk as shown in \cite{mazuelas:2020, mazuelas:2022} so that
\begin{align*}
R(\set{U}_j^{ k}) \leq R_j^\infty + \left(\|\B{\tau}_j^\infty - \B{\tau}_j^{ k} \|_\infty + \|\B{\lambda}_j^{ k} \|_\infty \right)\left\|\B{\mu}_j^\infty\right\|_1
\end{align*}
that leads to~\eqref{eq:ess_continual} 
using~\eqref{eq:pr2} and~\eqref{eq:19}.
\end{proof}

\section{Proof of Theorem~\ref{th:backward}} \label{ap:proofthbackward}

\begin{proof} 
To obtain bound in~\eqref{eq:N_bineq}, we proceed by induction. For $j = k$, we have that
$$n_j^j = n_k^k.$$
If~\eqref{eq:N_bineq} holds for the $(j+1)$-th task, the for the $j$-th task, we have that
\begin{align*}
n_{j}^{k} \geq n_j^j + \frac{n_{j+1}^{k}(1 + n_j^j  \|\B{d}_{j+1}\|_\infty) - n_j^j}{n_{j+1}^{k} \|\B{d}_{j+1}\|_\infty (1 + n_j^j \|\B{d}_{j+1}\|_\infty) + 1} \geq n_j^j + \frac{n_{j+1}^{k}(1 + n_j^j {d}) - n_j^j}{n_{j+1}^{k}{d}(1 + n_j^j{d}) + 1} 
\end{align*}
 where the second inequality is obtained because $\|\B{d}_j\|_\infty  \leq d$. Using that 
 \begin{equation*}
n_{j+1}^{k} \geq n \left(1+ \frac{(1+\alpha)^{2 j+1} -1 - \alpha}{\alpha(1+\alpha)^{2 j+1}+\alpha}+ \frac{(1+\alpha)^{2 (k-j-1)+1}- 1-\alpha}{\alpha(1+\alpha)^{2 (k-j-1)+1}+\alpha}\right)
\end{equation*}
the \ac{ess} of the $j$-th task satisfies
\begin{align*}
n_j^{ k} & \geq n_j^j + n \left(1+ \frac{(1+\alpha)^{2 (k-j)-1} - 1 - \alpha}{\alpha(1+\alpha)^{2 (k-j)-1}+\alpha}\right) \left({1 + \frac{n \left(1+ \frac{(1+\alpha)^{2 (k-j)-1} - 1 - \alpha}{\alpha(1+\alpha)^{2 (k-j)-1}+\alpha}\right)}{n d}}\right)^{-1}\\
& = n_j^j + n \frac{(1+\alpha)^{2 (k-j)}- 1}{\alpha(1+\alpha)^{2 (k-j)-1}+\alpha} \left({1 + \frac{\alpha^2}{\alpha+1}\left(1+ \frac{(1+\alpha)^{2 (k-j)-1} - 1 - \alpha}{\alpha(1+\alpha)^{2 (k-j)-1}+\alpha}\right)}\right)^{-1}
\end{align*}
where the second equality follows because $n d = \frac{\alpha^2}{\alpha+1}$ since ${\alpha} = \frac{ n d}{2} \left(\sqrt{1+\frac{4}{nd}} + 1\right)$. Then, we have that
\begin{align*}
n_j^{ k} \geq & n_j^j + n \frac{(1+\alpha)^{2 (k-j)}- 1}{\alpha(1+\alpha)^{2 (k-j)-1}+\alpha}\\
  &\cdot \left({\frac{((1+\alpha)^{2 (k-j)-1}+1)(\alpha+1+\alpha^2) + \alpha((1+\alpha)^{2 (k-j)-1} - 1 - \alpha)}{(\alpha+1)((1+\alpha)^{2 (k-j)-1}+1)}}\right)^{-1}\\
 \geq & n_j^j + n \frac{(1+\alpha)^{2 (k-j)}- 1}{\alpha(1+\alpha)^{2 (k-j)-1}+\alpha} \frac{(\alpha+1)((1+\alpha)^{2 (k-j)-1}+1)}{(1+\alpha)^{2 (k-j)+1}+1}.\end{align*}

Now, we obtain bounds for the \acs{ess} depending on the value value of $n d$. Such bounds are obtained similarly as in Theorem~\ref{th:forward} and we also denote by $\phi$ the golden ratio $\phi = 1.618\ldots$.
\begin{enumerate}
\item If ${n d} < \frac{1}{j^2}\Rightarrow \sqrt{n d} \leq \alpha \leq \sqrt{n d} \phi \leq  \frac{\phi}{j} \leq 1$ similarly as in the preceding case, then we have that  $n_{j}^{ k}$ satisfies
\begin{align*}
n_j^{ k} & \geq n_j^j + n \frac{1}{\alpha}\frac{\alpha({2 (k-j)})}{2+\alpha {2 (k-j)}} = n_j^j + n \frac{k-j}{1+\alpha {(k-j)}}  \geq n_j^j + n \frac{k-j}{1+\frac{\phi}{j} {(k-j)}}
\end{align*}
where the first inequality follows because $(1+\alpha)^{2 (k-j)-1} \geq 1 + \alpha(2 (k-j)-1)$ and the second inequality is obtained using $\alpha \leq \frac{\phi}{j}$.
\item If $ \frac{1}{j^2} \leq n d < 1 \Rightarrow \frac{1}{j}\leq \sqrt{n d} \leq \alpha \leq \sqrt{n d} \phi$ because ${\alpha} =  n d \frac{\sqrt{1 + \frac{4}{n d}} + 1}{2} = \sqrt{n d}\frac{\sqrt{n d + 4} +  \sqrt{n d}}{2}$, then we have that $n_{j}^{ k}$ satisfies
\begin{align*}
n_j^{ k} & \geq n_j^j \frac{n}{\alpha} \frac{(1+\alpha)^{2 (k-j)}- 1}{(1+\alpha)^{2 (k-j)}+1}  \geq n_j^j \frac{n}{\alpha} \frac{(1+\sqrt{nd})^{2 (k-j)}- 1}{(1+\sqrt{n d})^{2 (k-j)}+1}
\end{align*}
where the second inequality follows because  the \acs{ess} is monotonically increasing for $\alpha$ and $\alpha \geq n d$. Since $(1+  \sqrt{n d})^{2 (k-j)} \geq 1+ 2 \sqrt{n d} (k-j) $ and $k-j\geq 1$, we have that 
\begin{align*}
n_j^{ k} & \geq n_j^j + \frac{n}{\alpha} \frac{ \sqrt{n d}}{1 +  \sqrt{n d}}\geq n_j^j + {n} \frac{1}{\phi} \frac{1}{1 +  \sqrt{n d}}
\end{align*}
because $\alpha \leq \sqrt{n d} \phi$.
\item If $ n d \geq 1 \Rightarrow 1 \leq n d \leq \alpha \leq n d \phi$ because  ${\alpha} = n d \frac{\sqrt{1 + \frac{4}{n d}} + 1}{2}$, then we have that $n_{j}^{ k}$ satisfies
\begin{align*}
n_j^{ k} & \geq n_j^j + n \frac{1}{\alpha} \frac{2^{2 (k-j)}-1}{2^{2 (k-j)}+1}  \geq n_j^j + n\frac{1}{n d}\frac{1}{\phi}\frac{3}{5}
\end{align*}
where the first inequality follows because the \acs{ess} is monotonically increasing for $\alpha$ and $\alpha \geq 1$ and the second inequality is obtained using $k-j\geq 1$ and $ \alpha \leq n d \phi$.
\end{enumerate}
\end{proof}

\section{ESSs of the baseline approach that utilizes sliding windows}
\label{app:sliding_windows}

The \ac{ess} of the proposed forward learning techniques can be compared with the \ac{ess} of techniques based on sliding windows that are commonly used to adapt to evolving tasks \citep{bifet:2007, zhang2016sliding, tahmasbi2021driftsurf}. Such techniques obtain classification rules for each task using the sample sets corresponding with the $\overline{W}$ closest tasks with $\overline{W}$ the window size value. In the following, we first show the ESS using sliding windows with preceding tasks and then using sliding windows with preceding and succeeding tasks.

If the mean vector is obtained for each $j$-th task as the sample average of the $\overline{W}$ sample sets $D_{j-\overline{W}}, D_{j-\overline{W}+2}, \ldots, D_{j-1}$, then the \ac{ess} satisfies 
\begin{equation}
\label{eq:ess_wc}
n_{j, \overline{W}} \geq n \frac{6 \overline{W}}{(\overline{W} + 1) (2\overline{W}+1) n d + 6}
\end{equation}
and if the mean and confidence vectors are obtained for each $j$-th task with $j \in \{1, 2, \ldots, k\}$ as the sample average of the $\overline{W}$ closest sample sets $D_{j-\overline{W}+\widehat{W}+1}, D_{j-\overline{W}+\widehat{W}+2}, \ldots, D_{j}, \ldots, D_{j+\overline{W}-\widehat{W}}$ then, the \ac{ess} satisfies 
\begin{align*}
n_{j, \overline{W}, \widehat{W}} & \geq n \frac{6 \overline{W}}{(6 \widehat{W}^2 + 2 \overline{W}^2 + 3\overline{W}+1-6\widehat{W}\overline{W} -6\widehat{W}) nd + 6}
\end{align*}
with $1\leq \overline{W} \leq j-1$ and $\B{\sigma}_j, d$, and $n$ as in Theorem~\ref{th:forward}.
\begin{proof}
Let ${\tau}_{j, \overline{W}}^{(i)}$ be the $i$-th component of the sample average of the $\overline{W}$ closest sample sets to the $j$-th task, ${\tau}_{j, \overline{W}, \widehat{W}}^{(i)}$ be the sample average of the $\overline{W}$ sample sets closest the $j$-th task, and ${\tau}_{j}^{(i)}$ be the $i$-th component of the sample average corresponding to the $j$-th task.  We prove that  
\begin{align}
\label{eq:prob_ventana}
\mathbb{P}\left\{|{{\tau}_j^\infty}^{(i)} -\bar{\tau}_{j}^{(i)}| \leq {\kappa} \sqrt{2 \frac{1}{\bar{n}_{j}} \log\left(\frac{2m}{\delta}\right)}\right\}  \geq (1 - \delta)
\end{align}
for any component $i = 1, 2, \ldots, m$ with $\bar{\tau}_{j}^{(i)} = {\tau}_{j, \overline{W}}^{(i)}$ and $\bar{n}_{j} = n_{j, \overline{W}}$ and with $\bar{\tau}_{j}^{(i)} = {\tau}_{j, \overline{W}, \widehat{W}}^{(i)}$ and $\bar{n}_{j} =n_{j, \overline{W}, \widehat{W}}$.  For any $j = 1, 2, \ldots, k$, we have that 
\begin{align}
& {{\tau}_j^\infty}^{(i)} - {\tau}_{j, \overline{W}}^{(i)} =  {{\tau}_j^\infty}^{(i)} - \frac{1}{W}\left({\tau}_{j-\overline{W}}^{(i)} + {\tau}_{j-\overline{W}+1}^{(i)} + \ldots + {\tau}_{j-1}^{(i)}\right)\nonumber\\
= & \frac{1}{\overline{W}}\left(({{\tau}_j^\infty}^{(i)} -{\tau}_{j-\overline{W}}^{(i)}) + ({{\tau}_j^\infty}^{(i)} -{\tau}_{j-\overline{W}+1}^{(i)}) + \ldots + ({{\tau}_j^\infty}^{(i)} -{\tau}_{j-1}^{(i)})\right)\nonumber\\
\label{eq:vapp}
= & \frac{1}{\overline{W}}(({{\tau}_j^\infty}^{(i)} -{{\tau}_{j-\overline{W}}^\infty}^{(i)}- {v}_{j-\overline{W}}^{(i)}) + ({{\tau}_j^\infty}^{(i)} -{{\tau}_{j-\overline{W}+1}^\infty}^{(i)} -  {v}_{j-\overline{W}+1}^{(i)}) + \ldots \\
& + ({{\tau}_j^\infty}^{(i)} -{{\tau}_{j-1}^\infty}^{(i)} - {v}_{j-1}^{(i)}))\nonumber\\
\label{eq:wapp}
= & \frac{1}{\overline{W}}(({{\tau}_j^\infty}^{(i)} -{{\tau}_{j-\overline{W}+1}^\infty}^{(i)}+{w}_{j-\overline{W}+1}^{(i)}) + ({{\tau}_j^\infty}^{(i)} -{{\tau}_{j-\overline{W}+2}^\infty}^{(i)}+{w}_{j-\overline{W}+2}^{(i)}) + \ldots - \sum_{l = j-\overline{W}}^{k-1}{v}_{l}^{(i)}) \nonumber
\end{align}
where the first equality is by the definition of ${\tau}_{j, \overline{W}}^{(i)}$,~\eqref{eq:vapp} is due to $\B{v}_j = \frac{1}{n} \sum_{i = 1}^{n} \Phi(x_{j, i}, y_{j, i})$, and~\eqref{eq:wapp} is due to $\B{w}_j = \B{\tau}_j^\infty - \B{\tau}_{j-1}^\infty$.
Using the definition of $\B{w}_j$, we have that
\begin{align}
{{\tau}_j^\infty}^{(i)} - {\tau}_{j, \overline{W}}^{(i)} & = \frac{1}{\overline{W}} \left(\sum_{l = j-W+1}^j{w}_{l}^{(i)} + \sum_{l = j-W+2}^j {w}_l^{(i)} + \ldots + {w}_j^{(i)} - \sum_{l = j-\overline{W}}^{j-1}{v}_{l}^{(i)}\right)\nonumber\\
& = \frac{1}{\overline{W}}\sum_{l = j-\overline{W}+1}^j(l-j+\overline{W}){w}_{l}^{(i)} - \frac{1}{\overline{W}} \sum_{l = j-\overline{W}}^{j-1}{v}_{l}^{(i)}.\nonumber
\end{align}
The sub-Gaussian parameter of the error in the mean vector estimate is given by 
\begin{align}
\left({\rho}_{j, \overline{W}}^{(i)}\right)^2 & \leq \frac{1}{\overline{W}^2}\sum_{l = j-W+1}^j(l-j+\overline{W})^2\sigma(w_l^{(i)})^2+ \frac{1}{\overline{W}^2}\sum_{l = j-\overline{W}}^{j-1}\frac{\sigma(\Phi_l^{(i)})^2}{{n_l}}.\nonumber
\end{align}
Let $\kappa$ and $n_l$ be such that $\sigma(w_l^{(i)}) \preceq \kappa \sqrt{d}_{l}, \sigma(\Phi_l^{(i)}) \preceq \kappa \sigma_{l}$, and $n_l \geq n$ for any $l$, then we obtain
\begin{align}
\left({\rho}_{j, \overline{W}}^{(i)}\right)^2 & \leq \frac{1}{\overline{W}^2} \sum_{l = j-\overline{W}+1}^j(l-j+\overline{W})^2 \kappa^2 {d_{l}} + \frac{1}{\overline{W}^2}\sum_{i = j-\overline{W}}^{j-1} \kappa^2  \frac{{\sigma_{l}^2}}{{n}}\nonumber\\
  & \leq  \frac{1}{6 \overline{W}}(\overline{W}+1) (2 \overline{W} +1) \kappa^2 {d} + \kappa^2  \frac{{1}}{{\overline{W} n}}.
\end{align}
The inequality in~\eqref{eq:prob_ventana} is obtained using the union bound together with the Chernoff bound (concentration inequality) (see e.g., \citet{wainwright2019high}) for the random variables  ${{\tau}_j^\infty}^{(i)} - {\tau}_{j, \overline{W}}^{(i)}$ that are sub-Gaussian with parameter ${\rho}_{j, \overline{W}}^{(i)}$  and because $n_j \geq n$, $\|\B{\sigma}_j^2\|_\infty \leq 1$, and $\|\B{d}_j\|_\infty  \leq d$. 

Since ${\tau}_{j, \overline{W}, \widehat{W}}^{(i)}$ is given by the sample average of $\overline{W}$ sample sets, then we have that
 \begin{align}
{{\tau}_j^\infty}^{(i)} - {\tau}_{j, \overline{W}, \widehat{W}}^{(i)} = & {{\tau}_j^\infty}^{(i)} - \frac{1}{\overline{W}}\left({\tau}_{j-\widehat{W}+1}^{(i)} + {\tau}_{j-\widehat{W}+2}^{(i)} + \ldots + {\tau}_j^{(i)}+{\tau}_{j+1}^{(i)}+ \ldots + {\tau}_{j+\overline{W}-\widehat{W}}^{(i)}\right)\nonumber\\
= & \frac{1}{\overline{W}}\Big{(}({\tau_j^\infty}^{(i)} -{\tau}_{j-\widehat{W}+1}^{(i)}) + ({\tau_j^\infty}^{(i)} -{\tau}_{j-\widehat{W}+2}^{(i)}) + \ldots + ({\tau_j^\infty}^{(i)} -{\tau}_j^{(i)})\nonumber\\
& + \ldots + ({\tau_j^\infty}^{(i)} -{\tau}_{j+W-\widehat{W}}^{(i)})\Big{)}\nonumber\\
 =&  \frac{1}{W}\Big{(}({\tau_j^\infty}^{(i)} -{\tau_{j-\widehat{W}+1}^\infty}^{(i)}- {v_{j-\widehat{W}+1}}^{(i)}) + \ldots + ({\tau_j^\infty}^{(i)} -{\tau_{j}^\infty}^{(i)} - {v}_j^{(i)})\nonumber\\
 \label{eq:vapp2}
 & + \ldots + ({\tau_{j}^\infty}^{(i)} -{\tau_{j+\overline{W}-\widehat{W}}^\infty}^{(i)} - v_{j+\overline{W}-\widehat{W}}^{(i)})\Big{)}\\
= & \frac{1}{W}\Big{(}({\tau_j^\infty}^{(i)} -{\tau_{j-\widehat{W}+2}^\infty}^{(i)}+{w}_{j-\widehat{W}+2}^{(i)}) + \ldots \nonumber\\
\label{eq:wapp2}
& + ({\tau_{j}^\infty}^{(i)} -{\tau_{j+W-\widehat{W}-1}^\infty}^{(i)}-{w}_{j+\overline{W}-\widehat{W}}^{(i)}) - \sum_{l = j-\widehat{W}+1}^{j+\overline{W}-\widehat{W}}{v}_{l}^{(i)}\Big{)}
\end{align}
where the first equality is by the definition of ${\tau}_{j, \overline{W}, \widehat{W}}^{(i)}$, ~\eqref{eq:vapp2} is due to $\B{v}_k = \frac{1}{n} \sum_{i = 1}^{n} \Phi(x_{k, i}, y_{k, i})$, and~\eqref{eq:wapp2} is due to $\B{w}_k = \B{\tau}_k^\infty - \B{\tau}_{k-1}^\infty$. Using the definition of $\B{w}_k$, we have that
\begin{align*}
& {{\tau}_j^\infty}^{(i)} - {\tau}_{j, \overline{W}, \widehat{W}}^{(i)} = \frac{1}{W}\left(\sum_{l = j-\widehat{W}+2}^j{w}_{l}^{(i)} + \ldots - \sum_{l = j+1}^{j+\overline{W}-\widehat{W}}{w}_{l}^{(i)} - \sum_{l = j-\widehat{W}+1}^{j+\overline{W}-\widehat{W}}{v}_{l}^{(i)}\right)\\
& = \frac{1}{\overline{W}}\left(\sum_{l = j-\widehat{W}+2}^j (i-j+\widehat{W}-1){w}_{l}^{(i)}-\sum_{l = j+1}^{j+\overline{W}-\widehat{W}}(j+\overline{W}-\widehat{W}+1-i){w}_{l}^{(i)} - \sum_{l = j-\widehat{W}+1}^{j+\overline{W}-\widehat{W}}{v}_{l}^{(i)}\right).
\end{align*}
The sub-Gaussian parameter of the error in the mean vector estimate is given by 
\begin{align*}
(\rho_{j, \overline{W}, \widehat{W}}^{(i)})^2 \leq & \frac{1}{\overline{W}^2}\sum_{l = j-\widehat{W}+2}^j(i-j+\widehat{W}-1)^2 \sigma(w_l^{(i)})^2 \\
&+\frac{1}{\overline{W}^2}\sum_{l = j+1}^{j+\overline{W}-\widehat{W}}(j+\overline{W}-\widehat{W}+1-i)^2\sigma(w_l^{(i)})^2 + \frac{1}{\overline{W}^2}\sum_{l = j-\widehat{W}+1}^{j+\overline{W}-\widehat{W}}\frac{\sigma(\Phi_l^{(i)})^2}{n_l}.
\end{align*}
Let $\kappa$ and $n_l$ be such that $\sigma(w_l^{(i)}) \preceq \kappa \sqrt{d}_{l}, \sigma(\Phi_l^{(i)}) \preceq \kappa {\sigma}_{l}$, and $n_l \geq n$ for any $l$. Then, we have that
\begin{align*}
(\rho_{j, \overline{W}, \widehat{W}}^{(i)})^2 \leq & \frac{1}{\overline{W}^2}\sum_{l = j-\widehat{W}+2}^j(i-j+\widehat{W}-1)^2 \kappa^2 d_{l}^{(i)} \\
&+\frac{1}{\overline{W}^2}\sum_{l = j+1}^{j+\overline{W}-\widehat{W}}(j+\overline{W}-\widehat{W}+1-i)^2\kappa^2 d_{l}^{(i)} + \frac{1}{\overline{W}^2}\sum_{l = j-\widehat{W}+1}^{j+\overline{W}-\widehat{W}}\frac{\kappa^2 {\sigma_{l}^2}^{(i)}}{n} \\
\leq &\frac{\widehat{W}(\widehat{W}-1)(2 \widehat{W} - 1)}{6\overline{W}^2} d \\
& + \frac{(2 \widehat{W}^2 + 2 \overline{W}^2 + 3\overline{W}+1-\widehat{W}(4\overline{W}+3))(\overline{W}-\widehat{W})}{6\overline{W}^2} d + \frac{1}{\overline{W} n}\\
= &\frac{6 \widehat{W}^2 + 2 \overline{W}^2 + 3\overline{W}+1-6\widehat{W}\overline{W} -6\widehat{W}}{6\overline{W}} d + \frac{1}{\overline{W}n}
\end{align*}
The inequality in~\eqref{eq:prob_ventana} is obtained using the union bound together with the Chernoff bound (concentration inequality) for the random variables ${{\tau}_j^\infty}^{(i)} - {\tau}_{j, \overline{W}, \widehat{W}}^{(i)}$ that are sub-Gaussian with parameter $\rho_{j, \overline{W}, \widehat{W}}^{(i)}$ and because $n_j \geq n$, $\|\B{\sigma}_j^2\|_\infty \leq 1$, and $\|\B{d}_j\|_\infty  \leq d$. 
\end{proof}

\section{Optimization}\label{sec:lear}

Algorithm~\ref{alg:upmu} details the implementation of the optimization step of the proposed methodology.

\begin{algorithm}
\captionsetup{labelfont={bf}}
\caption{Optimization}
\label{alg:upmu}
\begin{algorithmic}
\State \textbf{Input:} \hspace{0.3cm}$\B{\tau}$, ${\B{\lambda}}$, ${\B{\mu}}$, and $\set{X}$
\State \textbf{Output:}\hspace{0.18cm}${\B{\mu}}, R(\mathcal{U})$
\For{$x \in \set{X}, \set{C} \subseteq \mathcal{Y}, \set{C}\neq\emptyset$}
\State $\V{F}, \V{h} \gets \text{append rows } \sum_{y \in \set{C}}\Phi(x, y)^{\top}/|\set{C}|, 1/|\set{C}|\text{ to } \V{F}, \V{h}$
\EndFor
\State $\B{\mu}(1) \gets {\B{\mu}}$, $\bar{\B{\mu}}{(1)} \gets {\B{\mu}}$
\For{$l = 1, 2, \ldots, K$}
\State {$a_l \gets 1/(l+1)^{3/2}$}, {$\theta_l \gets 2/(l+1)$, $\theta_{l+1} \gets 2/(l+2)$}
\State $\V{f}_i^{\top} \gets $ row of $\V{F}$ such that $\V{f}_i^{\top} \B{\mu}{(l)} - {h}^{(i)} =  {\max} \,\{ \V{F} \B{\mu}{(l)} - \V{h}\}$
\State $\bar{\B{\mu}}{(l+1)} \gets \B{\mu}{(l)} + a_l \left({\B{\tau}} - {\V{f}_i} - \B{\lambda} \text{sign} (\B{\mu}{(l)})\right)$
\State $\B{\mu}{(l+1)} \gets \bar{\B{\mu}}{(l+1)} + \theta_{l+1}(\theta_{l}^{-1} - 1)\left({\B{\mu}}{(l)} - \bar{\B{\mu}}{(l)}\right)$
\EndFor
				\State ${\B{\mu}} \gets \B{\mu}{(K+1)}$
		\State $R(\mathcal{U}) \gets 1 - {\B{\tau}}^{\top}\B{\mu} + \max \{\V{F} \B{\mu} - \V{h}\} + {\B{\lambda}}^{\top}|\B{\mu}|$
\end{algorithmic}
\end{algorithm}

\section{Extensions of the proposed methodology applied to CL}\label{app:cl_extension}

This section shows how the proposed methodology applied to \ac{CL} can be extended to situations in which a new sample set can correspond with a precedingly learned task.

A sample set $D$ corresponding with a $t$-th task for $t \in \{1, 2, \ldots, k\}$ can arrive at any time step. Since the $t$-th task is a precedingly learned task, we first update the sample average and \ac{MSE} vector $\B{\tau}_t$ and $\B{s}_t$ given by~\eqref{eq:tau1} of the $t$-th task using the new sample set; secondly, we update mean and \ac{MSE} vectors $\B{\tau}_j^j$ and $\B{s}_j^j$ as in~\eqref{eq:tau_general}-\eqref{eq:s_general} for each $j$-th task with $j \in \{t, t+1, \ldots, k\}$; and then, we update mean and \ac{MSE} vectors $\B{\tau}_j^k$ and $\B{s}_j^k$ as in~\eqref{eq:tau_general_back}-\eqref{eq:s_general_back} for each $j$-th task with $j \in \{1, 2, \ldots, k\}$. Such mean and \ac{MSE} vectors coincide with the mean and \ac{MSE} vectors obtained in Section~\ref{sec:evolution} using all samples corresponding with the $t$-th task at time step $t$. Then, taking $\B{\sigma}_j^2=\mathbb{V}\text{ar}_{\up{p}_j}\{\Phi\left(x, y\right)\}$ and $\B{d}_j = \mathbb{E}\{\B{w}_j^2\}$, the mean estimate $\B{\tau}_j^k$ is the unbiased linear estimator that has the minimum \ac{MSE}, and the updated $\B{s}_j^k$ is its \ac{MSE} (see Theorem~\ref{th:kalman}).

\section{Additional numerical results and implementation details}\label{app:res}

In this section we describe the datasets used for the numerical results in Section~\ref{sec:nr}, we provide further details for the numerical experimentations carried out, and include several additional results. Specifically, in the first set of additional results, we show the reliability of the presented performance guarantees; in the second set of additional results, we show the performance improvement leveraging information from preceding and succeeding tasks; and in the third set of additional results, we show the classification error and the running time for different hyper-parameter values. In addition, the code of the proposed methods is available on the web \url{https://github.com/MachineLearningBCAM/Supervised-learning-evolving-task-JMLR-2024}.
\begin{table}
\centering
\caption{Dataset characteristics.}
\label{tab:datasets}
\scalebox{0.8}{\begin{tabular}{llrrrrr}
\hline
Dataset & Type & Samples & $|\set{Y}|$ & Tasks & Reference \\ \hline
Rotating hyperplane & Synthetic & 30,000 & 2 &100 & \citep{pavlidis:2011}\\
BAF & Tabular & 1,000,000& 2 & 3,333 & \citep{jesus2022}\\
{Elec2} & Tabular &45,312 & 2 & 151 & \citep{bifet:2007}\\
Airlines & Tabular &539,383  & 2 & 1,797 &\citep{bifet:2010}\\
USPS & Tabular &2,930 & 2 & 9 & \citep{dekel2008forgetron}\\
Spam & Tabular &6,213 & 2 & 20 & \citep{sethi2017reliable}\\
Power supply &Tabular & 29,928 & 2 &  99 & \citep{tahmasbi2021driftsurf}\\
Yearbook & Images &37,921 & 2 & 10 & \citep{ginosar2015century} \\
ImageNet Noise & Images &12,000 & 2 & 10 &  \citep{mai2022online} \\
DomainNet & Images &6,256 & 4 & 6 & \citep{peng2019moment}\\
UTKFace & Images &23,500 & 2 & 94 & \citep{zhifei2017cvpr} \\
Rotated MNIST & Images &70,000 & 2 & 60 & \url{http://yann.lecun.com/exdb/mnist/} \\
Rotated MNIST i.i.d. & Images &70,000 & 2 & 60 & \url{http://yann.lecun.com/exdb/mnist/} \\
CLEAR & Images &10,490 & 3 & 10 & \citep{lin2021clear}\\
\hline
\end{tabular}}
\end{table}

The datasets used in Section~\ref{sec:nr} are publicly available~\citet{ginosar2015century, zhifei2017cvpr, peng2019moment, lin2021clear}, and  \url{http://yann.lecun.com/exdb/mnist/}. The summary of these datasets is provided in Table~\ref{tab:datasets} that shows the number of classes, the number of samples, and the number of tasks. In the following, we further describe the tasks and the time-dependency of each dataset used. 
\begin{itemize}
 \setlength\itemsep{0em}
\item The ``rotating hyperplane'' dataset that has been often used as benchmark for evolving environments \citep{pavlidis:2011}. 
In such dataset, a hyperplane in $D$-dimensional space $\sum_{i = 1}^D w_ix_i = 0$ rotates between consecutive tasks. Instances for $\sum_{i = 1}^D w_ix_i \geq 0$ correspond with class $1$ and instances for $\sum_{i = 1}^D w_ix_i < 0$ correspond with class $2$. In the following, we use multiple rotation angles of the hyperplane to evaluate the error in terms of the change in the distribution. 
\item The ``BAF'' dataset contains real-world bank account fraud detection information and the goal is to predict fraud or not fraud.
\item The ``Elec2'' dataset contains half-hourly energy-related information including the day of week, the time stamp, the New South Wales electricity demand, the Victoria electricity demand, and the scheduled electricity transfer between states; and the goal is to predict energy price increases or decreases  in New South Wales relative to a moving average of the last 24 hours.
\item The ``Airlines'' dataset contains records of flight schedules and the goal is to predict if a flight is delayed or not.
\item The ``USPS'' dataset contains numeric data obtained from the scanning of handwritten digits from envelopes by the U.S. Postal Service. 
\item The ``Spam'' dataset contains emails and the task is to predict if a email is malicious spam email or legitimate email.
\item The ``Yearbook'' dataset contains portraits' photographs over time and the goal is to predict males and females. Each task corresponds to portraits from one decade from 1905 to 2013.
\item The ``ImageNet noise'' dataset contains images with increasing noise over tasks and the goal is to predict if an image is a bird or a snake. The sequence of tasks corresponds to the noise factors $[0.0, 0.4, 0.8, 1.2, 1.6, 2.0, 2.4, 2.8, 3.2, 3.6]$ \citep{mai2022online}. 
\item The ``DomainNet'' dataset contains six different domains with decreasing realism and the goal is to predict if an image is an airplane, bus, ambulance, or police car. The sequence of tasks corresponds to the six domains: real, painting, infograph, clipart, sketch, and quickdraw. 
\item  The ``UTKFaces'' dataset contains face images in the wild with increasing age and the goal is to predict males and females. The sequence of tasks corresponds to face images with different ages from 0 to 116 years.
\item  The ``Rotated MNIST'' dataset contains rotated images with increasing angles over tasks and the goal is to predict if the number in an image is greater than $5$ or not. Each $j$-th task corresponds to a rotation angle randomly selected from $\left[\frac{180(j-1)}{k}, \frac{180j}{k}\right]$ degrees where $j \in \{1, 2, ..., k\}$ and $k$ is the number of tasks. 
\item  The ``Rotated MNIST'' dataset contains rotated images with increasing angles over tasks and the goal is to predict if the number in an image is greater than $5$ or not. Each $j$-th task corresponds to a rotation angle randomly selected from $\left[0, 360\right)$ degrees. 
\item The ``CLEAR'' dataset contains images with a natural temporal evolution of visual concepts in the real world and the goal is to predict if an image is soccer, hockey, or racing. Each task corresponds to one year from 2004 to 2014.
\end{itemize}

In Section~\ref{sec:nr}, we compare the results of IMRC methods with $6$ state-of-the-art-techniques \cite{lopez2017gradient, riemer2018learning, kirkpatrick2017overcoming, zhao2020handling, tahmasbi2021driftsurf, brzezinski2013reacting}. In the following, we briefly describe each method used.
\begin{itemize}
 \setlength\itemsep{0em}
\item Condor method \cite{zhao2020handling} is a technique developed for concept drift adaptation. The method provided by Zhao et al. is an ensemble method that adapts to evolving tasks by learning weighting the models in the ensemble at each time step.
\item DriftSurf method \cite{tahmasbi2021driftsurf} is a technique developed for concept drift adaptation. The method provided by Tahmasbi et al. adapts to evolving tasks by using a drift detection method. Such method allows to restart a new model when a change in the distribution is detected.
\item AUE method \cite{brzezinski2013reacting} is a technique developed for concept drift adaptation. The method provided by Brzezinski \& Stefanowski is an ensemble method that adapts to evolving tasks by incrementally updating all classifiers in the ensemble and weighting them with non-linear error functions. 
\item GEM method \cite{lopez2017gradient} is a technique developed for continual learning. The method provided by Lopez-Paz \& Ranzato learns each new task using a stochastic gradient descent with inequality constraints given by the losses of preceding tasks. Such constraints avoid the increase of the loss of each preceding tasks.
\item MER method \cite{riemer2018learning} is a technique developed for continual learning. The method provided by Riemer et al. learns each new task using sample sample sets that include random samples of preceding tasks. Such samples of preceding tasks are stored in a memory buffer. 
\item EWC method \cite{kirkpatrick2017overcoming} is a technique developed for continual learning. The method provided by Kirkpatrick et al. learns each new task regularizing the loss with regularization parameters given by the Fisher information.
\end{itemize}

   \begin{figure}
         \centering
                   \begin{subfigure}[t]{0.32\textwidth}
         \centering
                  \psfrag{Errorabcdefghijklmnopqrstuvwxyzabcdefghi}[l][l][0.35]{$R(\up{h}_j^{j-1})$}
                  \psfrag{Bound}[l][l][0.35]{$R(\set{U}_j^{j-1})$}
 \psfrag{Bound2}[l][l][0.35]{$R(\set{U}_j^{j-1})+ \left(|\B{\tau}_k^\infty - \B{\tau}_j^{j-1}| - \B{\lambda}_j^{j-1}\right)^{\top} \left|\B{\mu}_j^{j-1}\right|$}
                  \psfrag{Task}[][][0.5]{Task $k$}
                  \psfrag{Error probability}[][][0.5]{Error probability}
                  \psfrag{10}[][][0.5]{$10$}
                  \psfrag{30}[][][0.5]{$30$}
                  \psfrag{50}[][][0.5]{$50$}
                  \psfrag{70}[][][0.5]{$70$}
                  \psfrag{90}[][][0.5]{$90$}
                  \psfrag{0.5}[][][0.5]{$0.5$}
                  \psfrag{0.4}[][][0.5]{$0.4$}
                  \psfrag{0.3}[][][0.5]{$0.3$}
                  \psfrag{0.35}[][][0.5]{$0.35$}
                  \psfrag{0.45}[][][0.5]{$0.45$}
         \includegraphics[width=\textwidth]{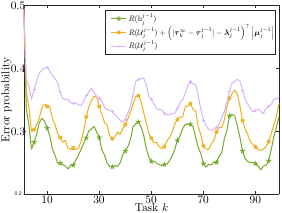}
                                \caption{Bounds for error probabilities using information from preceding tasks (SCD and MDA).}   
                                \label{fig:error_probability_100}
\end{subfigure}
\hfill
                   \begin{subfigure}[t]{0.32\textwidth}
         \centering
                  \psfrag{Errorabcdefghijklmnopqrstuvwxyzabcdefghijklmno}[l][l][0.5]{$R(\up{h}_j^{k})$}
                  \psfrag{Bound}[l][l][0.5]{$R(\set{U}_j^{k})$}
                   \psfrag{Bound2}[l][l][0.5]{$R(\set{U}_j^{k})+ \left(|\B{\tau}_j^\infty - \B{\tau}_j^{k}| - \B{\lambda}_j^{k}\right)^{\top} \left|\B{\mu}_j^{k}\right|$}
                  \psfrag{Task}[][][0.5]{Task $j$}
                  \psfrag{Error probability}[][][0.5]{Error probability}
                  \psfrag{LMRC}[l][l][0.6]{LMRC $R(\up{h}_j^{k})$}
                  \psfrag{10}[][][0.5]{$10$}
                  \psfrag{30}[][][0.5]{$30$}
                  \psfrag{50}[][][0.5]{$50$}
                  \psfrag{70}[][][0.5]{$70$}
                  \psfrag{90}[][][0.5]{$90$}
                  \psfrag{0.5}[][][0.5]{$0.5$}
                  \psfrag{0.4}[][][0.5]{$0.4$}
                  \psfrag{0.3}[][][0.5]{$0.3$}
                  \psfrag{0.2}[][][0.5]{$0.2$}
                  \psfrag{0.1}[][][0.5]{$0.1$}
                  \psfrag{0}[][][0.5]{$0$}
         \includegraphics[width=\textwidth]{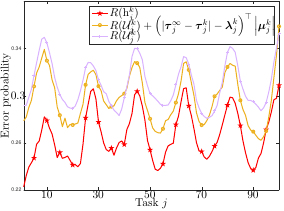}
                                \caption{Bounds for error probabilities using information from preceding, the target, and succeeding tasks (CL and MTL).}   
                                \label{fig:error_probability_continual_100}
\end{subfigure}
\hfill
     \begin{subfigure}[t]{0.32\textwidth}
                        \centering
     \psfrag{single-task learning}[l][l][0.5]{single-task learning}
         \psfrag{Forward n = 100}[l][l][0.5]{Forward $n = 100$}
         \psfrag{Forward and backward n = 10abcdefghijklmnop}[l][l][0.5]{Forward and backward $n = 10$}
         \psfrag{Forward n = 10}[l][l][0.5]{Forward $n = 10$}
           \psfrag{Forward and backward n = 100}[l][l][0.5]{Forward and backward $n = 100$}
             \psfrag{classification error/single task}[b][][0.5]{Classification error/single-task}
                                             \psfrag{10}[][][0.5]{10}
                                   \psfrag{30}[][][0.5]{30}
                                     \psfrag{50}[][][0.5]{50}
                                       \psfrag{70}[][][0.5]{70}
                                         \psfrag{90}[][][0.5]{90}
                                           \psfrag{0.1}[][][0.5]{0.1}
                                             \psfrag{0.8}[][][0.5]{0.8}
                                               \psfrag{0.6}[][][0.5]{0.6}
                                                 \psfrag{0.4}[][][0.5]{0.4}
                                                   \psfrag{1}[][][0.5]{1}
                                \psfrag{number of tasks}[][][0.5]{Number of tasks $k$}
         \includegraphics[width=\textwidth]{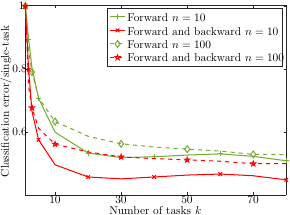}
              \caption{Classification error per number of tasks in ``Yearbook'' dataset.}
                                  \label{fig:n_tasks}
     \end{subfigure}
              \caption{Bounds for error probabilities with $n = 100$ samples per task and forward and backward learning can sharply boost performance and ESS as tasks arrive.}
                       \label{fig:Bounds}
\end{figure}

In the first set of additional results we further illustrate the reliability of the bounds using the synthetic dataset described in Section~\ref{sec:nr}. Specifically, we extend such results for the proposed methodology with $n = 100$ samples per task completing those in the main paper with $n = 10$ samples per task. Figures~\ref{fig:error_probability_100} and~\ref{fig:error_probability_continual_100} show, for each task, the averaged instantaneous bounds of probabilities of error in comparison with the true probabilities of error $R(\up{h}_{j}^{j-1})$ and $R(\up{h}_{j}^{k})$, respectively. Such figures show that the bounds $R(\set{U}_{j}^{j-1})$ and $R(\set{U}_{j}^{k})$ obtained at learning can offer accurate estimates for the probability of error of each task.
   
In the second set of additional results, we analyze the contribution of forward and backward learning to the final performance of the proposed methodology applied to \ac{MTL} and \ac{CL} using ``Yearbook'' dataset. In particular, we show the relationship among classification error, number of tasks, and sample size for single-task, forward, and forward and backward learning. These numerical results are obtained computing the classification error over all the sequences of consecutive tasks of length $k$ in the dataset. Then, we repeat such experiment $10$ times with randomly chosen training sets of size $n$. Figures~\ref{fig:n_tasks} shows the classification error of the proposed method divided by the classification error of single-task learning for different number of tasks with $n = 10$ and $n = 100$ sample sizes  using ``Yearbook'' dataset. Such figure shows that forward and backward learning can significantly improve the performance increasing the number of tasks. 

\begin{table}
\centering
\caption{Classification error of the proposed method using forward and backward learning varying $W$ and $b$.}
\label{tab:comp}
\setstretch{1.3}
\begin{adjustbox}{width=\textwidth,center}
\begin{tabular}{lrrrrrrrrrrrrrrrr}
\hline
Hyper-parameter                      & \multicolumn{2}{c}{$W = 2$} & \multicolumn{2}{c}{$W = 4$} & \multicolumn{2}{c}{$W = 6$} & \multicolumn{2}{c}{$b = 1$} & \multicolumn{2}{c}{$b = 2$} & \multicolumn{2}{c}{$b = 3$} & \multicolumn{2}{c}{$b = 4$} & \multicolumn{2}{c}{$b = 5$} \\
\hline
\multicolumn{1}{c}{Sample size $n$}& \multicolumn{1}{c}{10}& \multicolumn{1}{c}{100}& \multicolumn{1}{c}{10}& \multicolumn{1}{c}{100} & \multicolumn{1}{c}{10}& \multicolumn{1}{c}{100} & \multicolumn{1}{c}{10}& \multicolumn{1}{c}{100}& \multicolumn{1}{c}{10}& \multicolumn{1}{c}{100}& \multicolumn{1}{c}{10}& \multicolumn{1}{c}{100} & \multicolumn{1}{c}{10}& \multicolumn{1}{c}{100} & \multicolumn{1}{c}{10}& \multicolumn{1}{c}{100}\\ 
\hline
Yearbook       & .13         & .08                         & .13          & .09                      & .13                    & .09             & .14                & .10              & .09          & .14                         & .13                   & .08                 & .13        & .08                                            & .13          & .08                      \\
                             
{ImageNet noise} & .15                & .09                           & .15     & .09                          & .15     & .09                                & .15              & .09                      & .15                            & .09          & .15                  & .09                                          & .15     & .08                                      & .15                  & .08                  \\
       {DomainNet}      & .34       & .28                              & .32       & .27                       & .33                         & .28                  & .36       & .29                         & .35                          & .28      & .34       & .28                         & .34                & .28                       & .34           & .28                            \\
                             
{UTKFaces}       & .10           & .10                      & .10             & .10                    & .10                 & .10                & .10          & .10                       & .10                           & .10      & .10      & .10                         & .10         & .10                        & .10              & .10                   \\
                                {Rotated MNIST}  & .36      & .21                              & .35           & .21                   & .36        & .21                              & .36             & .22                    & .36         & .22                         & .36         & .21                           & .36              & .21                     & .36                  & .21           \\
{CLEAR}          & .09       & .05                     & .09      & .05                       & .09         & .06                   & .10         & .05                         & .09          & .05                        & .09           & .05                     & .09                 & .05                   & .08        & .05                      \\
                                \hline
\end{tabular}
\end{adjustbox}
\end{table}

In the third set of additional results, we assess the change in classification error and the running time varying the hyper-parameters. Table~\ref{tab:comp} shows the classification error of the proposed methodology using forward and backward learning varying the values of hyper-parameter for the window size $W$ and the number of backward steps $b$, completing those in the paper that show the results for $W = 2$ and $b = 3$. As shown in the table, the proposed methodology do not require a careful fine-tuning of hyper-parameters and similar performances are obtained by using different values. In addition, Table~\ref{tab:time} shows the mean running time per task in seconds for $b = 1, 2, \ldots, 5$ backward steps in comparison with the state-of-the-art techniques that learn a sequence of tasks. Such table shows that the methods proposed for backward learning do not require a significant increase in complexity, and the running time of the proposed method is similar to that of other state-of-the-art methods. 

\begin{table}[]
\centering
\caption{Running time of the proposed method using forward and backward learning in comparison with the state-of-the-art-techniques.}
 \label{tab:time}
\setstretch{1.3}
\begin{adjustbox}{width=\textwidth,center}\begin{tabular}{lrrrrrrrrrrrrrrrrrr}
\hline
Dataset                       &\multicolumn{2}{c}{GEM}  & \multicolumn{2}{c}{MER}& \multicolumn{2}{c}{EWC}&  \multicolumn{2}{c}{$b = 1$} & \multicolumn{2}{c}{$b = 2$} & \multicolumn{2}{c}{$b = 3$}  & \multicolumn{2}{c}{$b = 4$} & \multicolumn{2}{c}{$b = 5$} \\ 
\hline
\multicolumn{1}{c}{Sample size $n$}& \multicolumn{1}{c}{10}& \multicolumn{1}{c}{100} &  \multicolumn{1}{c}{10}& \multicolumn{1}{c}{100} & \multicolumn{1}{c}{10}& \multicolumn{1}{c}{100} & \multicolumn{1}{c}{10}& \multicolumn{1}{c}{100} & \multicolumn{1}{c}{10}& \multicolumn{1}{c}{100} & \multicolumn{1}{c}{10}& \multicolumn{1}{c}{100}& \multicolumn{1}{c}{10}& \multicolumn{1}{c}{100}& \multicolumn{1}{c}{10}& \multicolumn{1}{c}{100} \\ \hline
Yearbook  &  0.10  &0.48&   0.17 &3.73  & 0.36  &3.03     & 0.10                   & 0.32                                      & 0.11                     & 0.40                                      & 0.13               & 0.49                                        & 0.17                      & 0.58                                               & 0.18         & 0.66                                          \\
                           ImageNet noise  &0.01&0.04&0.08&1.05 &0.03&0.25 & 0.26               & 0.49                                       & 0.26                 & 0.53                                          & 0.28       & 0.56                                                   & 0.30                           & 0.59                                & 0.44                             & 0.60                         \\                
{DomainNet}      &0.01&0.02&0.07&0.90&0.02&0.16  & 0.52              & 8.46                                         & 0.54          & 8.98                                              & 0.54         & 9.51                                                     & 0.56           & 9.70                                        & 0.57     & 9.92                                                   \\
                               {UTKFaces}     &0.31&0.18&0.13&3.46& 0.25&2.25    & 0.11          & 0.35                                                     & 0.12             & 0.40                                                & 0.13           & 0.49                                                   & 0.16                                                & 0.57                & 0.19            & 0.66                                        \\
                                
{Rotated MNIST}  &0.18&1.09&0.21&4.09&0.59&5.30 & 0.14             & 0.47                                                    & 0.17                                & 0.60                                 & 0.21      & 0.74                                                         & 0.25                       & 0.88                                     & 0.29     & 1.01                                                       \\
                                
{CLEAR}         &0.01&0.03&0.07&1.06&0.03&0.25  & 0.24              & 1.31                                                   & 0.25     & 1.41                                                              & 0.26          & 1.47                                                    & 0.27           & 1.60                                                & 0.36      & 1.69                                                 \\
                                \hline                    
\end{tabular}
\end{adjustbox}
\end{table}

 \
\newpage
\bibliography{biblio}

\end{document}